\documentclass[10pt,twocolumn,letterpaper]{article}
\pdfoutput=1

\usepackage{3dv}
\usepackage{times}
\usepackage{epsfig}
\usepackage{subcaption}
\usepackage{graphicx}
\usepackage{amsmath}
\usepackage{amssymb}
\usepackage{xcolor}

\usepackage{multirow}
\usepackage{bm}

\usepackage{titling}

\pretitle{\vspace*{-3pt}\begin{center}\Large\bf}
\posttitle{\par\vspace*{24pt}\end{center}\vspace*{-24pt}}

\usepackage[pagebackref=true,breaklinks=true,letterpaper=true,colorlinks,bookmarks=false]{hyperref}

\usepackage{subfiles}
\usepackage{authblk}

\DeclareMathOperator*{\argmin}{arg\!\min}

\threedvfinalcopy 

\def\threedvPaperID{62} 

\ifthreedvfinal\fi
\begin{document}

\date{}

\pagenumbering{gobble}
%
\title{Learning to Align Images using Weak Geometric Supervision}
%
%
\author{Jing Dong$^{1,2}$ \quad Byron Boots$^{1}$ \quad Frank Dellaert$^{1}$ \quad Ranveer Chandra$^{2}$ \quad Sudipta N. Sinha$^{2}$}
\affil{$^1$ Georgia Institute of Technology \qquad $^2$ Microsoft Research}

\maketitle

\begin{abstract}
Image alignment tasks require accurate pixel correspondences,
which are
usually recovered by matching local feature descriptors.
Such descriptors are often derived using supervised learning on
existing datasets with ground truth correspondences.
However, the cost of creating such datasets is usually prohibitive.
In this paper, we propose a new approach to align two
images related by an unknown 2D homography where
the local descriptor is learned from scratch from the images and the
homography is estimated simultaneously. Our key insight is that
a siamese convolutional neural network can be trained jointly
while iteratively updating the homography parameters by
optimizing a single loss function. Our method is currently weakly
supervised because the input images need to be roughly aligned.

We have used this method to align images of different modalities
such as RGB and near-infra-red (NIR) without using any prior labeled
data. Images automatically aligned by our method were then used to
train descriptors that generalize to new images. We also evaluated
our method on RGB images. On the HPatches benchmark~[2], our method
achieves comparable accuracy to deep local descriptors that were
trained offline in a supervised setting.
\end{abstract}

\section{Introduction}


\begin{figure}[t]
\begin{center}
\includegraphics[width=1\linewidth]{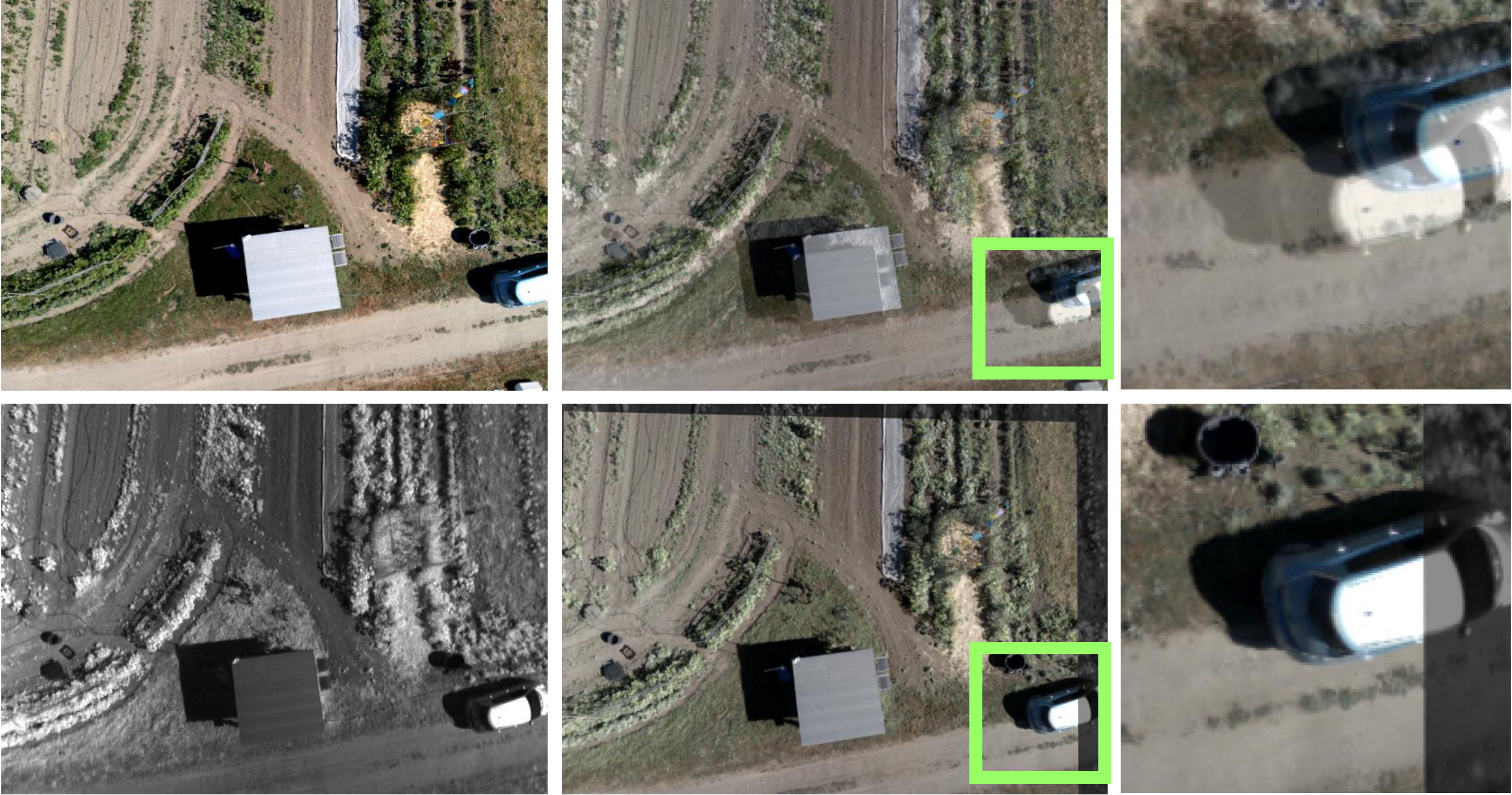}
\hspace{2cm}(a) \hspace{2.6cm} (b) \hspace{2.2cm} (c)
\end{center}
\vspace{-6mm}
\caption{Aligning images of a farm taken with a UAV. (a) RGB (top) and NIR (bottom) input images. (b) Original alignment (top) and our result (bottom). (c) Zoom-in view.}
\label{fig:idea}
\end{figure}

Finding pixel correspondences between multiple images is a fundamental problem in computer vision.
It is a crucial ingredient in 3D reconstruction, object recognition, image analysis and editing.
The correspondence problem is usually solved by extracting sparse or dense local feature
descriptors in the images and matching the descriptors across images, sometimes using additional
geometric constraints. Features, such as SIFT~\cite{Lowe2004}, SURF~\cite{Bay2008}, DAISY~\cite{Tola2010} etc.
provide partial invariance to scale, viewpoint and lighting change
and have led to great progress in image matching.

Invariant descriptors could also be learned using data-driven approaches based on convolutional
neural networks (CNN)~\cite{Simo2015,Zagoruyko2015,Han2015,Balntas2016,Balntas2016tfeat,Kumar2016,Yi2016}.
These CNN-based descriptors are trained in a fully supervised setting using datasets containing
ground truth image correspondence information~\cite{Brown2011}. Such datasets have been created
by leveraging techniques for sparse and dense 3D reconstruction that relied on classical SIFT features
~\cite{Winder2007}. However, these datasets only contain corresponding RGB image patches of the same
scene. Therefore, they are unsuitable for training descriptors that can be used for aligning images
of different modality~\cite{Hermosillo2002,Shen2014}, or aligning semantically related images with different
appearance\cite{Liu2016,Taniai2016,Ham2016,Rocco2017a,Rocco2017b}. The main difficulty is that
recovering image correspondence and alignment in other situations requires datasets with ground
truth that either do not exist or are prohibitively costly to acquire.

In this paper, we investigate the problem of aligning two images of a rigid scene when neither
handcrafted nor pretrained features are available. Assuming that we know the images can be
aligned using a specific family of geometric transforms, we propose to learn local feature
descriptors from scratch while simultaneously estimating the parameters of the geometric
transform specific to the image pair.

The core idea in our work is that of jointly learning the descriptor \ie training a
siamese convolutional neural network (CNN) and simultaneously estimating the parameters
of the geometric transform. To that end, we propose to solve a joint optimization problem
where the updates to the geometric parameters are also computed using backpropagation
in the same way as for the network parameters. In this work, we assume the transform to be a
2D homography and that the input images are roughly aligned, which makes our
problem setting weakly supervised. We implement a multi-scale version of the optimization
on image pyramids in order to handle images with greater initial misalignments.

Even though our training procedure resembles that for supervised learning with siamese networks,
there are some important differences. First, the training sets for standard siamese networks
are typically fixed and only the weights are updated using backpropagation and SGD. However, in
our case, the set of positive pairs in our training set changes during training \ie. our estimate of the \textit{true}
correspondences gradually improves as the patches are resampled using the updated homography
estimate computed in the previous iteration. The improvements to the homography
also reduces the total training loss, as the progressively learned embedding produces
a more well defined separation between the positive and negative pairs of patches from
the two images.

Secondly, when an image pair is successfully aligned by our method, the trained network parameters 
are likely to be unreliable due to overfitting on the limited training data in the two images.
However, with these alignments, we can now automatically build up a dataset with
precise correspondences which could then be used to train a local descriptor with better
generalization performance to new scenes. We evaluate this idea first for aligning
RGB--NIR images from scratch. We also evaluate on color images in the HPatches benchmark~\cite{Balntas2017}
and show that our alignments computed from scratch given weak supervision are competitive
to those computed by state of the art supervised methods that were trained offline on massive
amounts of data.

\section{Related Work}

\noindent \textbf{Local Descriptors.}
Hand-designed local features are still widely used in computer vision, \eg
SIFT~\cite{Lowe2004}, SURF~\cite{Bay2008}, DAISY~\cite{Tola2010}, DSP-SIFT~\cite{Dong2015}
and A-SIFT~\cite{Yu2011}. While they offer partial invariance and computational efficiency,
the use of learning in descriptor design promises even higher accuracy and robustness.
Winder et al.~\cite{Winder2007} proposed to tune descriptor hyperparameters using a training
set of ground truth correspondences to improve accuracy. Brown et al.~\cite{Brown2011} extended
the framework to learn descriptors that minimized the error of a nearest neighbor classifier.
Later, Simonyan et al~\cite{Simonyan2014} showed that further accuracy could be obtained
using convex optimization to learn good strategies for pooling and dimensionality reduction.

\noindent \textbf{Siamese Networks and CNNs.}
Bromley et al.~\cite{Bromley1994} proposed siamese networks for verifying whether image pairs were related.
These neural networks output feature vectors which can be compared using a suitable metric.
Two identical networks (\ie with parameters tied) are trained with matching and non-matching pairs
and the network weights are tuned to learn invariance. Siamese networks were used to learn
discriminative metrics~\cite{Chopra2005}, learn nonlinear dimensionality reduction~\cite{Hadsell2006} and has
seen a resurgence in recent years for learning local descriptors, that we discuss next.

Simo et al.~\cite{Simo2015} use siamese networks to train CNN descriptors, proposing an efficient
training procedure that searches for difficult positive and negative pairs and includes them
while training their model iteratively. DeepCompare~\cite{Zagoruyko2015}, MatchNet~\cite{Han2015}
both use CNNs to compute descriptors but also to compute the similarity score. In principle
they are related to MC-CNN~\cite{zbontar2016} proposed for dense stereo matching.
These method tend to be more accurate but much more computationally expensive.
DeepCompare~\cite{Kumar2016}, PN-Net~\cite{Balntas2016} and TFeat~\cite{Balntas2016tfeat} replace siamese networks with
triplets networks. The triplet loss encourages positives pairs to have higher similarity
compared to negative pairs such that the pairs share a data point which provides better
context. LIFT~\cite{Yi2016} is yet another CNN model trained to predict 2D keypoints with
orientations in addition to descriptors.
All these methods use supervised learning and need ground truth correspondence. In contrast,
we do not need ground truth but assume that the input images are
related via an unknown 2D homography and are approximately aligned.

CNNs have been used in other ways for image correspondence problems.
Deep Matching~\cite{Weinzaepfel2013} is a multiscale semi-dense matching method
that, inspired by CNNs, interleaves convolutions and max-pooling on image pyramids.
Later, it was extended to a fully-trainable variant~\cite{Thewlis2016}.
Rocco et al~\cite{Rocco2017a,Rocco2017b} propose methods for semantical alignment of
arbitary image pairs going from a supervised to a weakly supervised setting. Although,
our work shares a similar motivation, our network is designed to learn local descriptors
in the siamese setting whereas their CNN architecture are trained to learn global
representations.

\noindent \textbf{Datasets.}
The recent siamese networks have mostly been trained on the Multi-View Stereo Correspondence (MVSCorr) dataset~\cite{Brown2011}
built using multi-view stereo and structure from motion on Internet photo collections.
Schmidt et al.~\cite{Schmidt2017} used dense RGB-D SLAM instead of multi-view stereo in a similar spirit
and Bergamo et al.~\cite{Bergamo2013} leveraged correspondences obtained from structure from motion
to train discriminative codebooks for place recognition.
The HPatches dataset~\cite{Balntas2017} was created to benchmark local descriptor performance
and provides consistent evaluation metrics for different tasks like patch retrieval and image matching.
Another recent benchmark~\cite{Schonberger2017} measures the impact of different descriptors
by checking whether they lead to more accurate image-based 3d reconstructions.

\noindent \textbf{Multi-modal image alignment.}
Mutual information is a widely used method for image alignment~\cite{Viola1997},
popular for medical image registration~\cite{Maes1997}. However, the procedure
is iterative and relies on good initialization and more difficult to use on
natural color images or when matching local patches~\cite{Andronache2008}.
While variational methods for multi-modal image matching were explored earlier on~\cite{Hermosillo2002},
recent focus has been on learning keypoints and descriptors suitable for multispectral
images~\cite{Firmenichy2011,Brown2011b} or developing robust matching costs~\cite{Shen2014} for dense
multi-modal optical flow estimation. For multi-modal images, computing
an automatic initialization robust to arbitrary differences in image scale, position,
rotation etc. is still challenging although robust bootstrapping techniques have
shown promise~\cite{Yang2007}.

\noindent \textbf{Other learning-based methods.}
Techniques for learning to align images have been studied before.
Miller et al.~\cite{Miller2006} proposed \textit{congealing}, a procedure
to jointly align multiple images by minimizing entropy across stacks of aligned pixels.
Huang et al. extended this line of work to the unsupervised setting using handcrafted features
~\cite{Huang2007} and later by learning the features from scratch~\cite{Huang2012}.
These methods work for well defined categories such as faces, and can be used to automate dataset creation~\cite{Yi2014}.
Our work has a similar motivation but we focus on matching images of arbitrary scenes.
Recently, the inverse compositional Lucas-Kanade algorithm was used with convolutional feature maps~\cite{chang2017clkn}
and was also used to improve spatial transformer networks~\cite{lin2017inverse}. However, these models must be trained in
a supervised manner. Our focus in this work is instead on weakly supervised learning for local descriptors.

\section{Preliminaries}

Let us briefly review how local descriptors are learned using siamese networks and how
these networks are trained. The neural network takes square patches as input and
outputs a feature descriptor vector.
We denote $n\,\times\,n$ pixel patches taken from a $c$ channel image
as $\mathbf{x}\,\in\,\mathbb{R}^{n\,\times\,n\,\times\,c}$.
The network outputs a $d$--dimensional vector that is denoted as
$\bm{f}(\mathbf{x};\theta) \in \mathbb{R}^{d}$.
Here, $\theta$ is a vector denoting the network parameters (or weights).
We sometimes use $\bm{f}(\mathbf{x})$ instead of $\bm{f}(\mathbf{x};\theta)$
for notational brevity.

The training data consists of two sets of pairs of patches. Each patch pair is denoted by
$\{\mathbf{x}, \mathbf{x}^\prime\}$. The first set denoted by $\mathcal{P}$ contains true
correspondences \ie pairs of patches from different images that depict the same scene
point or object. The negative set denoted by $\mathcal{N}$ contains pairs of patches
that are not corresponding and depict different scene points and objects.
Training the network involves learning an embedding, where
$\Vert \bm{f}(\mathbf{x}) - \bm{f}(\mathbf{x}^\prime) \Vert_2$ is small for positive pairs $\{\mathbf{x}, \mathbf{x}^\prime\} \in \mathcal{P}$, and
the distance is large for negative pairs $\{\mathbf{x}, \mathbf{x}^\prime\} \in \mathcal{N}$.
The loss function used during training is called contrastive loss~\cite{Simo2015,Balntas2016, Balntas2016tfeat,Hadsell2006}.
It is defined for input pairs and has the following form.
\begin{equation}
\begin{split}
\bm{L}_0(\mathbf{x},\mathbf{x}^\prime; \theta) & = \Vert \bm{f}(\mathbf{x};\theta) - \bm{f}(\mathbf{x}^\prime;\theta)\Vert_2\\
\bm{L}_1(\mathbf{x},\mathbf{x}^\prime; \theta) & = \max(\,0,\,\mu - \Vert \bm{f}(\mathbf{x};\theta) - \bm{f}(\mathbf{x}^\prime;\theta)\Vert_2)
\end{split}
\label{eq1}
\end{equation}
The hyperparameter $\mu$ denotes a margin and is often set to the value $1$.
The loss function $\bm{L}_0$ encourages pairs of vectors to have smaller pairwise distances.
In contrast, the hinge loss $\bm{L}_1$ encourages the pairwise distances to increase and imposes
a penalty when those distances become smaller than the margin $\mu$.
The network parameters $\theta$ are computed by solving the following optimization problem.
\begin{equation}
\argmin_{\theta} \Big( \sum_{i\,=\,1}^{|\mathcal{P}|} \bm{L}_0(\mathbf{x}_i, \mathbf{x}_i^\prime; \theta) + \sum_{j\,=\,1}^{|\mathcal{N}|} \bm{L}_1(\mathbf{x}_j, \mathbf{x}_j^\prime; \theta) \Big)
\label{eq:argmin-siamese}
\end{equation}

Stochastic gradient descent (SGD) is often used for this problem and cross-validation is
used to pick the best model.

\section{Algorithm}\label{sec:algo}

Given an image pair $\{I, I^\prime\}$, we assume that we know the family of 2D warping
functions $\mathbf{w}(\cdot\,; \psi)\!:\!\mathbb{R}^2\!\to\!\mathbb{R}^2$ with parameter
vector $\psi$ that transforms a pixel location or a patch in $I$ to the corresponding
location or patch in $I^\prime$. We will now present our method to estimate the parameters
$\psi$ while simultaneously computing the network parameters $\theta$. The resulting
embedding $\bm{f}(\mathbf{x}; \theta)$ will be specific to $I$ and $I^\prime$.

Before describing our weakly supervised method is full detail, we first discuss a simpler
learning problem, that of learning $\theta$ when the parameter $\psi$ is known. This problem
can be solved using supervised learning. We then motivate how this method can be extended
to the weakly supervised setting and finally present the proposed approach to learn $\psi$
and $\theta$ jointly using a joint optimization framework. Finally, we also present a hybrid
variant of siamese and pseudo siamese networks which is useful in cases where the input image
modalities differ a lot \eg RGB and NIR.

\subsection{Descriptor learning from an aligned image pair}\label{sec:sup-siamese}

When the alignment parameters $\psi$ for an image pair are known, we can use the warping
function to extract corresponding patches from the images to train $\bm{f}(\mathbf{x; \theta})$.
We first select a set of image patches in the first image $I$. For each such patch $\mathbf{x}$
in $I$, we find the warped patch $\mathbf{w}(\mathbf{x}; \psi)$ in $I^\prime$ and insert them
into the set of positive pairs $\mathcal{P}$. Similarly, we construct the set of negative pairs
$\mathcal{N}$ by randomly sampling patches in $I^\prime$ whose location in the image are at
least $\tau$ pixels away from the location of the true corresponding patch. The model can now
be trained by minimizing a pairwise loss function, similar to the one in Equation~\ref{eq1}.
\begin{equation}
\begin{split}
\bm{L}_0(\mathbf{x}; \psi, \theta) & =\Vert \bm{f}(\mathbf{x}; \theta) - \bm{f}(\mathbf{w}(\mathbf{x}; \psi); \theta) \Vert_2 \\
\bm{L}_1(\mathbf{x}, \mathbf{x}^\prime; \theta) & = \max(\,0,\,\mu - \Vert \bm{f}(\mathbf{x}; \theta) - \bm{f}(\mathbf{x}^\prime; \theta) \Vert_2)
\end{split}
\label{eq:loss-sup}
\end{equation}
The network parameters are learned by minimizing the following objective over the pairs in
$\mathcal{P}$ and $\mathcal{N}$.
\begin{equation}
\argmin_{\theta} \Big( \sum_{i\,=\,1}^{|\mathcal{P}|} \bm{L}_0(\mathbf{x}_i; \psi, \theta) + \sum_{j\,=\,1}^{|\mathcal{N}|} \bm{L}_1(\mathbf{x}_j, \mathbf{x}^\prime_j; \theta) \Big)
\label{eq:argmin-siamese-sup}
\end{equation}
This method still uses supervised learning but does so in an uncommon setting.
The embedding $\bm{f}(\mathbf{x}; \theta)$ is being learned from only a single
image pair. Hence, the training set is small and the model is very likely
to overfit to the data. When the alignment parameter $\psi$ is accurate, the
embedding is still meaningful for this image. We will now analyze what happens
when the alignment is inaccurate and the correspondences it induces are imprecise.

\subsection{From supervised to weakly supervised}\label{sec:weak-sup-siamese}

\begin{figure}[t]
\begin{center}
\includegraphics[width=0.9\linewidth]{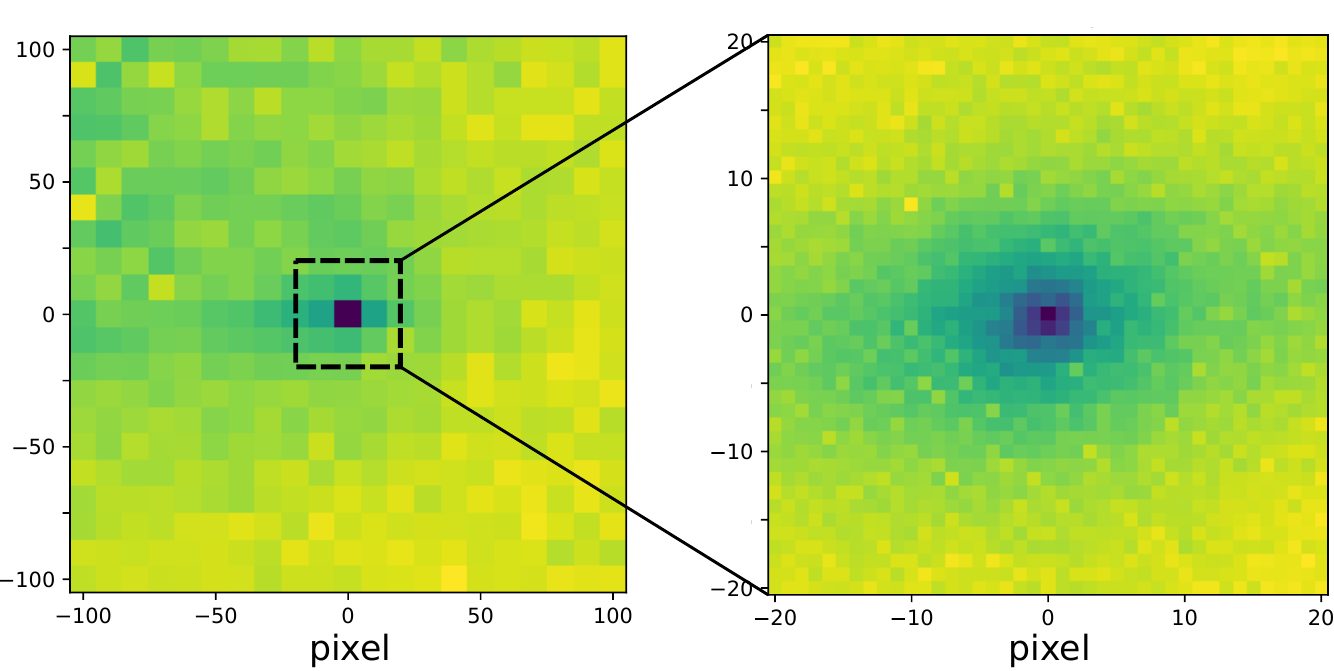}
\end{center}
\caption{Visualization of final training losses (see Eq.~\ref{eq:loss-sup}) for supervised method on an image pair, where 2D
translations (up to $\pm$ 100 pixels) were added to the true locations in the second image. Darker color indicates lower loss.}
\label{fig:pairwise-loss-vis}
\end{figure}

To simulate the effect of imprecise alignment on an image pair, we added
$x$ and $y$ translational offsets to the true warping function to generate several
imprecise candidates.
For each alignment candidate, we applied the supervised method just described previously to
learn a different embedding $\bm{f}(\mathbf{x}; \theta)$ from scratch and recorded
the loss obtained at the end of training. Figure~\ref{fig:pairwise-loss-vis} shows
a visualization of the training loss for all $x$ and $y$ offsets up to $\pm$ 100 pixels.

The visualization shows the effect of misalignment on the training loss. There is
a well defined minimum at the center, \ie when there is no misalignment. More over,
the cost surface is smooth and has a stable gradient near the center and does
not have any significant spurious local minima. This observation motivated the
question. \emph{Can we minimize the same pairwise loss to also iteratively estimate
the alignment $\psi$ as we train the neural network?}

This leads to our \emph{self-supervised} method to jointly learn the descriptor and the warping parameters. We still use the pairwise loss function defined in Eq.~\ref{eq:loss-sup}, but the warping parameter $\psi$ is no longer assumed to be known. Instead it is a variable in the loss function optimization.
\begin{equation}
\theta^*, \psi^* =
\argmin_{\theta, \psi}
\Big( \sum_{i\,=\,1}^{|\mathcal{P}|} \bm{L}_0(\mathbf{x}_i; \psi, \theta) + \sum_{j\,=\,1}^{|\mathcal{N}|} \bm{L}_1(\mathbf{x}_j, \mathbf{x}^\prime_j; \theta) \Big)
\label{eq:argmin-selfsup}
\end{equation}
%

Our self-supervised method does not need any transformation supervision. However, the training loss as shown
will in general be nonconvex and local minima can exist. With iterative optimization techniques
based on gradient descent, there is no guarantee of convergence to the correct estimate of $\psi$ starting
from an arbitrary initial value. Thus, our method is \emph{weakly supervised}. It needs a reasonable initialization
of $\psi$ or requires the input images to be roughly aligned. While this sounds like a limitation, coarse
alignment is available in many cases and our empirical evidence shows that our method can handle a fairly
large amount of misalignment.

\subsection{Joint alignment and weakly supervised learning}
\label{sec:weak-sup-implement}

\begin{figure}[t]
\begin{center}
\includegraphics[width=0.96\linewidth]{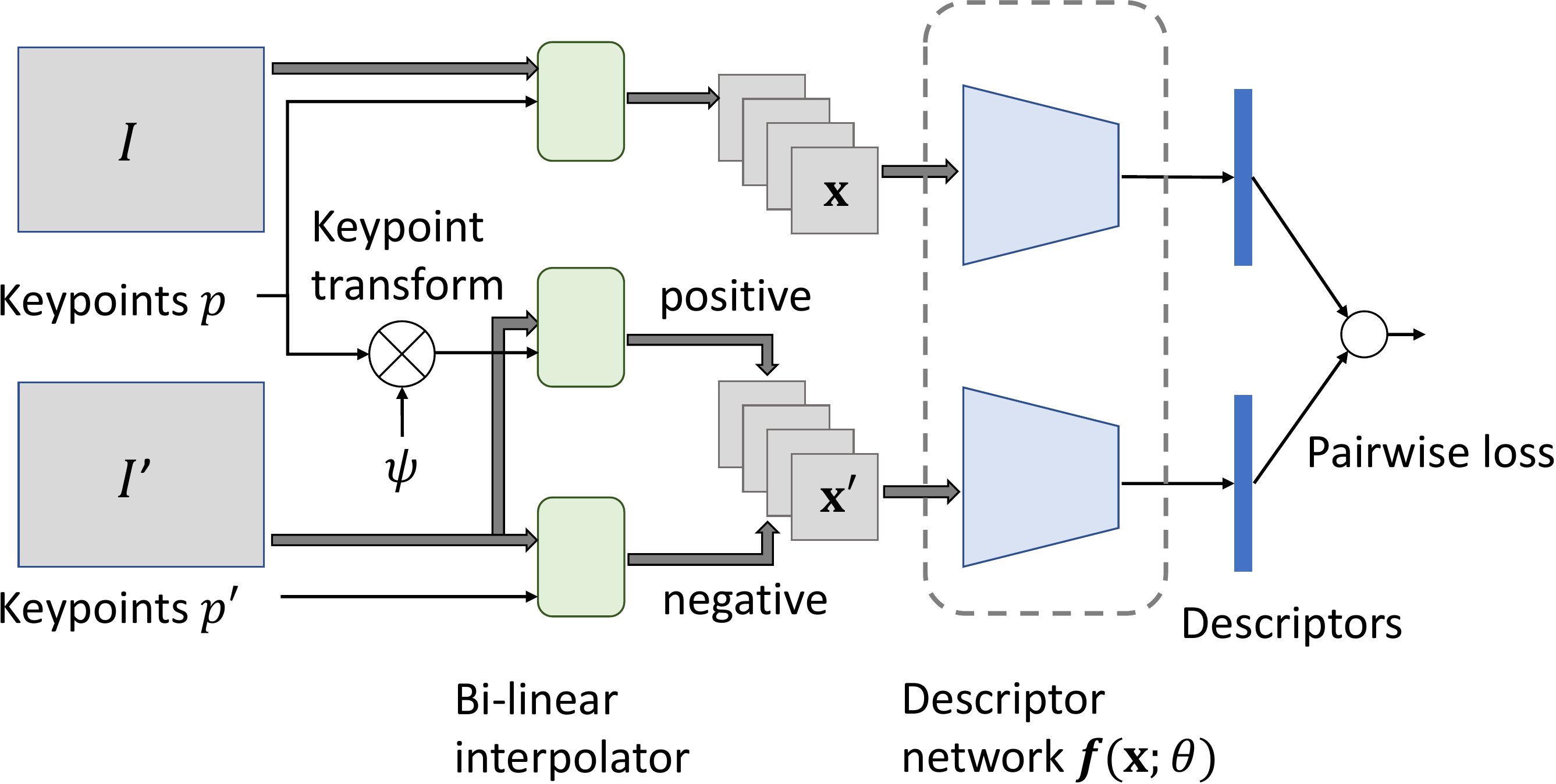}
\end{center}
\vspace{-2mm}
\caption{Overview of the proposed joint model.}
\label{fig:pipeline}
\end{figure}

In this paper, we assume that the warping model $\mathbf{w}$ is based on a 2D homography. Thus, we can handle
image pairs from a purely zooming or rotating camera or overlapping images of a planar
scene from one or more cameras.

We now describe the joint optimization for the homography case in more details. To generate patches from an image,
we extract multiple randomly sampled 2D keypoints from the image. Each keypoint
$\mathbf{p}$ has a position $(x,y)$, orientation $\phi$ and scale $s$, from which we can
resample a $n \times n$ square patch using bilinear interpolation. Mathematically, we write this as
$\mathbf{x}=\text{B}(\mathbf{p}; I)$.
To obtain corresponding patches, we transform the keypoint $\mathbf{p}$ in the image $I$ using the
homography to obtain the transformed keypoint $\mathbf{p}^\prime$ in $I^\prime$ and then resample the patch
associated with $\mathbf{p}^\prime$.
Here $\mathbf{w}_k$ is the same warping function as $\mathbf{w}$, but $\mathbf{w}_k$ transforms keypoints
instead of patches, which will be detailed in Section~\ref{sec:warp-keypoint}. Mathematically, we have,
\begin{equation}
\mathbf{x}\!=\!\text{B}(\mathbf{p}; I), \; \mathbf{w}(\mathbf{x}; \psi)\!=\!\text{B}(\mathbf{p}^\prime; I^\prime)\!=\! \text{B}(\mathbf{w}_k(\mathbf{p}; \psi); I^\prime).
\label{eqbilinear}
\end{equation}

Bilinear interpolation is differentiable with respect to the homography parameters. When we substitute
$\mathbf{w}(\mathbf{x}; \psi)$ from Eq.~\ref{eqbilinear} into Eqs.~\ref{eq:loss-sup} and \ref{eq:argmin-selfsup}, the
new training loss remains differentiable with respect to $\theta$ and $\psi$. Thus, we can use
backpropagation to compute the derivatives. Figure~\ref{fig:pipeline} illustrates our model.
Since the parameter vector $\psi$ gets iteratively updated, we compute the positive pairs
using the updated homography from scratch and regenerate the positive
training set on-the-fly in each iteration. However, the set of negative pairs remains fixed throughout.
The neural network architecture, implementation details of keypoint warping and the homography
parametrization are discussed in Section~\ref{sec:details}.

\subsection{Partially shared pseudo siamese network}
\label{sec:pseudo-siamese}

\begin{figure}[t]
\begin{center}
\includegraphics[width=0.85\linewidth]{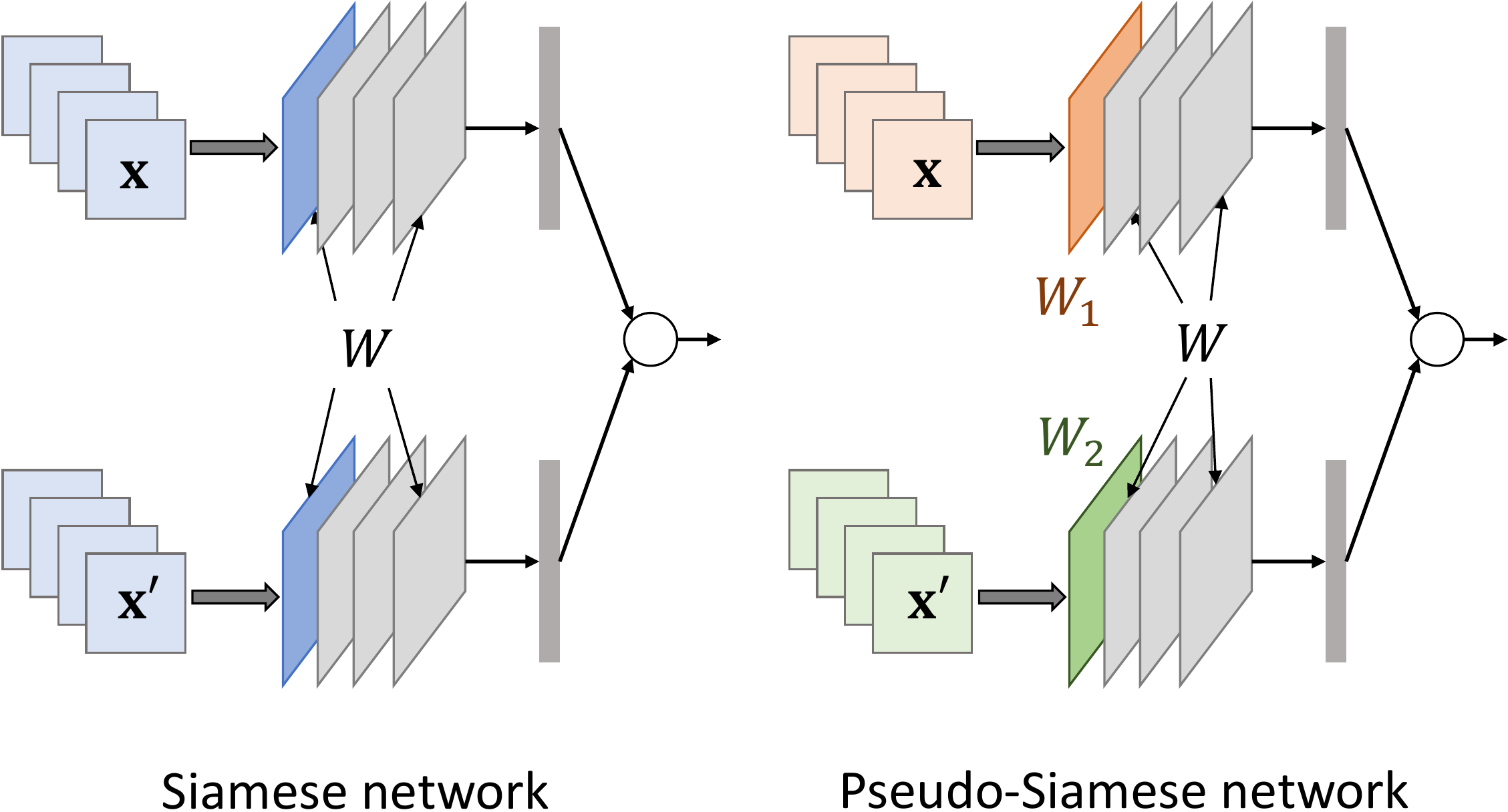}
\end{center}
\caption{We train standard siamese networks and a special form of pseudo-siamese networks with unshared weights in the first layer
to adapt to different image modalities.} 
\label{fig:pseudo-siamese}
\end{figure}

Since siamese networks uses shared weights for both input pairs, they may perform poorly
when the input image pairs have different modalities, \eg RGB and NIR, since one set of parameters
may not capture the statistics of both modalities.
Similar ideas for unsharing network weights have
been investigated in domain adaptation and transfer learning~\cite{Rozantsev2018,Ma2012}.
In \emph{pseudo-siamese} networks, both networks have the same structure but
have different copies of the weights. They have been used for matching different modalities~\cite{Aguilera2016,Mou2017}.
but do not provide a significant gain when both inputs have the same modality~\cite{Zagoruyko2015}.
With twice the parameters, pseudo-siamese networks tend to perform worse than siamese networks
when training data is limited.

We explore a special case of pseudo-siameses networks where the weights of only the first
layer of both networks are different \ie the first layer is unshared, but the remaining parameters of
the network are shared (see Figure~\ref{fig:pseudo-siamese}). This allows the first layer to adapt to the
different modalities but only adds a small number of parameters to the model. This is important for us since
the training sets are quite small.

\section{Implementation details}
\label{sec:details}

In this section, we present important details for implementing our approach
and describe our CNN architecture.

\subsection{Homography parameterization for SGD}
\label{sec:homgraphy-rep}
\begin{figure}[t]
\begin{center}
\includegraphics[width=0.49\linewidth]{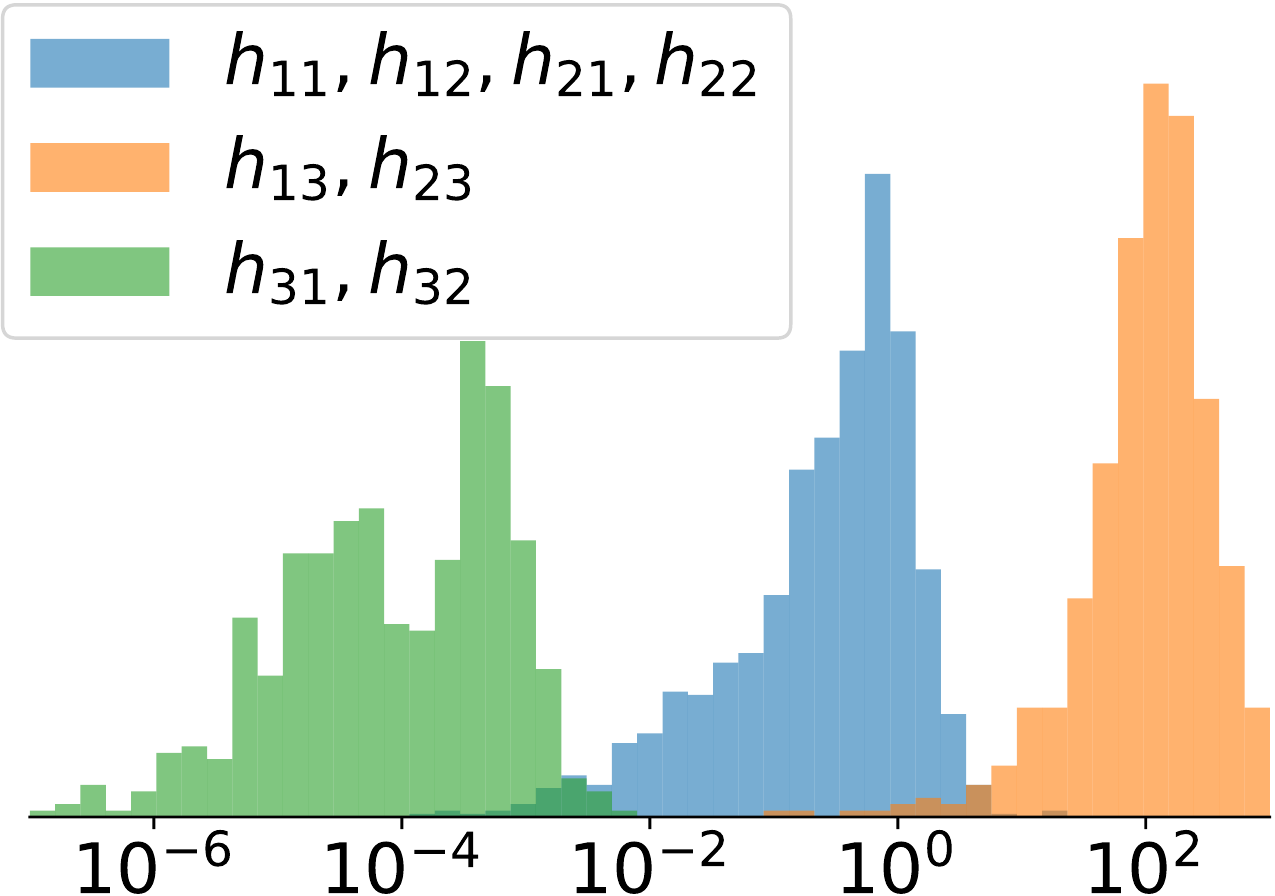}
\hfill
\includegraphics[width=0.49\linewidth]{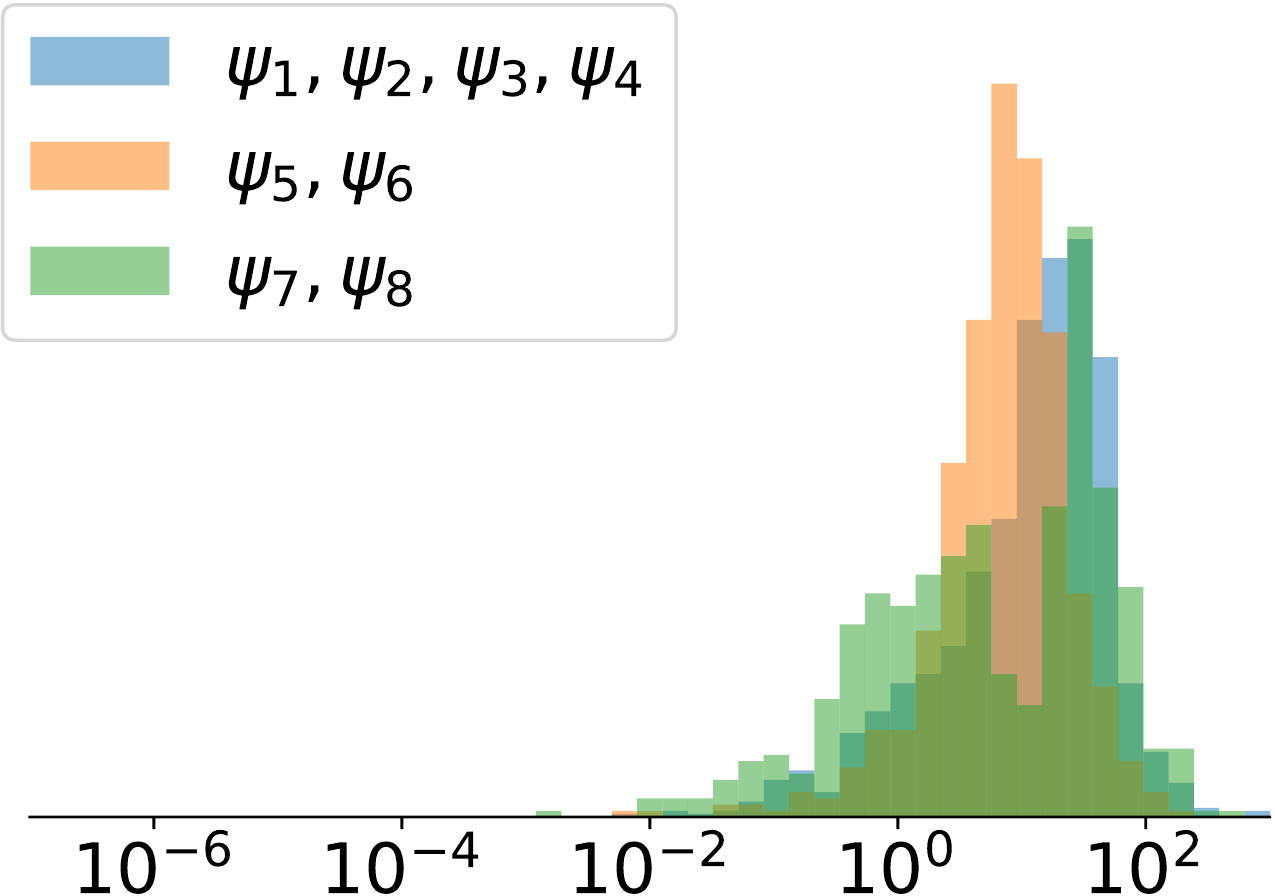}
\end{center}
\caption{Homography parameter's distributions of dataset~\cite{Balntas2017} in log-scale, left is before normalization in $H$, and right is normalization in $\psi$ with $\alpha=64$.}
\label{fig:h-distribution}
\end{figure}

Unlike second-order methods \eg Gauss-Newton, first-order methods like SGD
are not curvature-aware and have trouble optimizing ill-conditioned problems
when different dimensions of the parameter space differ a lot in scale.
The usual 8-DOF parameterization for the homography $H$ has
eight parameters often have very different scales.
\begin{equation}
H=\begin{pmatrix}
h_{11}& h_{12}& h_{13}\\
h_{21}& h_{22}& h_{23}\\
h_{31}& h_{32}& 1\\
\end{pmatrix}.
\end{equation}
Figure~\ref{fig:h-distribution} shows the scale variation of the eight
parameters for some homographies in the HPatches dataset~\cite{Balntas2017}.
Therefore, we choose a well-conditioned homography parametrization that is more
suitable for use with SGD. Based on \emph{dimensional analysis}~\cite{Bridgman1922}
in physics, we check the \textit{unit} of $H$ the homography mapping although, there are no physical
units.
\begin{align}
x^\prime\!=\!\frac{h_{11}x+h_{12}y+h_{13}}{h_{31}x+h_{32}y+1},
y^\prime\!=\!\frac{h_{21}x+h_{22}y+h_{23}}{h_{31}x+h_{32}y+1}.
\label{eq:homography-trans}
\end{align}
Here, $x$, $y$, $x^\prime$ and $y^\prime$ are in pixels. So to reach dimensional homogeneity, $h_{11}, h_{12}, h_{21}, h_{22}$ must have \textit{unit} $1$, $h_{13}, h_{23}$ must have $\text{pixel}$ \textit{unit}, and $h_{31}, h_{32}$ must have $\text{pixel}^{-1}$ \textit{unit}. Thus, we normalize $H$ using the image dimension $w \times h$ and this gives the following 8-dimensional vector for $\psi$.
\begin{equation}
\alpha\big[h_{11}-1, h_{12}, h_{21}, h_{22}-1, \frac{h_{13}}{w}, \frac{h_{23}}{h},h_{31}w, h_{32}h \big]
\end{equation}
Here, $\alpha$ is a scale hyperparameter that we set to $64$.
This parametrization is scale normalized (see Figure~\ref{fig:h-distribution}) and
conveniently maps $\psi=0$ to the identity homography.

\subsection{Keypoint transformation under homography}
\label{sec:warp-keypoint}

The position $(x^\prime,y^\prime)$ of a warped keypoint $\mathbf{p}^\prime$ is given by
Eq.~\ref{eq:homography-trans}. To find the orientation $\phi^\prime$ and scale $s^\prime$
of the point $\mathbf{p}^\prime$, we treat
$\mathbf{p}$ as a 2D vector $[v_0, v_1]^T$ with orientation $\phi$ and length $s$
with origin $(x,y)$ in image $I$. Then, we have
\begin{align}
&v_0 \propto s\cos\phi, &v_1 \propto s\sin\phi\label{eq:kvec}
\end{align}
Warping a keypoint is the same as finding the warped vector $(v_0^\prime, v_1^\prime)^T$ in $I^\prime$.
Assuming the vectors are short, we can use finite differences on Eq.~\ref{eq:homography-trans}
to compute the values of
\begin{align}
v_0^\prime \approx \frac{h_{11}v_0+h_{12}v_1}{h_{31}x+h_{32}y+1},
v_1^\prime \approx \frac{h_{21}v_0+h_{22}v_1}{h_{31}x+h_{32}y+1}
\label{eq:diffkvec}
\end{align}
The orientation and scale of $\mathbf{p}^\prime$ can be computed by
substituting values of $v_0^\prime$ and $v_1^\prime$ from Eq.~\ref{eq:diffkvec}
into these equations.
\begin{align}
\phi^\prime = \arctan\Big(\frac{v_1^\prime}{v_0^\prime}\Big),
s^\prime =\sqrt{(v_0^\prime)^2 + (v_1^\prime)^2}
\end{align}
%
%
%

\subsection{CNN training}
\label{sec:cnn}

To obtain image patches to train our model, we select keypoint sets $\{\mathbf{p}\}$ and $\{\mathbf{p}^\prime\}$ with
sufficient randomization to reduce the effect of overfitting. We sample approximately 4000 keypoints in each image.
The positions of these keypoints are sampled from areas with sufficient contrast by selecting areas were the local
gradient magnitude exceeds 0.05\footnote{image intensities have mean 0 and standard deviation 1} and selecting pixels randomly from
a uniform distribution. The orientation is sampled uniformly from the range $[0, 2\pi]$ and the scale values are sampled
such that $\log_2 s$ is distributed uniformly in the interval $[0, 4]$.
We also check that the warped keypoint $\mathbf{p}^\prime$ lies inside the boundary of image $I^\prime$.
Although the cropped patches extracted from $\{\mathbf{p}\}$ and $\{\mathbf{p}^\prime\}$ spatially overlap,
selecting the orientation and scales randomly provides effective data augmentation.

We use a shallow network architecture: {\tt Conv(16,32,\\5,1)-Tanh-MaxPool(12,-,2,2)-Conv(6,64,3,\\1)-Tanh-FC(256)}.
The parameters of each layer are denoted
by \textsc{(Input,Channel,Kernel,Stride)}. The patch size is $16 \times 16$ pixels and the descriptor dimension is $d\!=\!256$.
Our design is based on prior work~\cite{Balntas2016tfeat,Simo2015} but other architectures could be used as well.
Since we have relatively fewer and small patches to train our model, we favor architectures with lower capacity to reduce the risk of overfitting.
The optimization is done using SGD with batch size of 64 and
momentum of 0.9, with an initial learning rate of $10^{-4}$ which is temporally annealed.  

To address the issue of potentially large initial misalignments, the training is done in a coarse-to-fine fashion on an image pyramid, where the coarsest level has about $80$ pixels on the longer side. The joint alignment and network training starts at coarsest pyramid level after which the estimated parameter $\psi^*$ is used to initialize $\psi$ at the next level. The network parameters $\theta^*$ are discarded as descriptors learned at coarser scales are not suitable for patches at finer scales. The parameters $\psi^*$ and $\theta^*$ from the finest level are saved.

\section{Experiments}

We first evaluate our method on matching color images from HPatches~\cite{Balntas2017}
to compare with existing methods.
We then test our method for aligning aerial RGB and NIR images that were captured using an off the shelf drone.
Our implementation is based on Tensorflow~\cite{Tensorflow2015}.
All experiments were done using a NVIDIA GTX 1080Ti GPU, and aligning each image pair takes about 90 seconds.
On both datasets, we quantitatively evaluate ours descriptor to SIFT~\cite{Lowe2004} and DAISY~\cite{Tola2010}. We
also compare to four learned approaches -- DeepCompare~\cite{Zagoruyko2015} with one (-s) and two stream (-s2s) siamese networks, DeepDesc~\cite{Simo2015} and DeepPatchMatch~\cite{Kumar2016} which are currently amongst the top ranked methods on HPatches and were trained on the MVSCorr dataset~\cite{Brown2011}.
We use the \emph{mean average precision (mAP)} metric to evaluate descriptor performance.
To evaluate an image pair ($I$,$I^\prime$), we first extract many pairs of corresponding patches
from them. We then extract descriptors from the patches. Then, for each descriptor extracted from $I$, we
independently compute the exact nearest neighbor within the set of descriptors extracted from $I^\prime$, in the $L2$
distance sense. The average precision (AP) is the fraction of times the correct nearest neighbor was found.
The mean AP (mAP) is the average AP across multiple image pairs.

\subsection{Evaluation on the HPatches dataset}
HPatches~\cite{Balntas2017} contains image pairs rated at~\textsc{Easy},~\textsc{Hard} and \textsc{Tough} difficulty levels for 116 scenes where 59 scenes have viewpoint variations and the rest vary in illumination.
The evaluation protocol in HPatches involves calculating mAP on pre-extracted pairs of grayscale patches.
We implement this protocol but using our own patches so that descriptors
computed from RGB patches can also be evaluated. To generate the evaluation patches,
we detect up to 2000 SIFT keypoints~\cite{Lowe2004} in the first image, warp the underlying patches by the
ground truth homographies and sampled the corresponding patches in the second image.

\begin{table*}[!t]
\centering
\begin{tabular}{llccccccc}
\hline
\multirow{2}{*}{Colorspace} & \multirow{2}{*}{Method} & \multicolumn{3}{c}{Viewpoint} & \multicolumn{3}{c}{Illumination} & \multirow{2}{*}{Average} \\
& & \textsc{Easy} & \textsc{Hard} & \textsc{Tough} & \textsc{Easy} & \textsc{Hard} & \textsc{Tough} &\\
\hline
\multirow{8}{*}{Grayscale}&SIFT~\cite{Lowe2004} & 0.485 & 0.297 & 0.192 & 0.549 & 0.431 & 0.372 & 0.388\\
&DAISY~\cite{Tola2010} & 0.683 & 0.491 & 0.336 & 0.579 & 0.416 & 0.327 & 0.472\\
&DeepCompare-s~\cite{Zagoruyko2015} & 0.781 &  0.542& 0.361 & 0.779 & 0.632 & 0.532 & 0.605\\
&DeepCompare-s2s~\cite{Zagoruyko2015} & \bf{0.803} & \bf{0.581} & \bf{0.396} & 0.776 & 0.631 & 0.542 & \bf{0.622}\\
&DeepDesc~\cite{Simo2015} & 0.781& 0.554 & 0.360 & \bf{0.792} & \bf{0.655} & 0.547 & 0.615\\
&DeepPatchMatch~\cite{Kumar2016} & 0.770 & 0.541 & 0.359 & 0.760 & 0.610 & 0.509 & 0.592\\
&\bf{Ours-s-Grayscale} & 0.700 & 0.426 & 0.309 & 0.790 & 0.644 & \bf{0.549} & 0.570\\
&\bf{Ours-ps-Grayscale} & 0.729 & 0.446& 0.290 & 0.760 & 0.624 & 0.520 & 0.562\\
\hline
\multirow{2}{*}{RGB}&\bf{Ours-s-RGB} & 0.763 & 0.525 & \bf{0.397} & \bf{0.813} & \bf{0.679} & \bf{0.607} & 0.631\\
&\bf{Ours-ps-RGB} & 0.781 & 0.563 & 0.388 & 0.808 & 0.664 & 0.592 & \bf{0.633}\\
\hline
\end{tabular}
\caption{HPatches evaluation: mAP for six baselines and variants of our method. For the six groups (columns), the best grayscale method is in bold. When the RGB method has a higher mAP than the best grayscale method, it is also in bold.}
\label{table:results_hpatch}
\end{table*}

\subsubsection{Descriptor performance evaluation}

We evaluate four variants of our method for grayscale and RGB patches combined with
the siamese network (\textbf{Ours-s}) and the pseudo siamese network (\textbf{Ours-ps}) respectively.
The descriptor mAP evaluation is summarized in Table~\ref{table:results_hpatch}. All six
baselines were based on grayscale patches. The coarse alignment needed by our method for
initialization was simulated by adding random 2D translational perturbations to the true homography,
where the shift is equal to 5\% of the image size (see later for results with larger shifts).

In this evaluation, \textbf{Ours-s-Grayscale} is ranked lower than pretrained CNN but the gap in mAP is small --
0.57 (ours) \vs 0.59--0.62~\cite{Zagoruyko2015,Simo2015,Kumar2016}.
This is expected, since those CNNs were trained offline on a massive patch dataset whereas
in our case, the network is optimized from scratch on each image pair.
SIFT and DAISY have much lower mAP of 0.39 and 0.47 respectively.
However, our RGB networks have the highest overall accuracy. The
pseudo siamese (mAP = 0.633) and the siamese network (mAP = 0.631)
both had similar accuracies.
The siamese variant \textbf{Ours-s-RGB} has the highest mAP in four out of the six subgroups.
This evaluation shows that the descriptors learnt from scratch by our method are representative
even though we do not expect them to generalize to new images.

\subsubsection{Analyzing robustness to initial alignment error}

\begin{figure}[t]
\begin{center}
\includegraphics[width=0.49\linewidth]{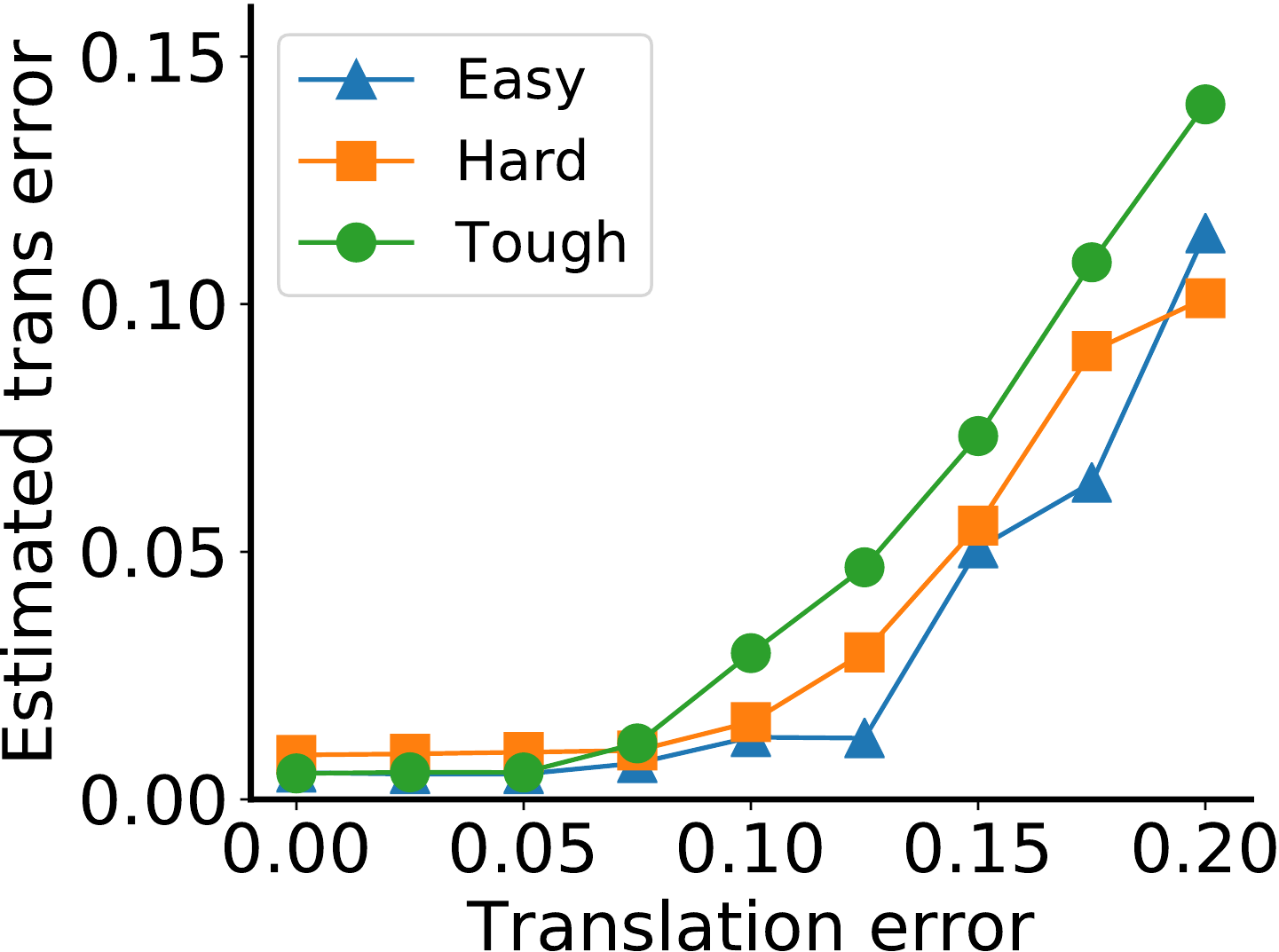}
\hfill
\includegraphics[width=0.49\linewidth]{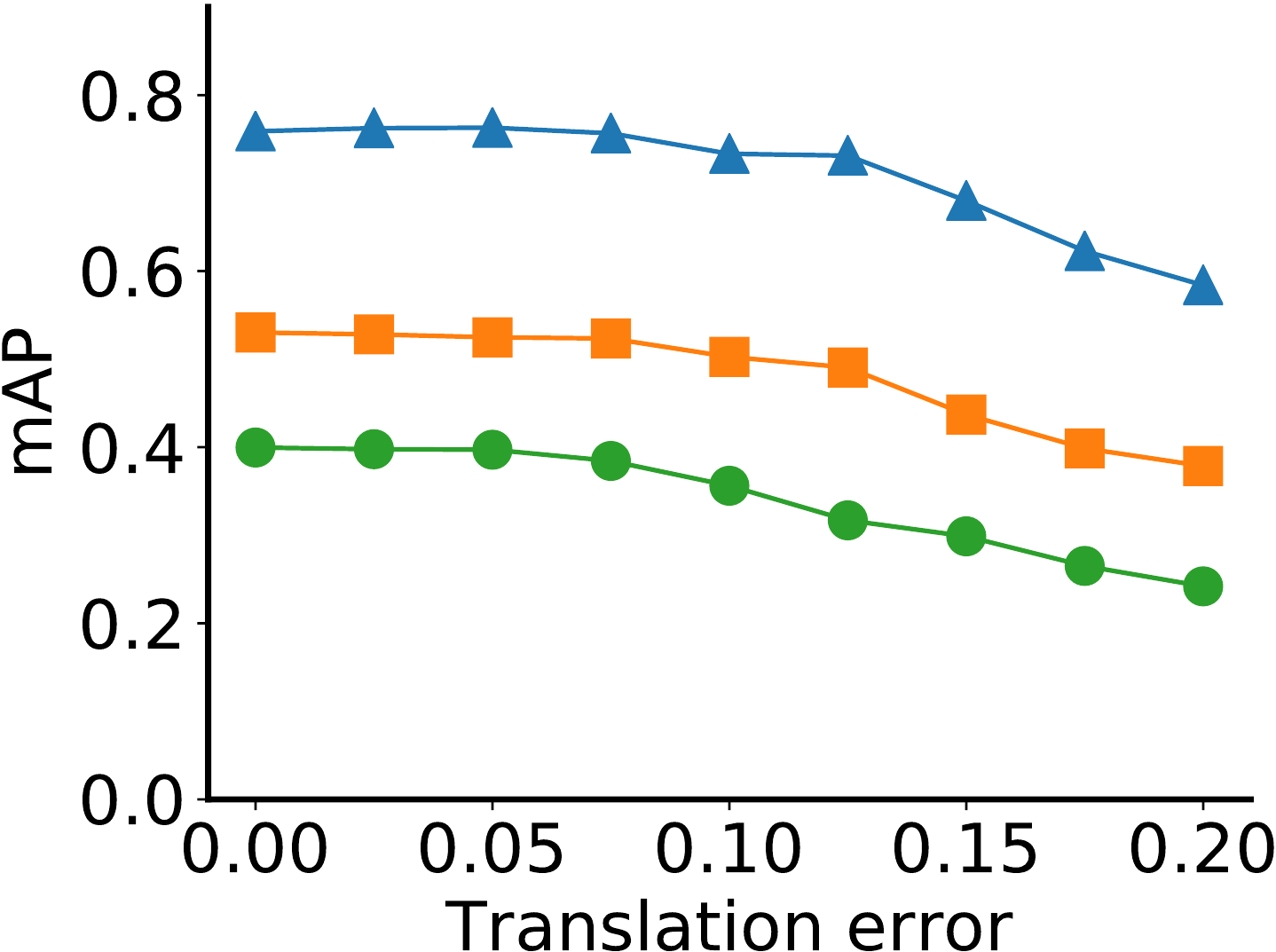}
\includegraphics[width=0.49\linewidth]{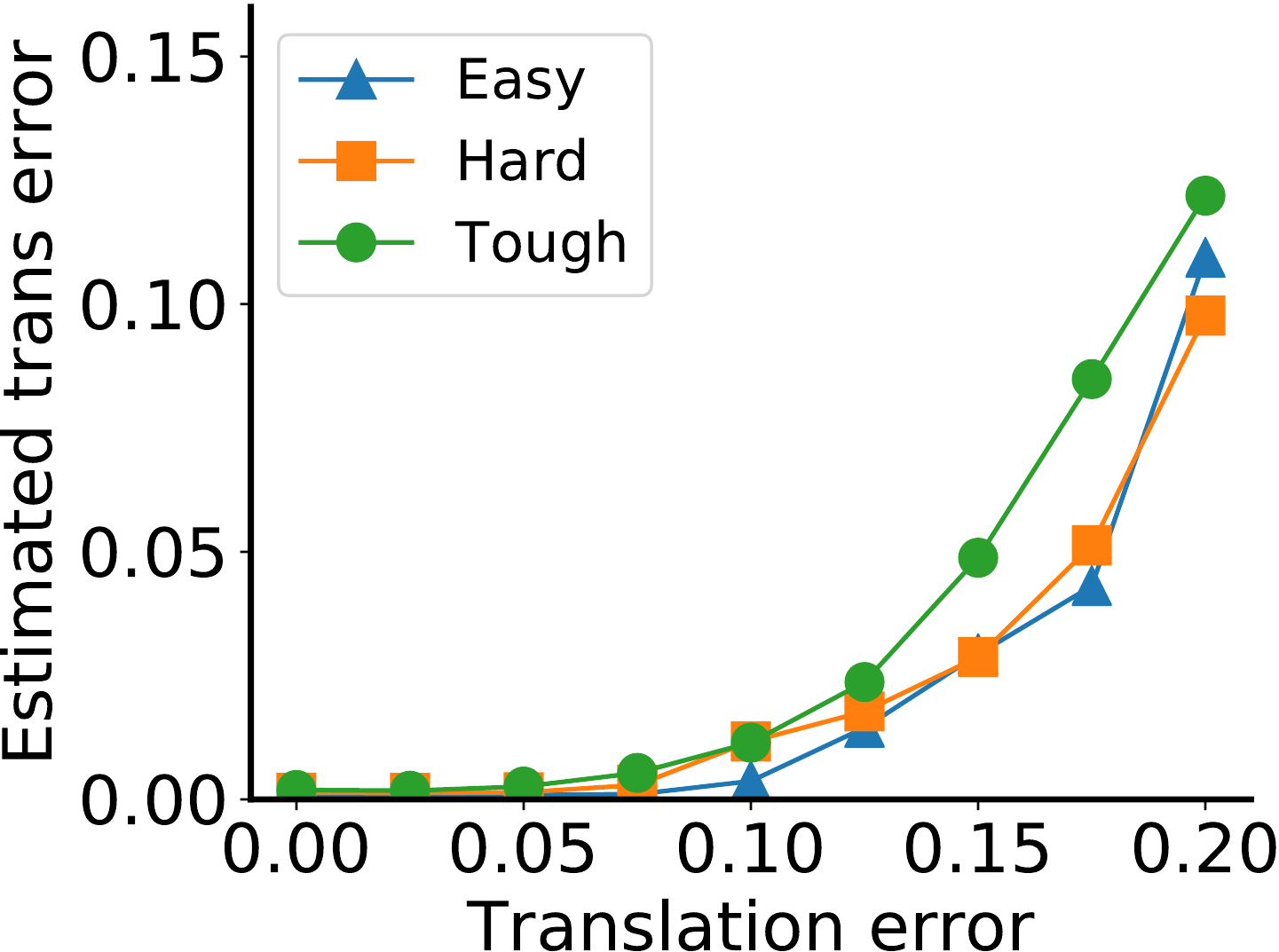}
\hfill
\includegraphics[width=0.49\linewidth]{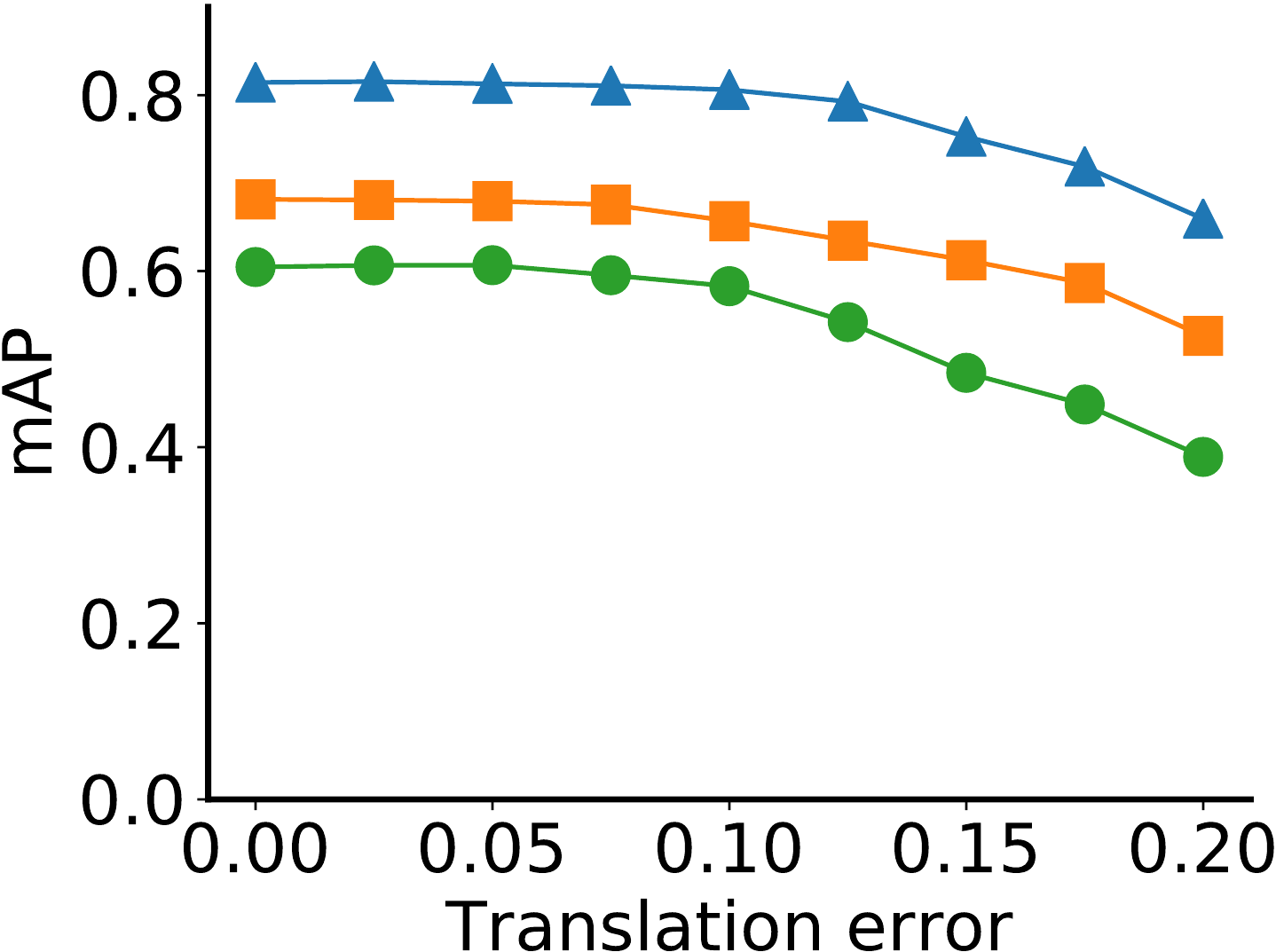}
\end{center}
\caption{The average error of the estimated homography and mAP score of our method on~\textsc{Easy},~\textsc{Hard} and \textsc{Tough} pairs for a range of initial translation errors (expressed as a fraction of the image size). Upper and lower rows are for "v" and "i" groups respectively in HPatches~\cite{Balntas2017} (see text).}
\label{fig:h-err-hpatch}
\end{figure}

We also analyze the robustness of our method to increasing amounts of error in the initial alignment.
Figure~\ref{fig:h-err-hpatch} shows the results of these experiments. The plots show both the homography
estimation error and the mAP scores for models trained on the six subgroups starting with a different
amount of translational perturbation.
The homography error is equal to the average warping error of the true feature matches under the estimated homography,
after normalizing the error by the image size.
The plots show that the homography error is very low for small perturbations (up to 7.5\%) which indicates
accurate alignment. The alignment error does increase with higher perturbation. However, it is worth noting
that here, we adhere to the standard descriptor evaluation protocol and avoid RANSAC and geometric constraints
to \textit{robustly} estimate the homography which would have easily yielded more accurate alignments.

For perturbations between 0.0--0.1, the mAP curves are mostly flat \ie the mAP drop is fairly small. The mAP drops by an amount
between 0.01 and 0.04 for the six subgroups.
In these plots, a perturbation of 0.1 is equal to a 2D shift of 10\% of the image size. As the perturbation increases, the
accuracy of our method drops gradually. At 0.2, the mAP is still reasonably high for the~\textsc{Easy} and \textsc{Hard}
groups whereas performance drops more for the \textsc{Tough} group.

\subsection{Evaluation on RGB and NIR images}


Next, we report our evaluation of RGB--NIR image alignment which has useful applications in precision agriculture (see example
in Figure~\ref{fig:idea}). The main difficulty here is due to frequent gradient reversal and the
lack of correlation in the visible spectrum and the NIR band (770--810nm).
We collected 50 RGB-NIR image pairs and manually annotated sparse correspondences in them and then recovered ground truth homographies.
We split these into two sets -- a training set of 40 pairs and a test set of 10 pairs. We evaluate the descriptor performance
and homography estimation accuracy on the training set, which we present next.

\subsubsection{Descriptor performance evaluation}\label{sec:drone_desc}

We calculate the mAP of our descriptors on the 40 training pairs using the same protocol used for HPatches.
In each case, an identity transformation was used for initialization.
We compared our method with the six baselines where the color patches were converted to grayscale.
The first column of Table.~\ref{table:results_drone} summarizes the descriptor performance evaluation for this
case. Our method performs significantly outperforms all existing methods. The existing learned methods were not
trained for these modalities and are not applicable when datasets with ground truth correspondences are unavailable.
In this case, the pseudo-siamese network (mAP = 0.603) is much more accurate than the siamese network (mAP = 0.404)
for the reason discussed earlier.

\begin{table}[!t]
\centering
\begin{tabular}{lcc}
\hline
\multirow{2}{*}{Method} & \multicolumn{2}{c}{RGB vs. NIR} \\ & Train Set(40) & Test Set(10) \\
\hline
SIFT & 0.098 & 0.085 \\
DAISY & 0.030  & 0.040 \\
DeepCompare-s & 0.068 & 0.071\\
DeepCompare-s2s  & 0.101   & 0.098\\
DeepDesc & 0.070 & 0.066\\
DeepPatchMatch & 0.089 & 0.080\\
\bf{Ours-s}  & 0.404 & --\\
\bf{Ours-ps}  & \bf{0.603} & --\\
\bf{Automatic-s (supervised)}  & -- & 0.475 \\
\bf{Automatic-ps (supervised)} & -- & \bf{0.556} \\
\hline
\end{tabular}
\caption{Evaluation of RGB--NIR image alignment.
The first column shows the mAP on the 40 training pairs where our method clearly outperforms all baselines.
The second column shows the mAP on the 10 test pairs. Those descriptors were trained in supervised fashion
on a training set generated using our automatic weakly supervised method.}
\label{table:results_drone}
\end{table}

\subsubsection{Analyzing robustness to initial alignment error}

\begin{figure}[t]
\begin{center}
\includegraphics[width=0.47\linewidth]{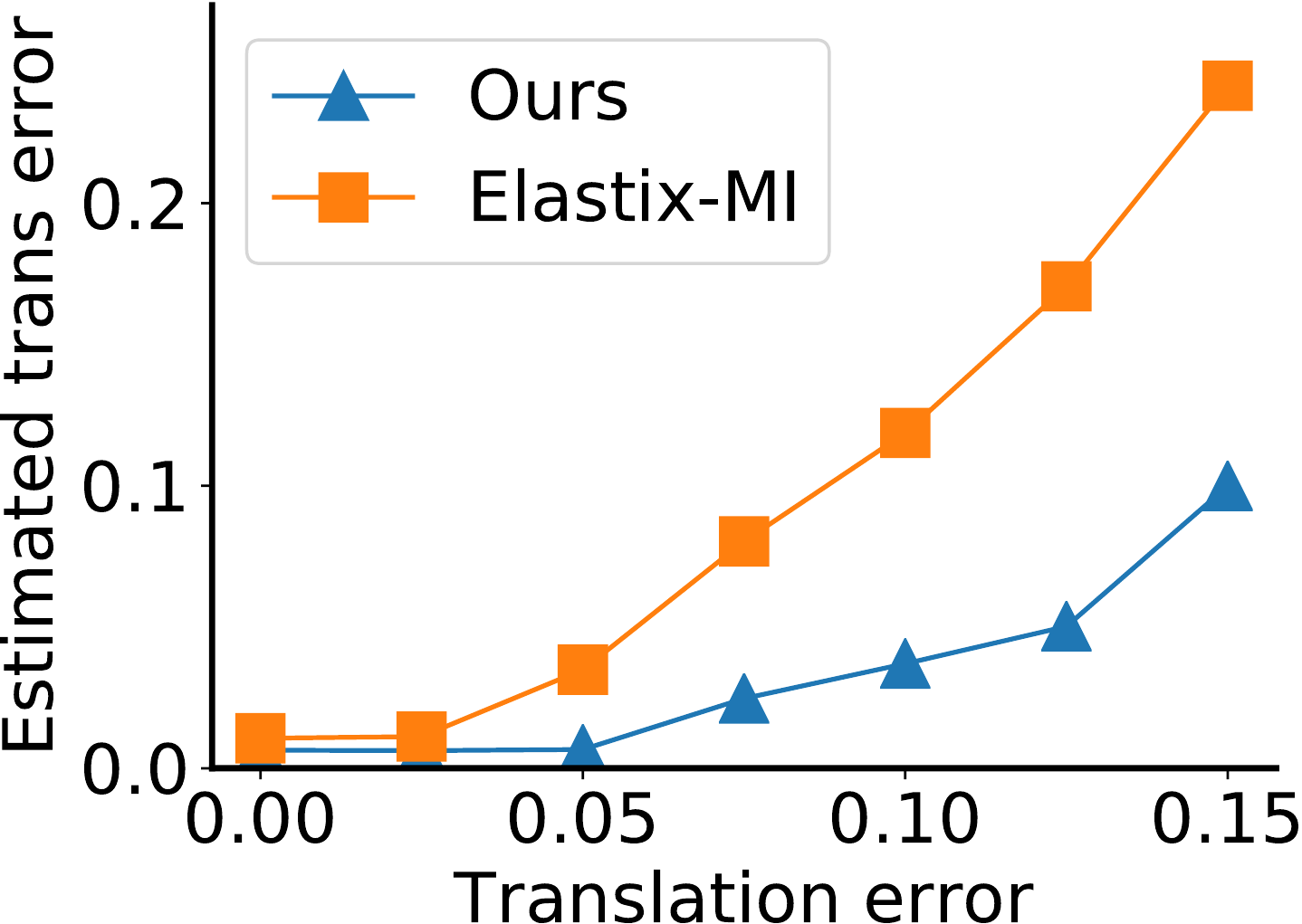}
\hfill
\includegraphics[width=0.47\linewidth]{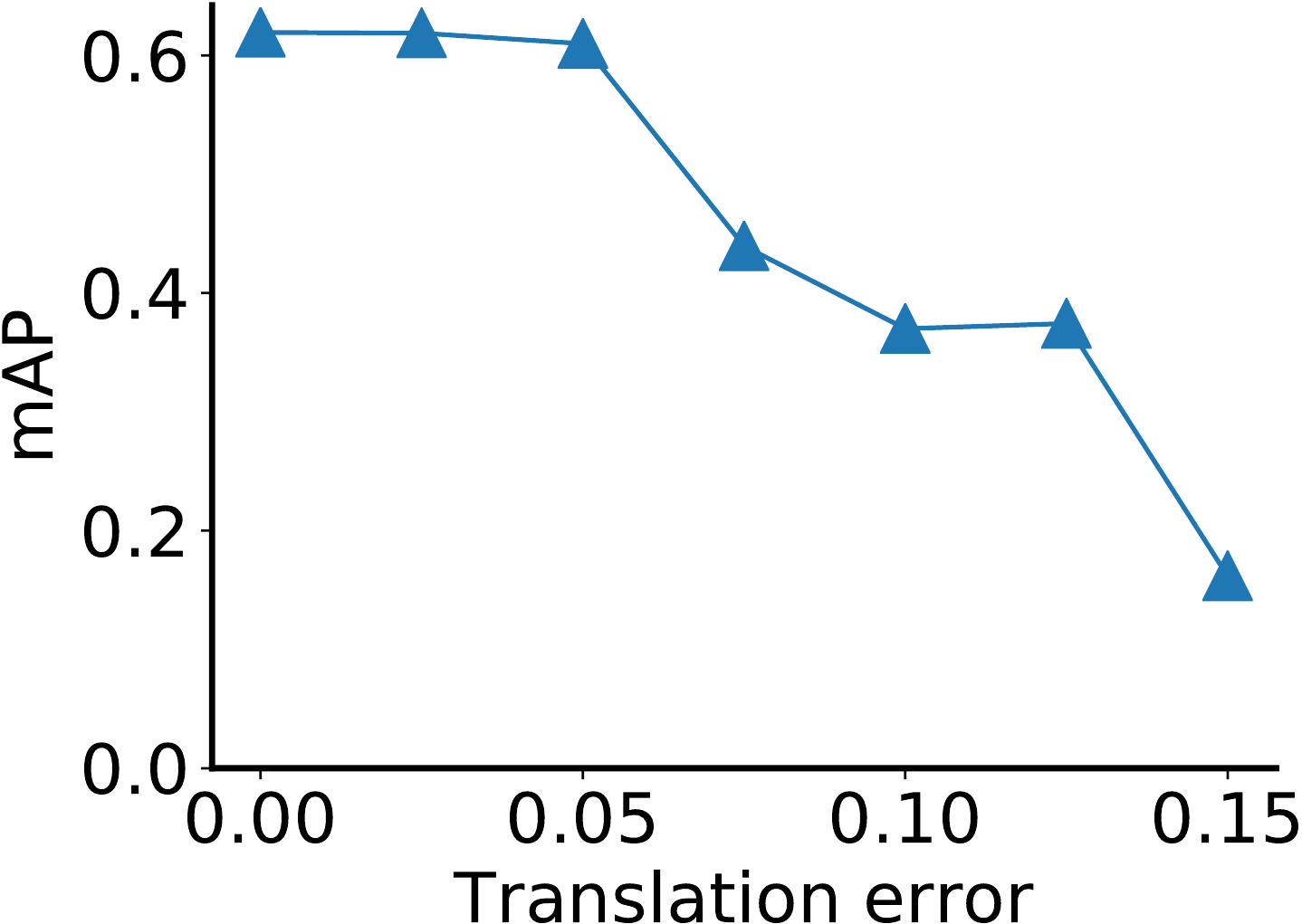}
\end{center}
\caption{Homography errors (alignment accuracy) and mAP (descriptor performance) for our method and {\tt Elastix}~\cite{klein2010elastix} which uses mutual information for image registration when the error in the initial alignment is varied. Our method can handle much more initial misalignment.}
\label{fig:h-err-drone}
\end{figure}

Once again, we evaluate our method by adding translational shifts to the ground truth alignment and simulate initializations of increasing difficulty.
Since mutual information is sometimes used to align images of different modalities~\cite{Viola1997}, we compare our pseudo-siamese network to an
advanced mutual information based image registration method -- {\tt Elastix}~\cite{klein2010elastix} which uses Mattes' mutual information metric~\cite{mattes2003}, with the same coarse to fine pyramid resolution as our method.
The results are shown in Figure~\ref{fig:h-err-drone}. With increasing translational perturbation, the mAP decreases gradually and the homography error increases as expected. However, our method is consistently more reliable than {\tt Elastix}. In particular, our method has near-zero homography error (\ie all pairs in training set are accurately aligned ) up to a perturbation of 0.05 whereas {\tt Elastix} is accurate only up to a much smaller perturbation of 0.025.

\subsubsection{Automatic descriptor learning}

We now evaluate the effectiveness of our method for automatically building a dataset without human annotation.
We train a supervised descriptor on such a dataset and test the descriptor performance.
We run our method on 40 RGB/NIR image pairs. Using the homography obtained from each pair we
extract correspondences. Specifically, SIFT keypoints extracted from the RGB images are
warped by the estimated homography. We construct a set of 62K positive pairs in the training set.
The supervised descriptor is a siamese CNN architecture (see Section~\ref{sec:cnn}), and trained using
the same hyperparameters described earlier except that fewer training epochs were used.

This supervised descriptor network is evaluated using the same protocol as before but on the 10 test image pairs. The last two rows of Table.~\ref{table:results_drone} shows the mAP for this method. The mAP of 0.556 is comparable to that of our weakly supervised method and much higher than all the existing methods. These results were obtained without tuning the network's architecture or its hyperparameters and show that our weakly supervised method is feasible for automatically learning local descriptors from coarsely aligned images.


\section{Conclusion}

We proposed a new weakly supervised method to align two images related by an
unknown 2D homography when a coarse alignment is known. The key
idea here is to train a local descriptor siamese network from scratch while
jointly estimating the homography. We show how the descriptor parameters and
the geometric parameters can be updated jointly within a single optimization.
There are several avenues for future work. We will explore the idea of learning one
network from several image pairs while estimating a unique alignment for each pair
and also investigate extensions for making the method fully automatic. Finally,
being able to handle general image pairs which have significant parallax would make the
approach more widely applicable.

\paragraph{Acknowledgements.}
Part of the work was done while author Jing Dong was an intern at Microsoft Research.
This work was also supported in part by National Institute of Food and Agriculture, USDA, under 2014-67021-22556.

\nocite{*}

{\small
\bibliographystyle{ieee}
\bibliography{ref}
}

\newpage
\clearpage

\setcounter{page}{1}
\setcounter{section}{0}
\setcounter{figure}{0}
\setcounter{table}{0}
\setcounter{equation}{0}
\setcounter{footnote}{0}

\renewcommand{\thepage}{A\arabic{page}}
\renewcommand{\thesection}{\Alph{section}}
\renewcommand{\thefigure}{A\arabic{figure}}
\renewcommand{\thetable}{A\arabic{table}}
\renewcommand{\theequation}{A\arabic{equation}}

\makeatletter
\def\@thanks{}
\makeatother

%
%
\pagenumbering{gobble}
\title{Supplementary Material:\\ Learning to Align Images using Weak Geometric Supervision}

\maketitle

In the supplementary material, we include visualizations of intermediate results and our final feature matching and alignment results on several pairs from the HPatches dataset and our dataset of aerial RGB--NIR images.

\section{Visualization of keypoints used for training}

In the main paper, we mentioned that the keypoints were random selected from the two images and then used to resample image patches to train our siamese (or pseudo-siamese) network. Figure~1 shows an example of an image with 200 random sampled keypoints shown. It gives an indication about the underlying patches that were used for training.
We would like to point out that there are no keypoints in the homogeneous areas of the image since we ensure that keypoints are extracted from regions where the gradient magnitude exceeds a minimum threshold. This helps us to avoid sampling too many uninformative patches. Secondly, notice that there is significant overlap between nearby image patches. During training we sample 4000 keypoints in each image, so the amount of overlap is quite high. This might seem highly redundant. However, we ensure that all these patches are assigned a random scale and orientation and we resize the original patches to patches with $16 \times 16$ pixels.
The randomization acts as a form of data augmentation that is effective in our setting where the amount of training data is quite scarce.

\begin{figure}[t]
\begin{center}
\includegraphics[width=0.9\linewidth]{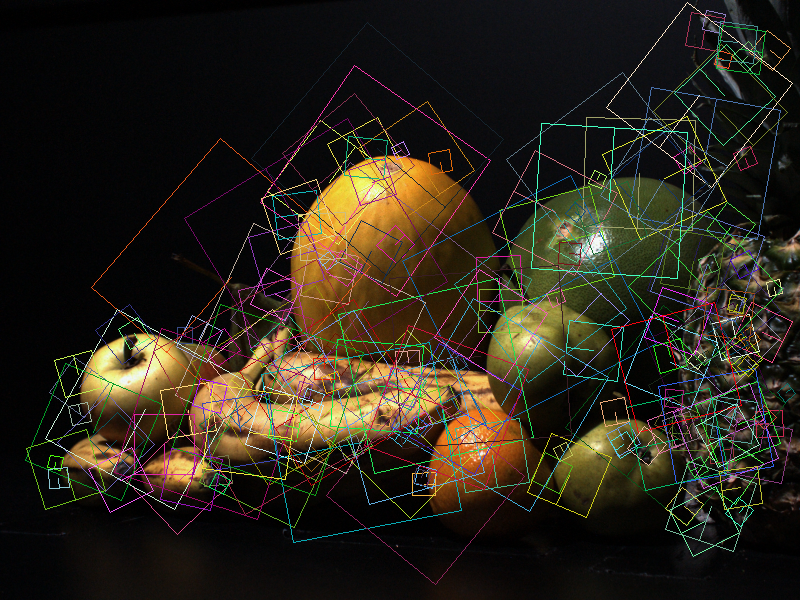}
\end{center}
\caption{An image from the HPatches dataset. We show 200 of the random keypoint frames extracted from this image, each of which has a random scale and orientation. The patches used to train our network are extracted from these keypoint frames.}
\vspace{-3mm}
\end{figure}

\section{Additional alignment and matching results}

In this section, we show additional feature descriptor matching results and image alignment results obtained using our method.
We generated the feature matching results as follows. We first extract a set of SIFT keypoints in each image. We then calculate feature descriptors for those keypoints using our network. Then for each descriptor in the first image, we search for the nearest neighbor amongst all descriptors in the second image, where the notion of neighbors is defined based on $L^2$ distance. We filter these candidate matches using a simple geometric verification step based on a 4-pt RANSAC method (using a minimal solver to compute homographies from four point--point matches).

We generated the results on the HPatches image pairs using our siamese network whereas for the RGB and NIR images, we used our partially shared pseudo-siamese network to compute feature descriptors. For all the qualitative results shown here, the initial alignment was obtained by adding a random 2D translational perturbation to the ground truth homographies where the amount of translation was equal to 15\% of the image size.

In Figures~2--7 we show six examples of results on HPatches pairs, whereas in Figures~8--11, we show four examples of RGB and NIR image alignment.
The feature matches in the RGB and NIR pairs are typically sparser compared to the matches in the RGB pairs. This is because the underlying SIFT keypoints are not very repeatable. However our learned descriptors seem to be reliable enough to compute an accurate alignment.

\begin{figure*}[t]
\begin{subfigure}[b]{1.0\textwidth}
\includegraphics[width=0.36\linewidth]{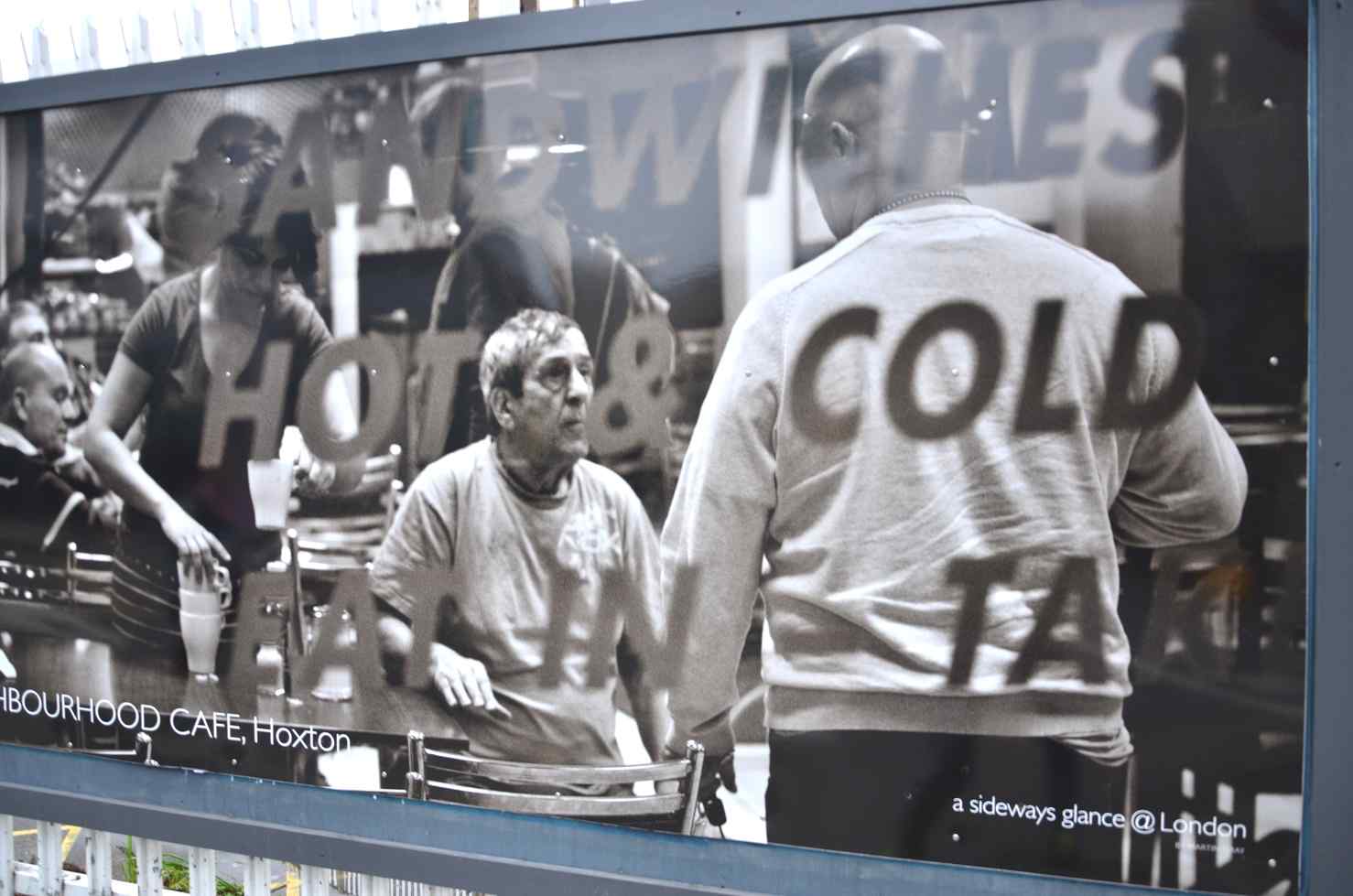}
\hfill
\includegraphics[width=0.36\linewidth]{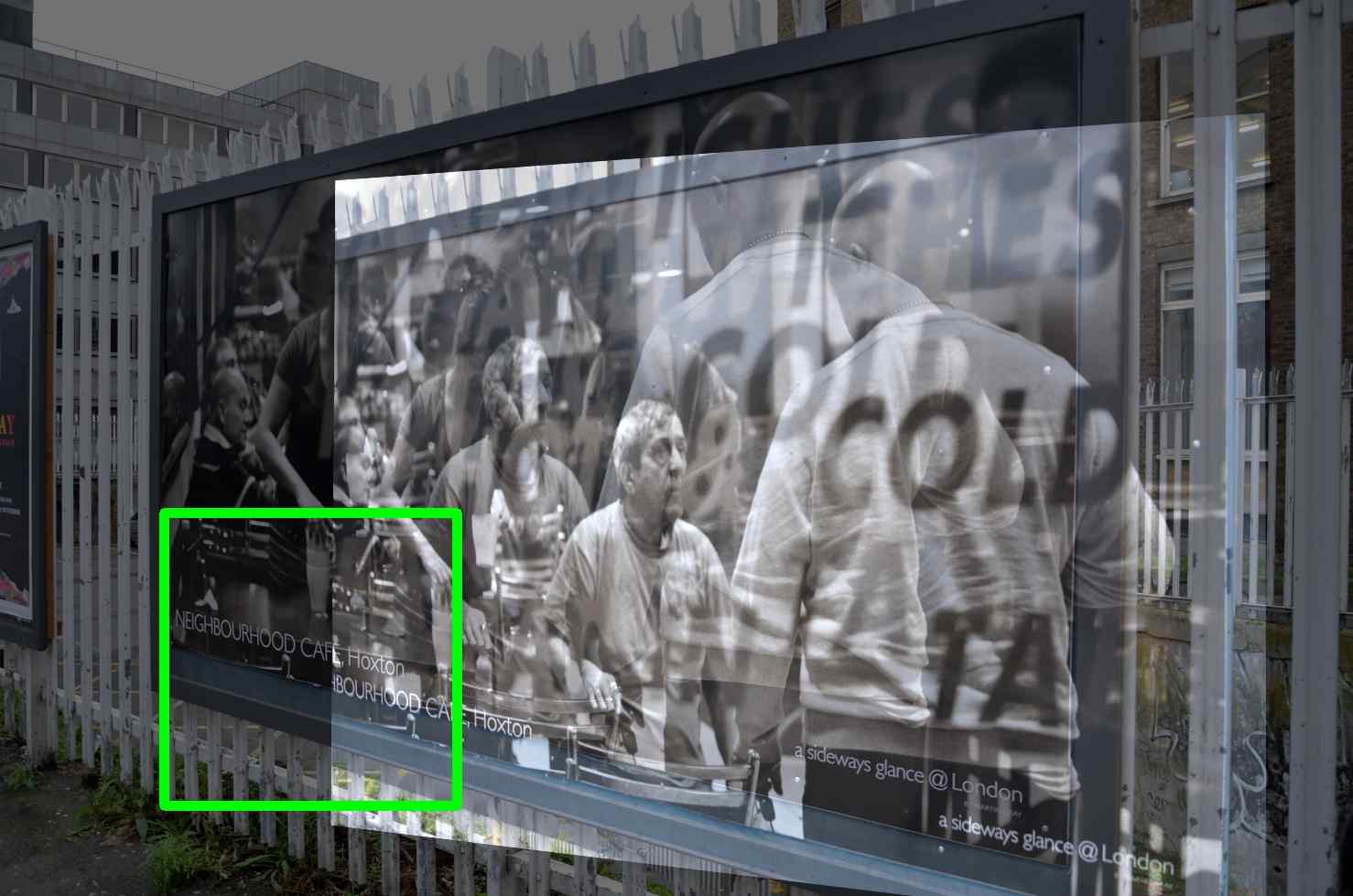}
\hfill
\includegraphics[width=0.238\linewidth]{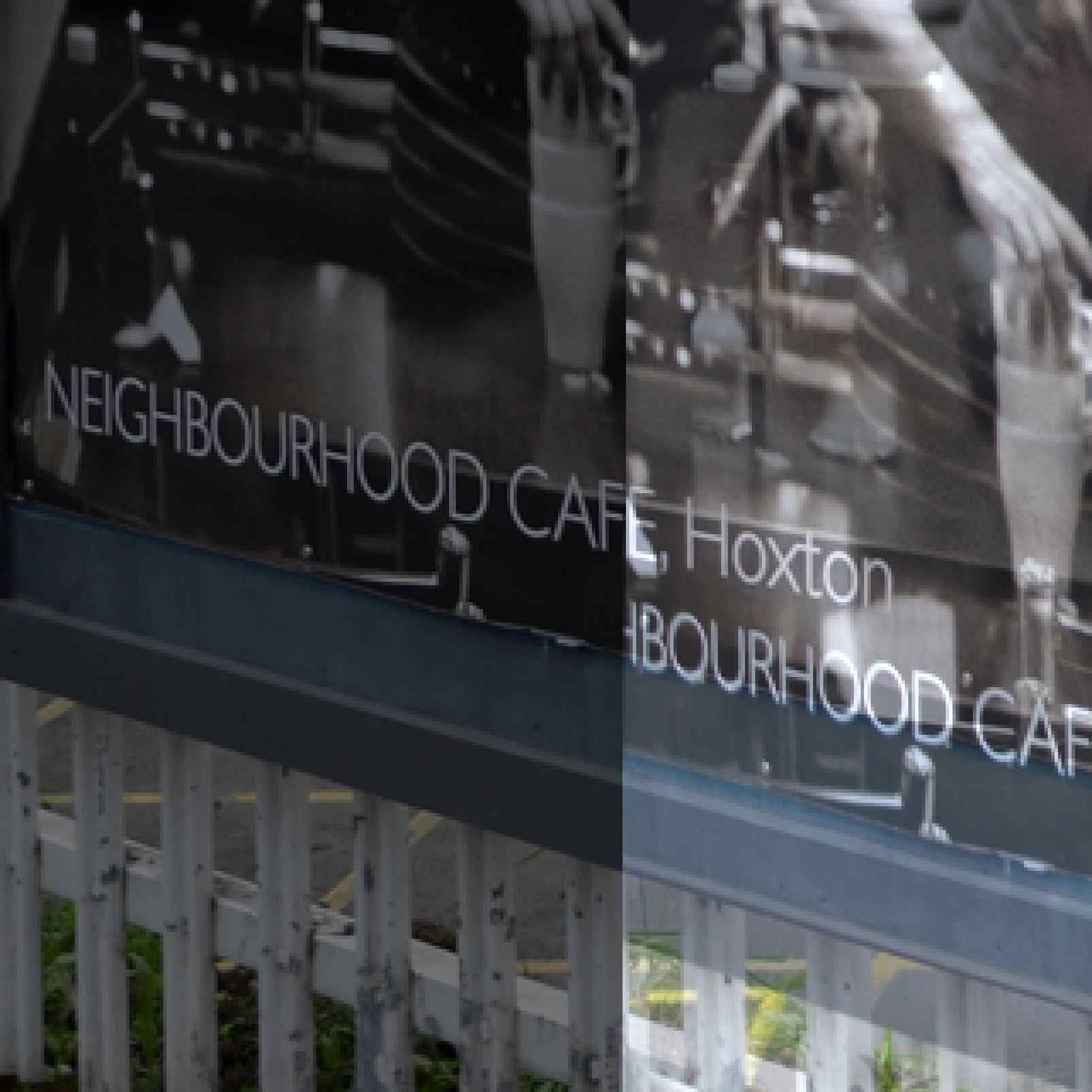}
\par\medskip
\includegraphics[width=0.36\linewidth]{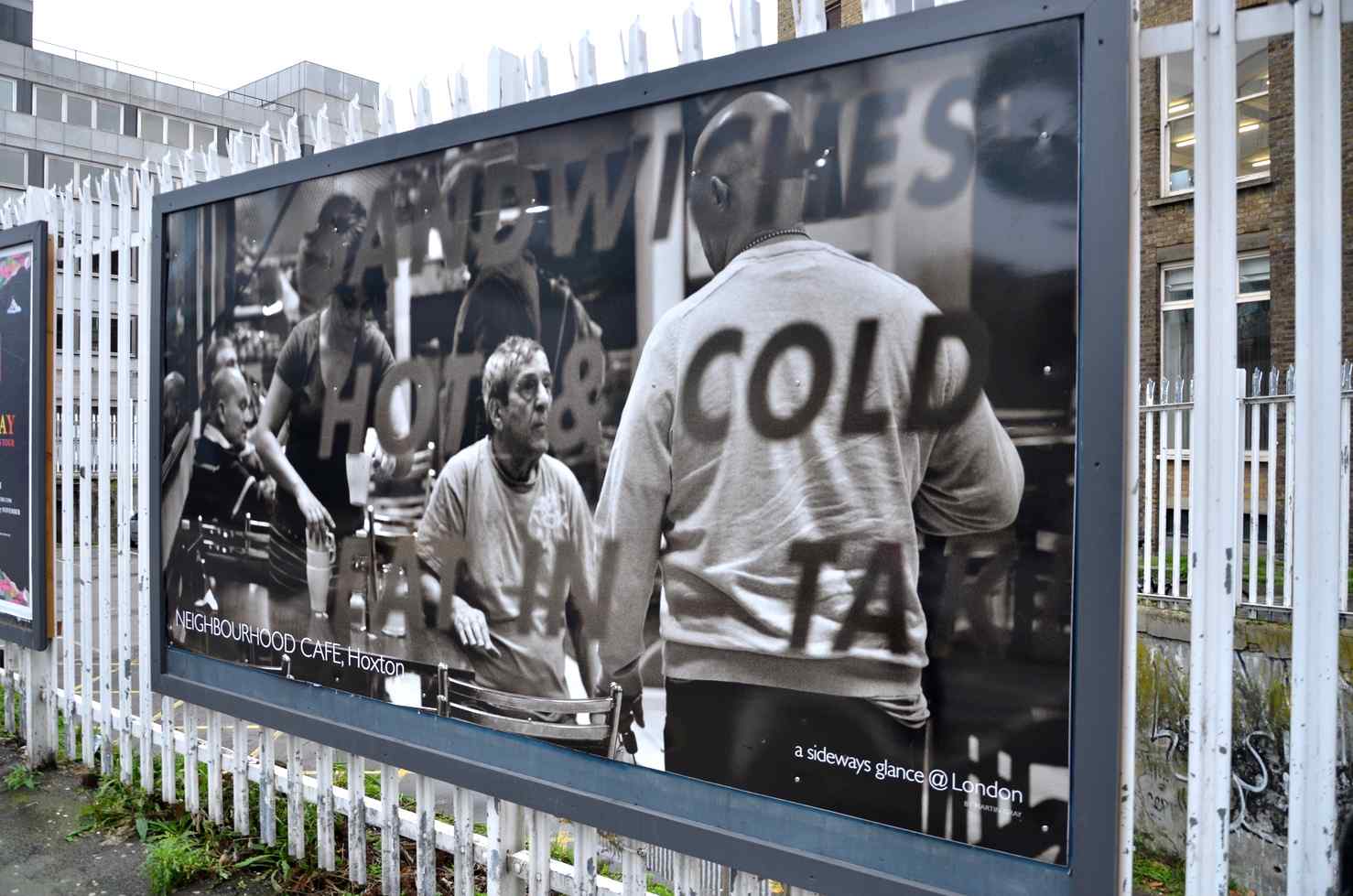}
\hfill
\includegraphics[width=0.36\linewidth]{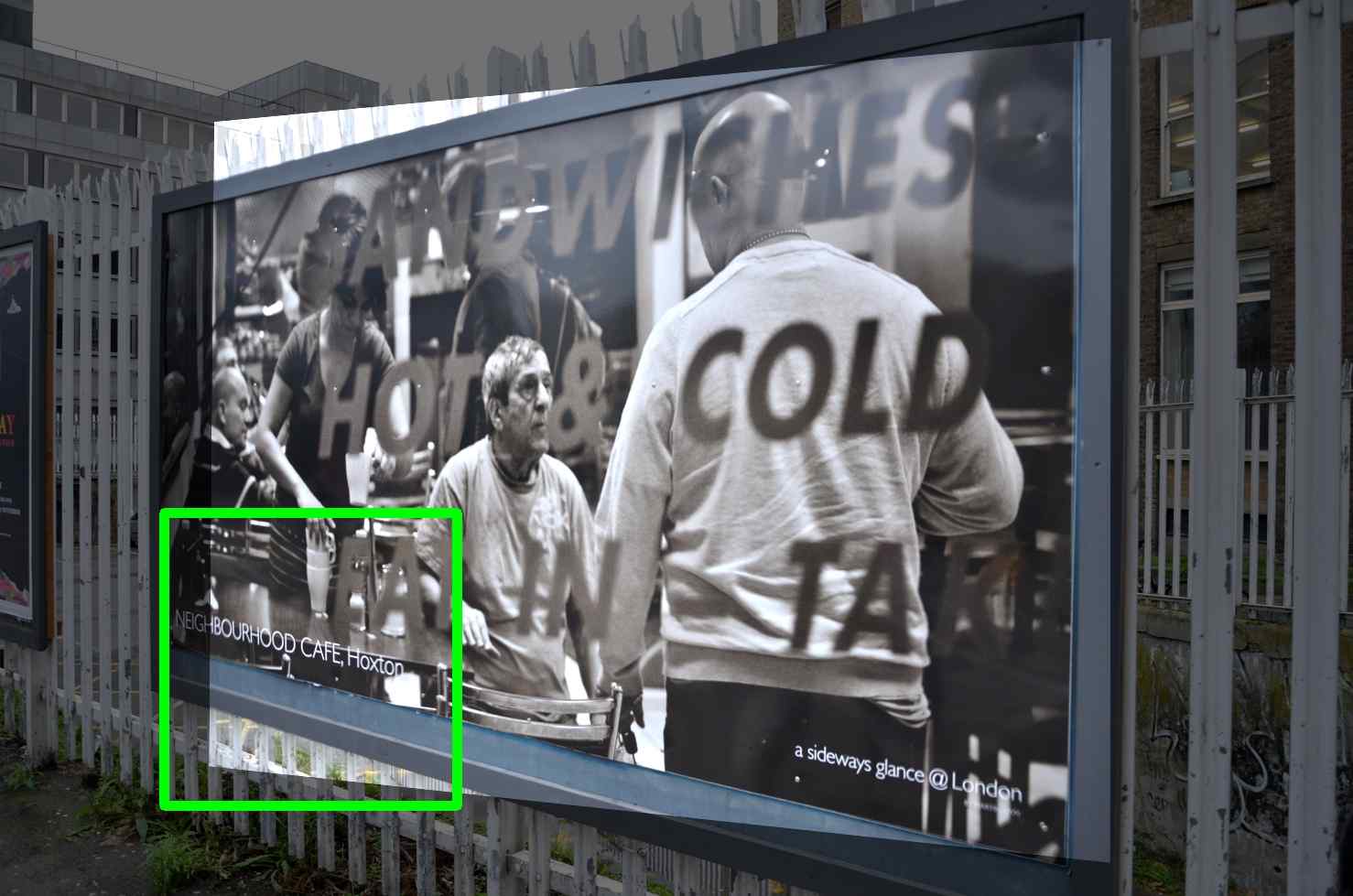}
\hfill
\includegraphics[width=0.238\linewidth]{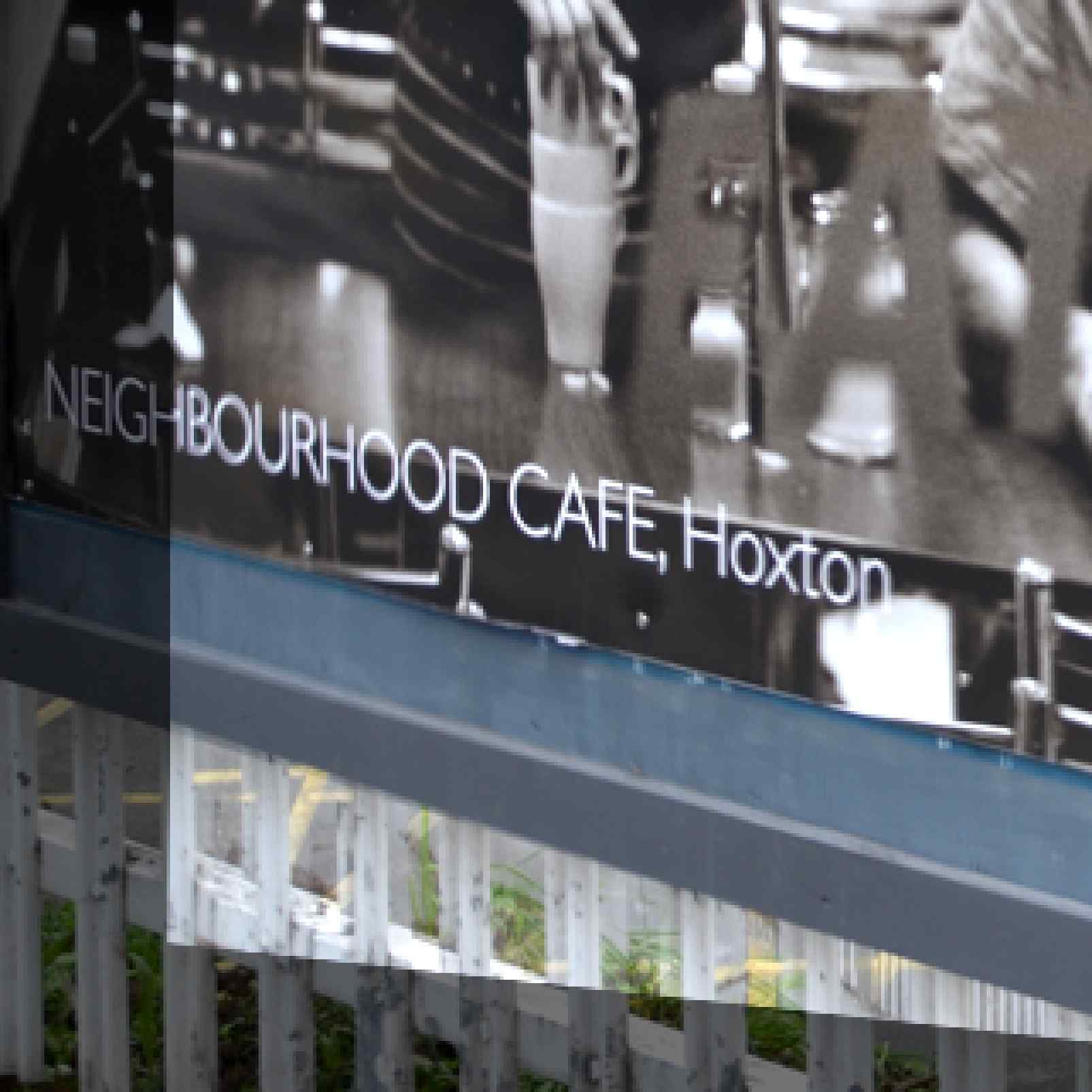}
\caption{Alignment Result}
\end{subfigure}
\par\bigskip
\begin{subfigure}[b]{1.0\textwidth}
\begin{center}
\includegraphics[width=0.8\linewidth]{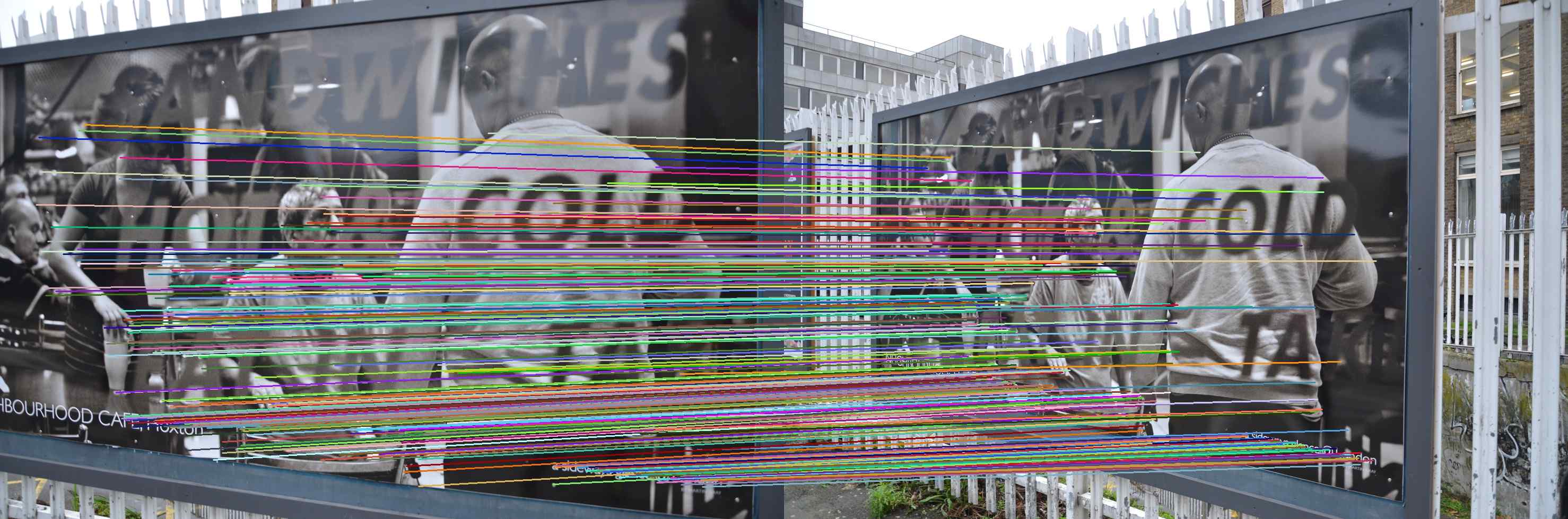}
\caption{Matching Result}
\end{center}
\end{subfigure}
\caption{
\textsc{Tough} pair from HPatches (v\_coffeehouse):
(a) \textsc{Left:} The two input images. \textsc{Middle:} The result in the upper row is the initial alignment. Here both images have been overlaid and blended with each other. The lower row shows the alignment computed by our estimated homography. \textsc{Right:} Zoomed-in views of a specific region of the image shows the quality of the initial and final alignments. (b) The sparse feature matches between the two images. These are computed by extracting SIFT keypoints and matching feature descriptors learned using our method. The matches are filtered using a geometric verification step that robustly fits a homography and finds inlier matches.
The outliers are not shown.
}
\end{figure*}

\begin{figure*}[t]
\begin{subfigure}[b]{1.0\textwidth}
\includegraphics[width=0.36\linewidth]{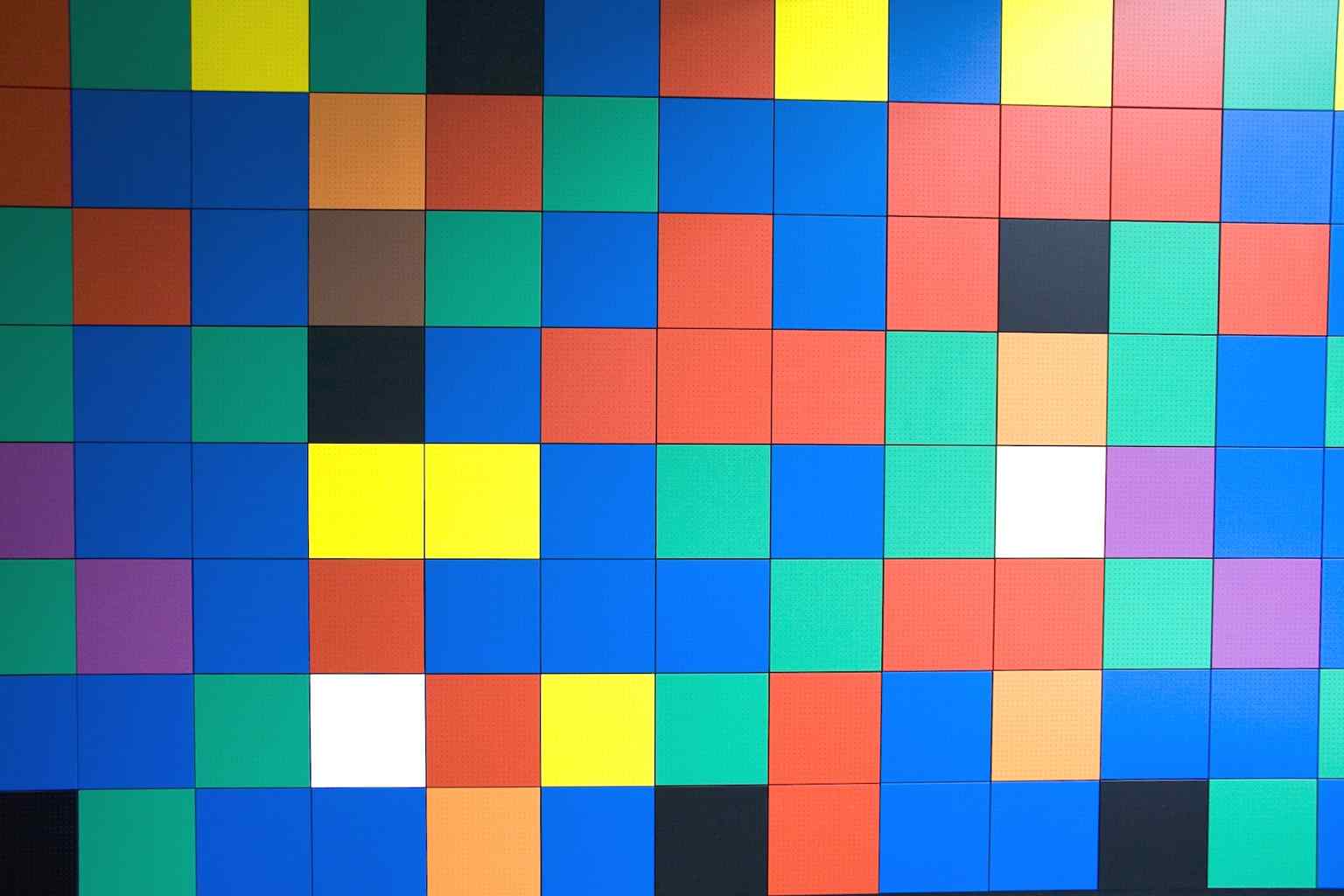}
\hfill
\includegraphics[width=0.36\linewidth]{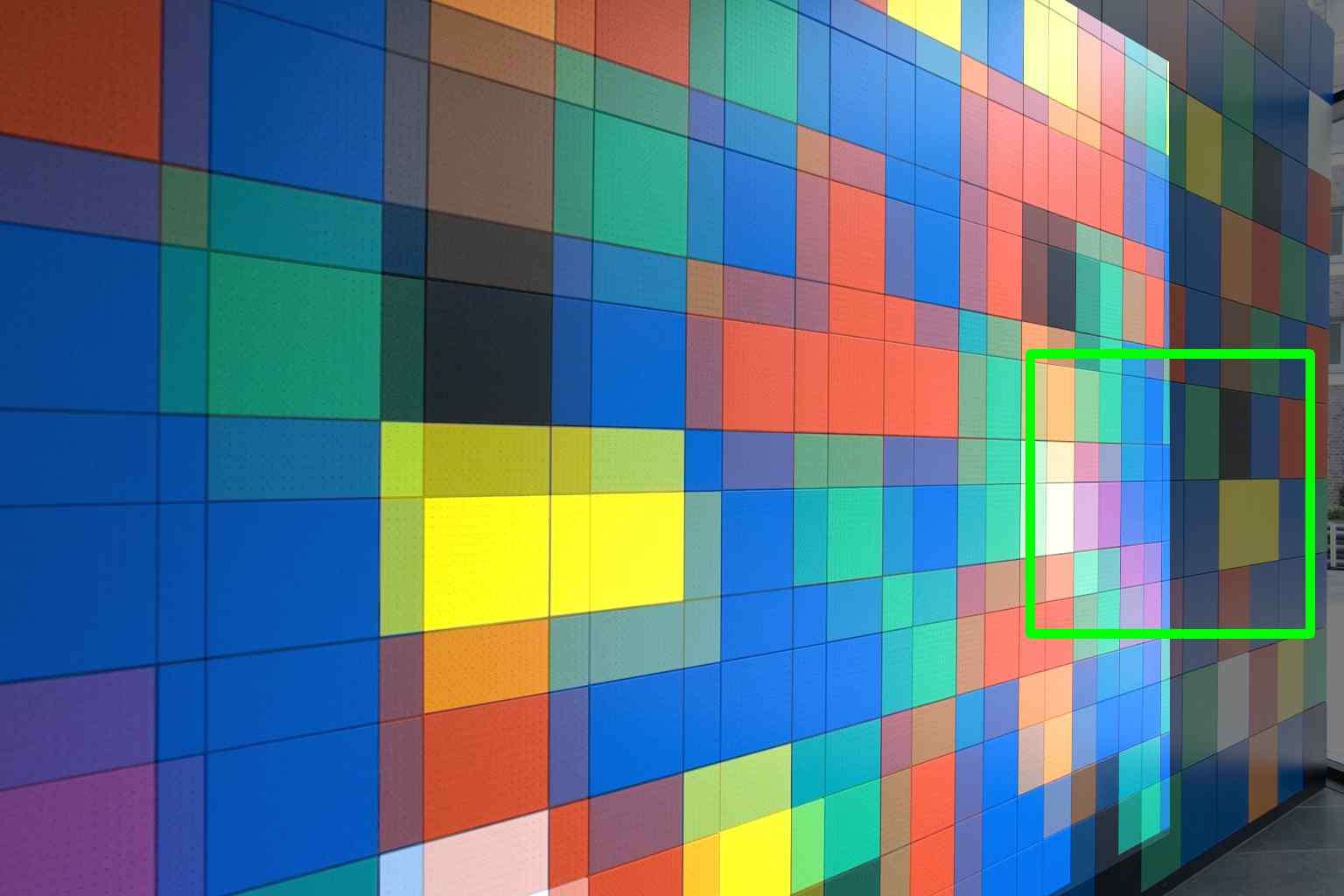}
\hfill
\includegraphics[width=0.24\linewidth]{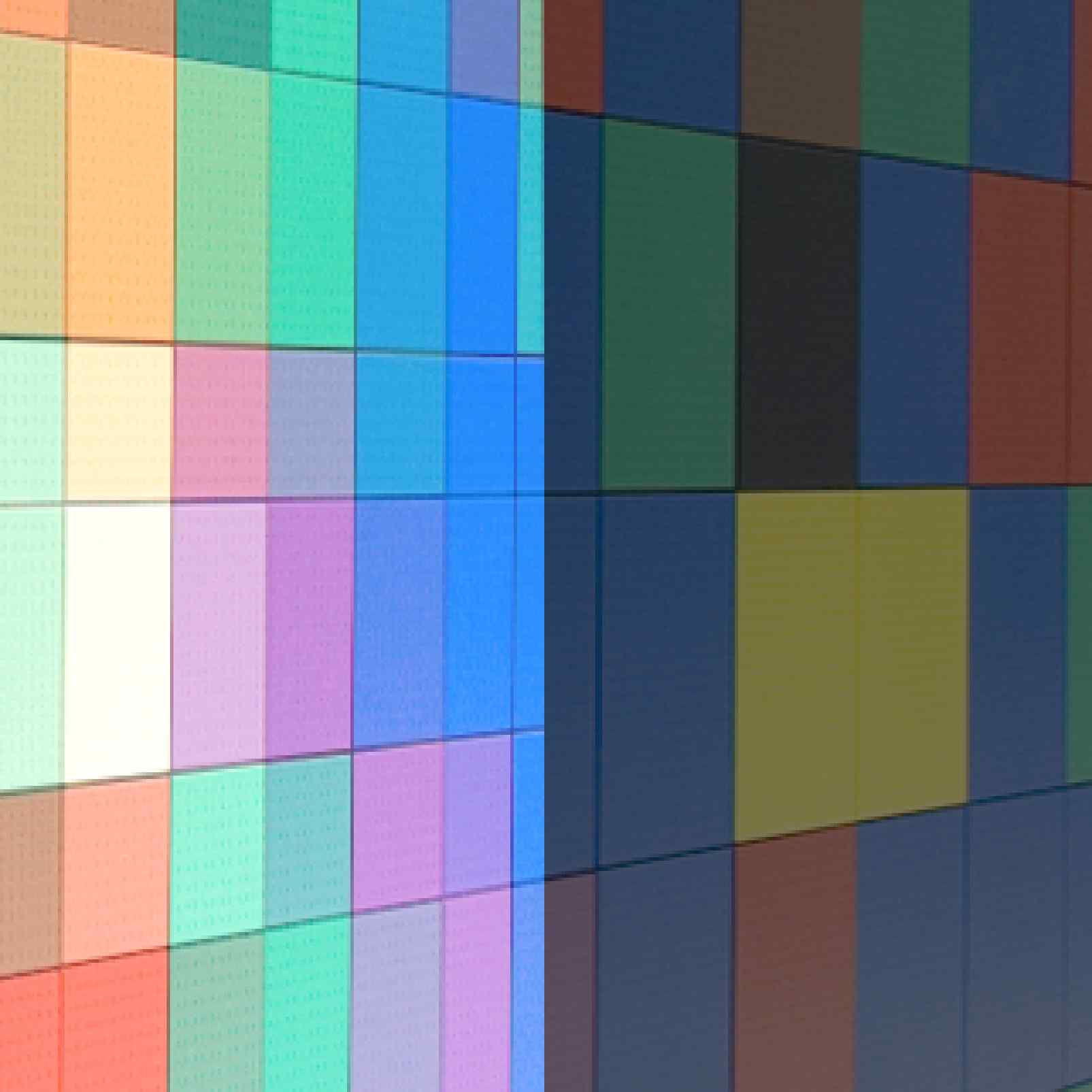}
\par\medskip
\includegraphics[width=0.36\linewidth]{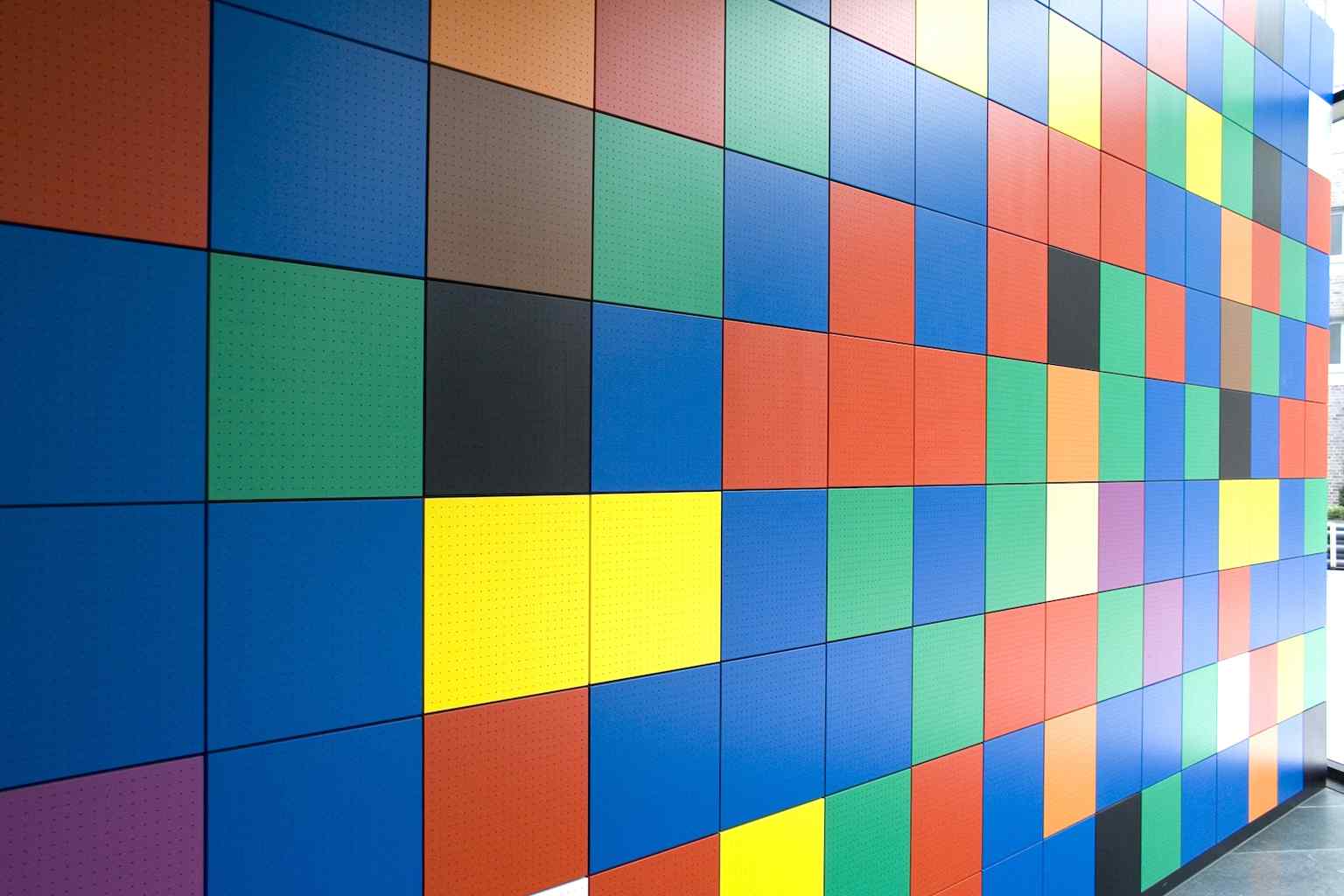}
\hfill
\includegraphics[width=0.36\linewidth]{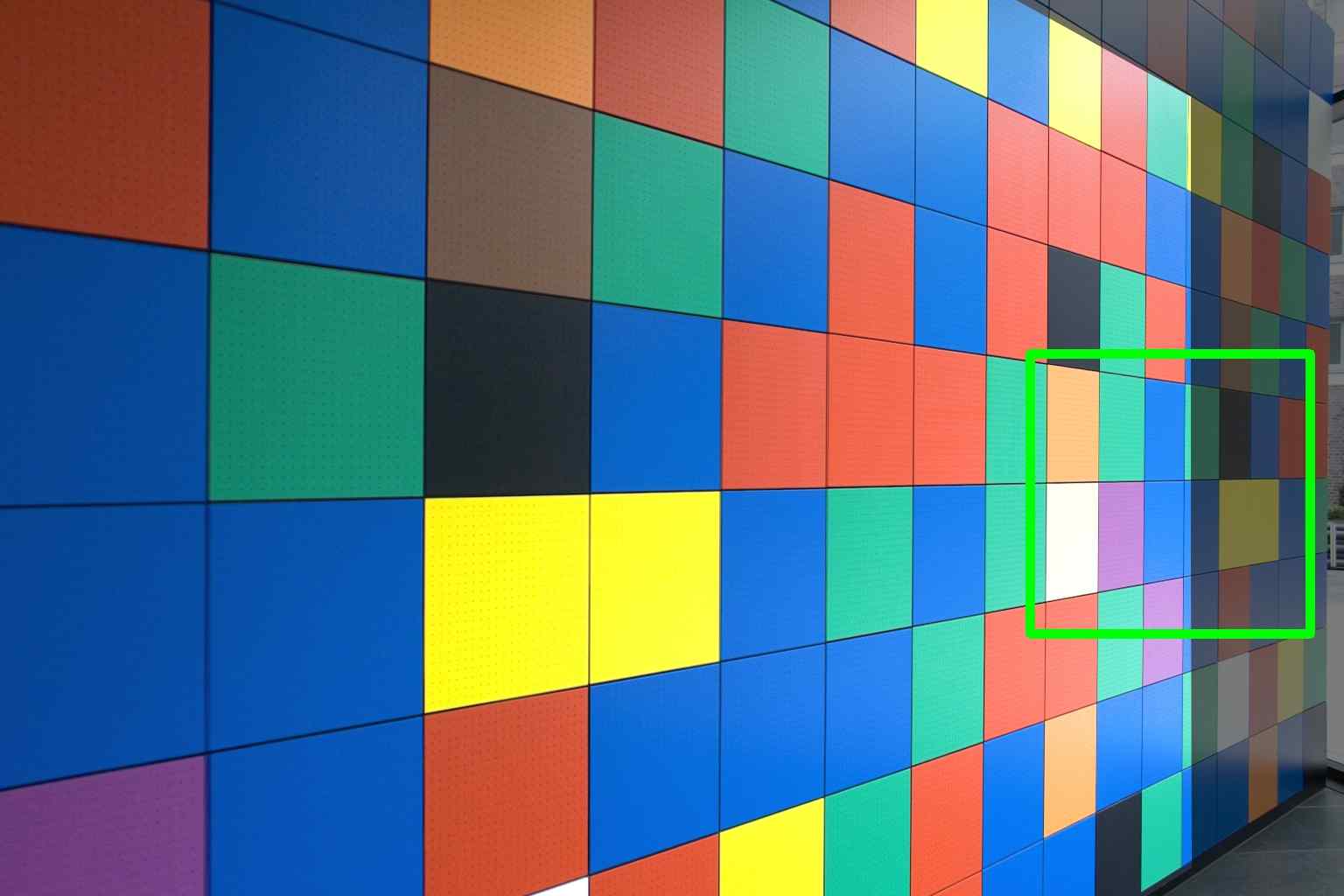}
\hfill
\includegraphics[width=0.24\linewidth]{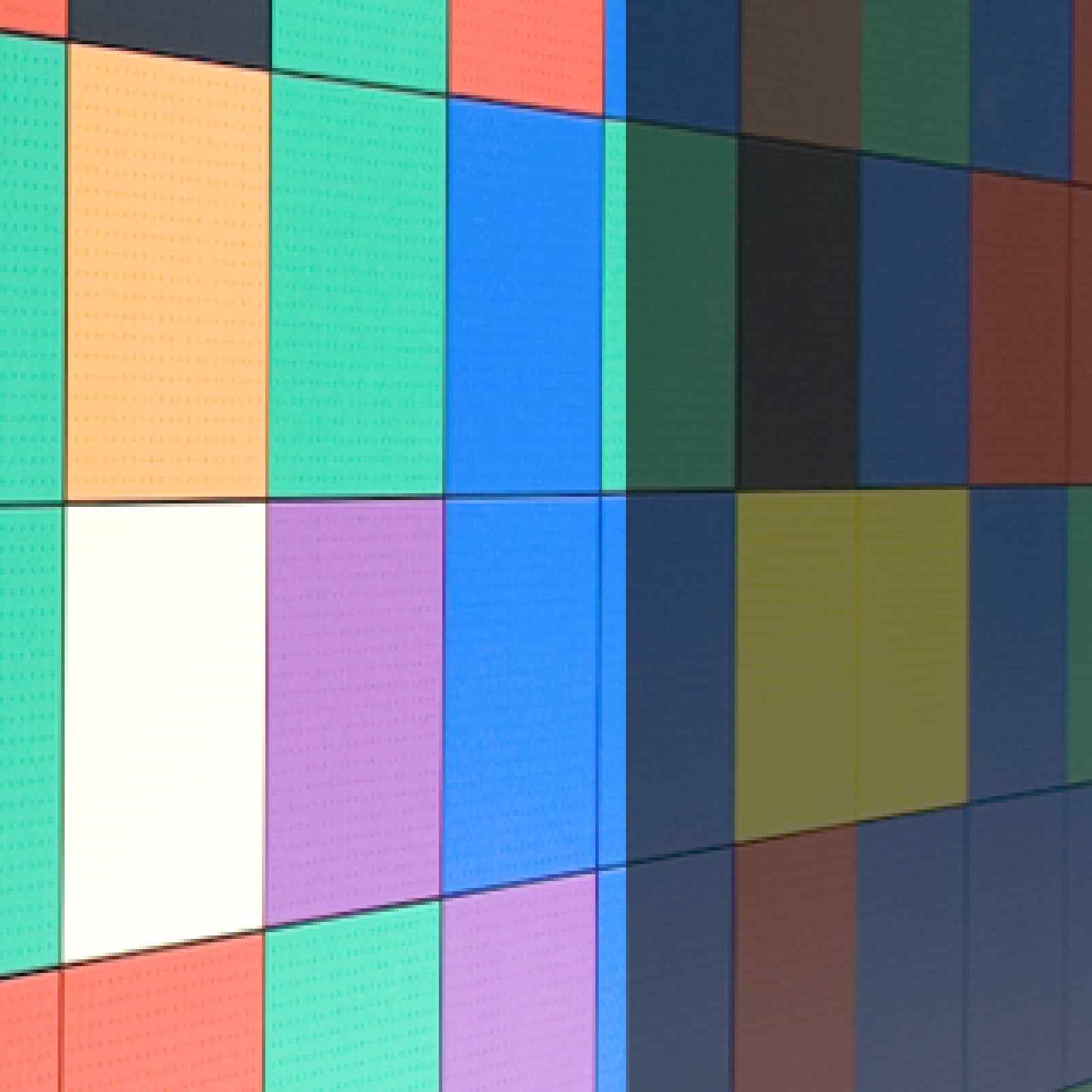}
\caption{Alignment result}
\end{subfigure}
\par\bigskip
\begin{subfigure}[b]{1.0\textwidth}
\begin{center}
\includegraphics[width=0.8\linewidth]{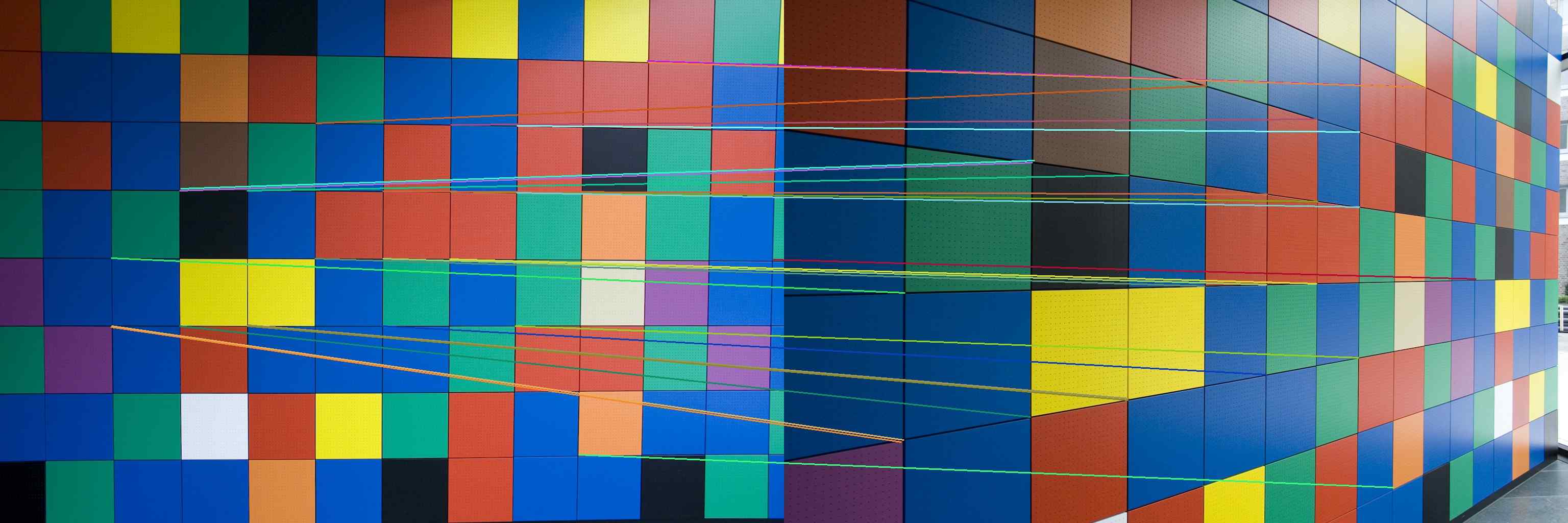}
\caption{Matching result}
\end{center}
\end{subfigure}
\caption{
\textsc{Hard} pair from HPatches (v\_colors):
(a) \textsc{Left:} The two input images. \textsc{Middle:} The result in the upper row is the initial alignment. Here both images have been overlaid and blended with each other. The lower row shows the alignment computed by our estimated homography. \textsc{Right:} Zoomed-in views of a specific region of the image shows the quality of the initial and final alignments. (b) The sparse feature matches between the two images. These are computed by extracting SIFT keypoints and matching feature descriptors learned using our method. The matches are filtered using a geometric verification step that robustly fits a homography and finds inlier matches.
The outliers are not shown.
}
\end{figure*}

\begin{figure*}[t]
\begin{subfigure}[b]{1.0\textwidth}
\includegraphics[width=0.35\linewidth]{figs/results/img1_i_contruction}
\hfill
\includegraphics[width=0.35\linewidth]{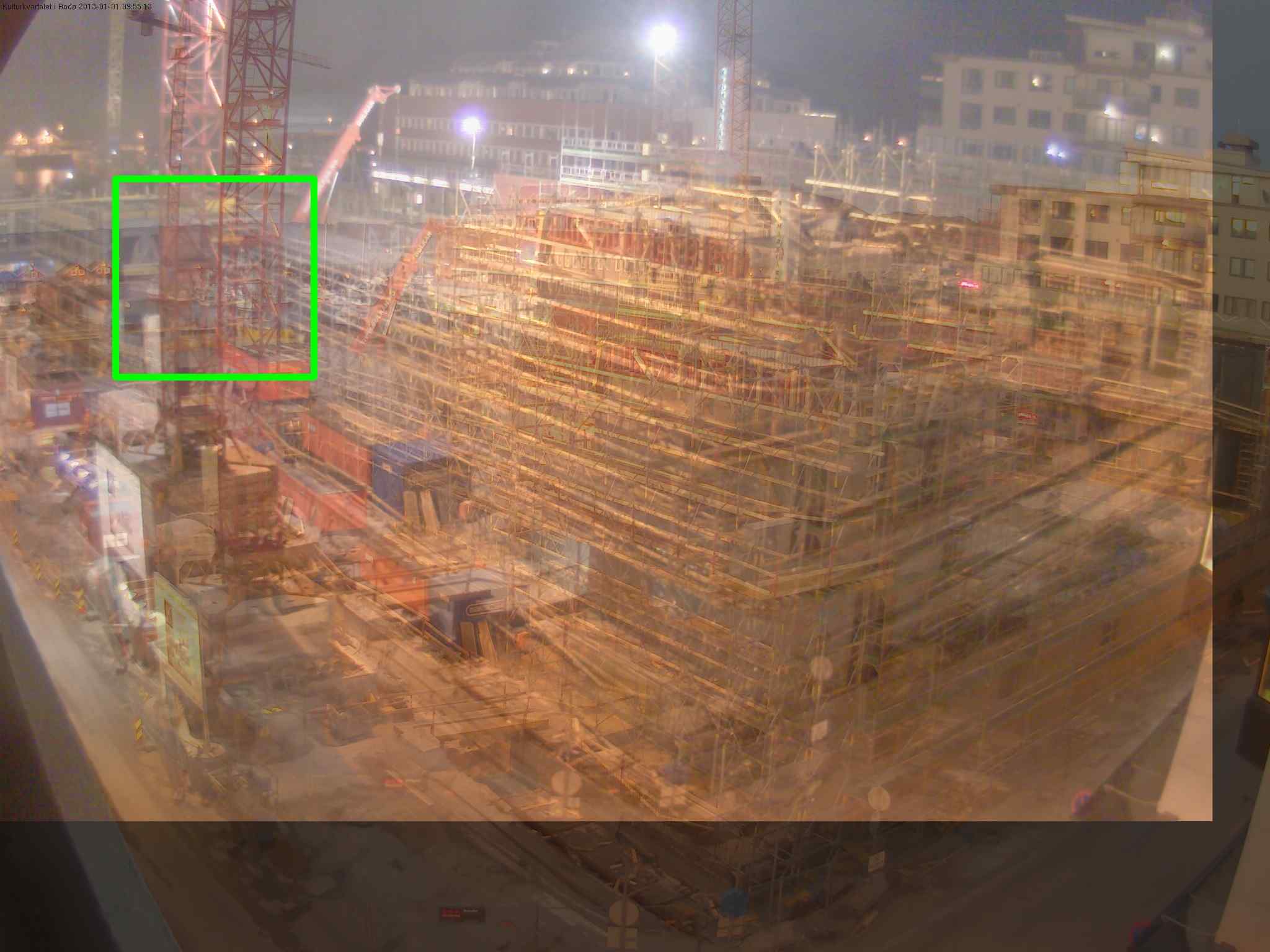}
\hfill
\includegraphics[width=0.26\linewidth]{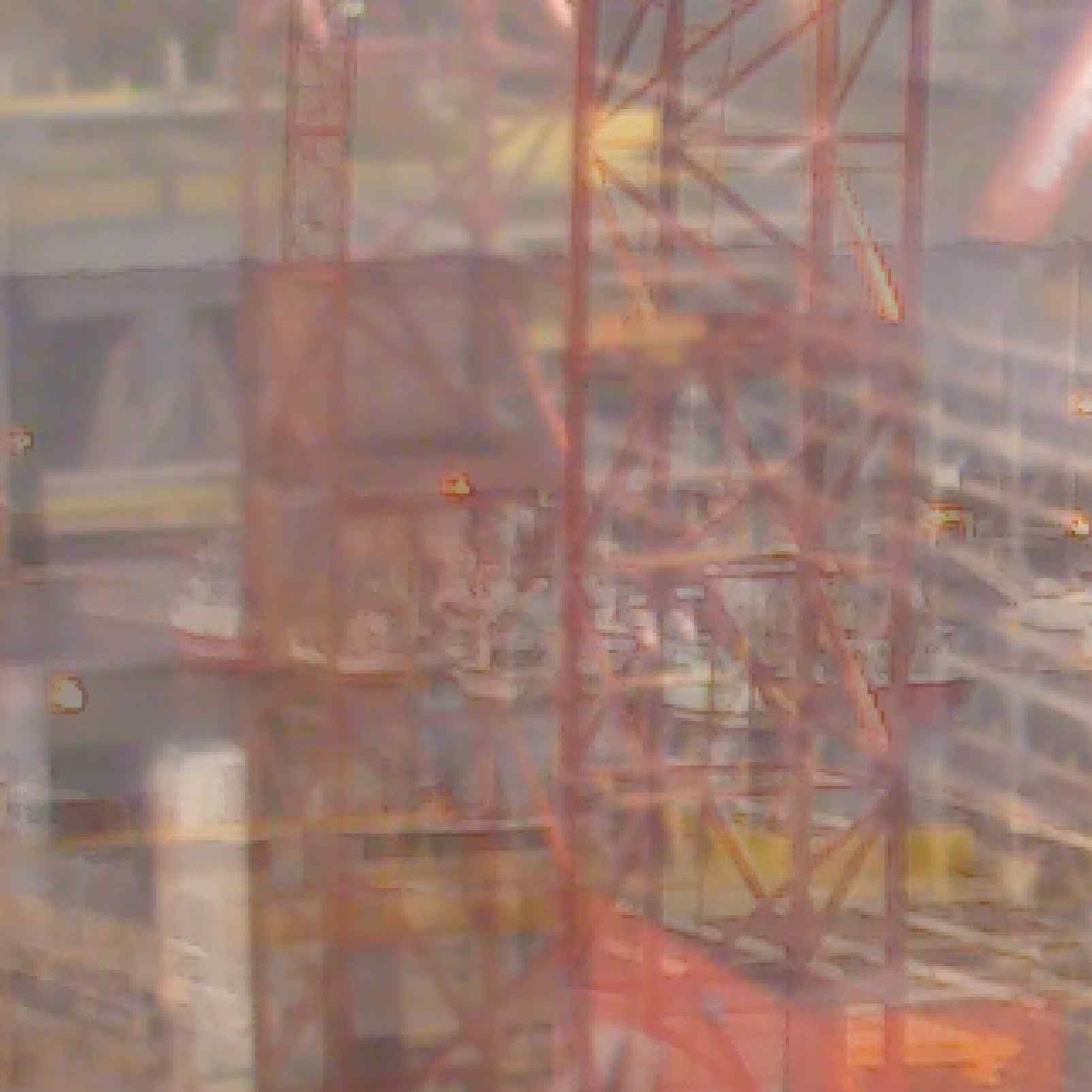}
\par\medskip
\includegraphics[width=0.35\linewidth]{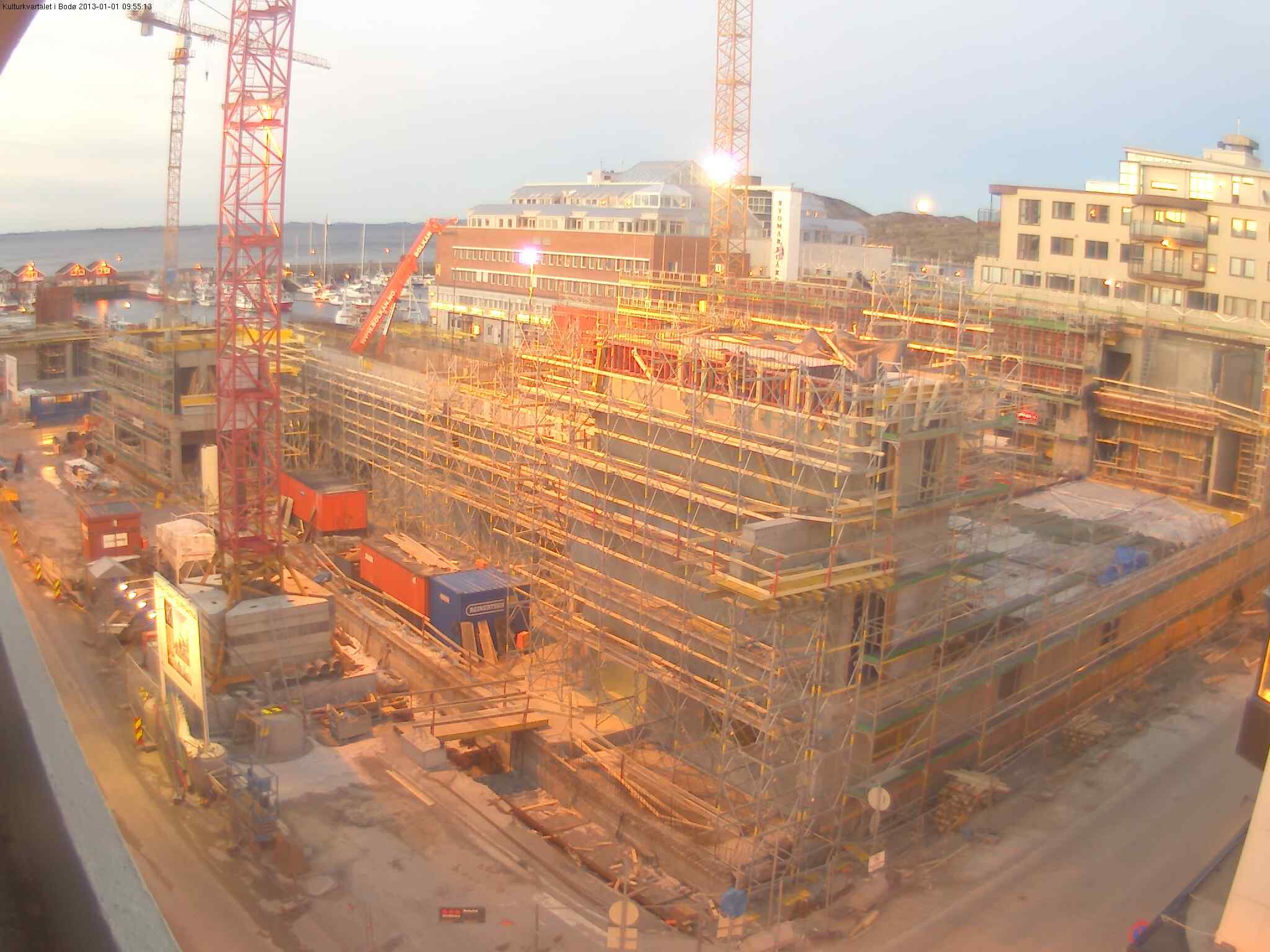}
\hfill
\includegraphics[width=0.35\linewidth]{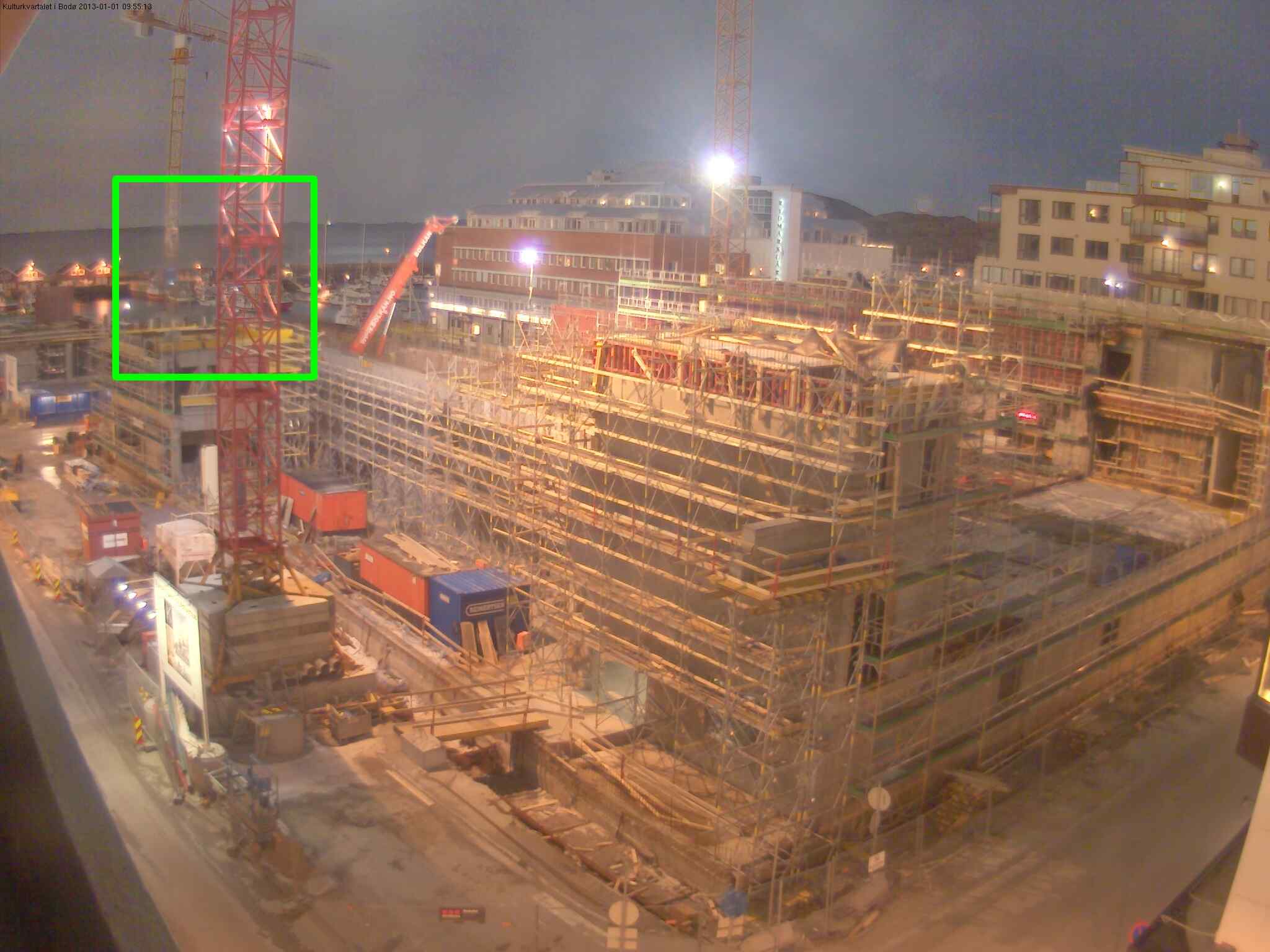}
\hfill
\includegraphics[width=0.26\linewidth]{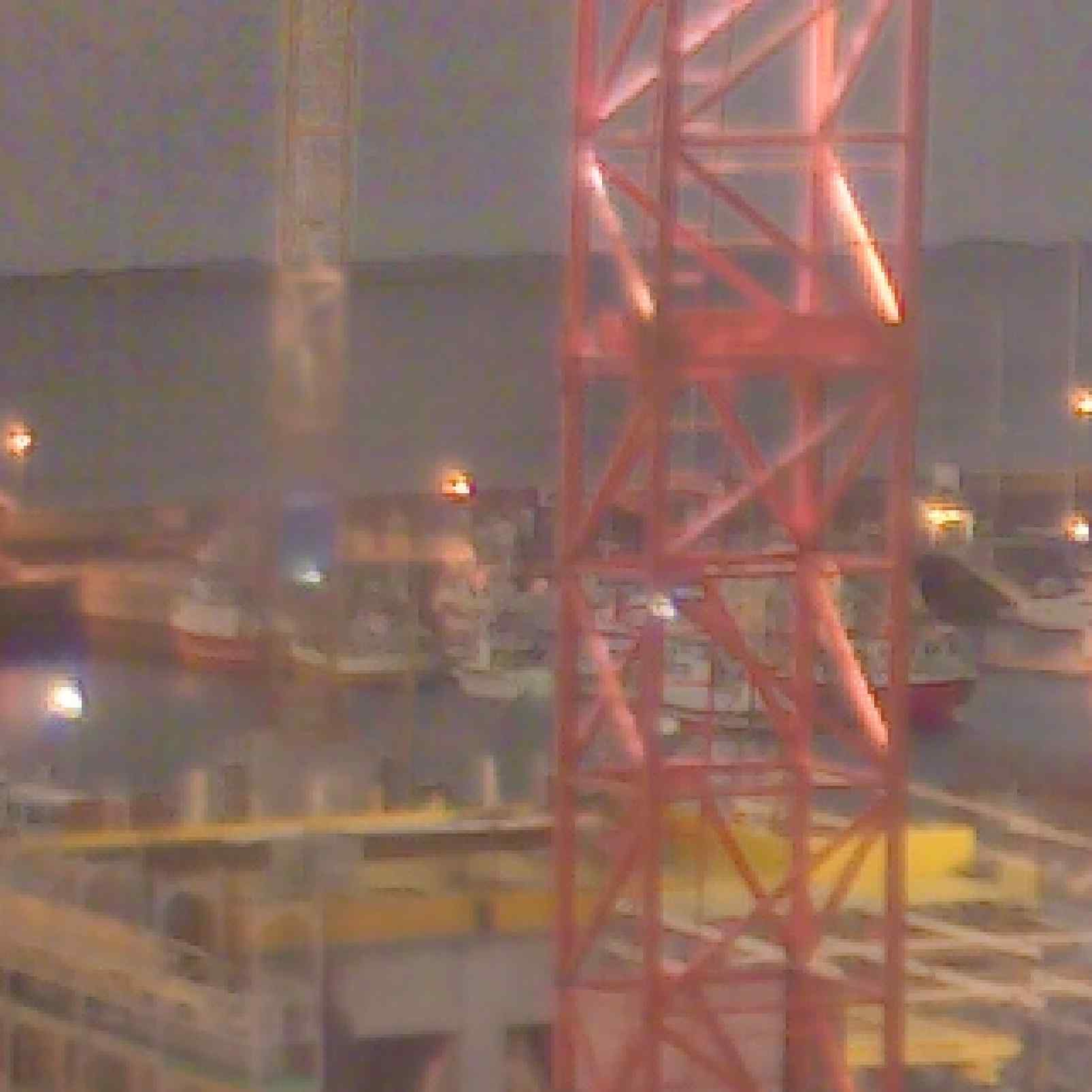}
\caption{Alignment result}
\end{subfigure}
\par\bigskip
\begin{subfigure}[b]{1.0\textwidth}
\begin{center}
\includegraphics[width=0.8\linewidth]{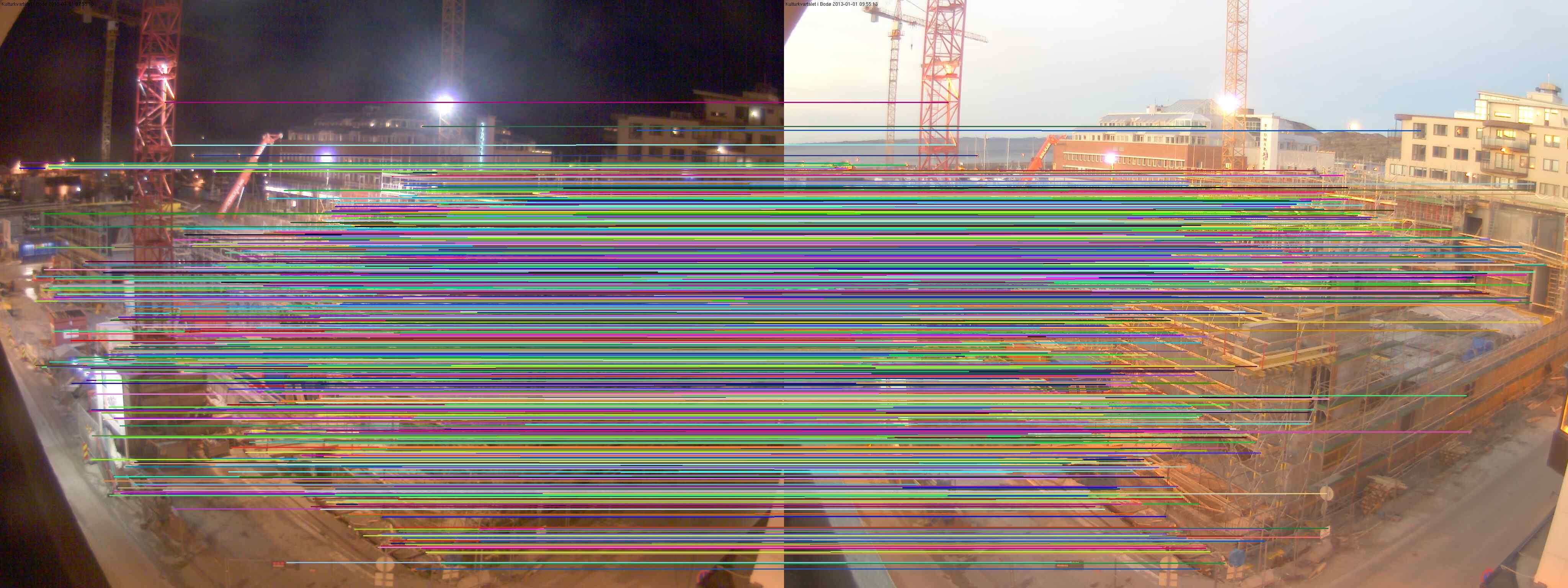}
\caption{Matching result}
\end{center}
\end{subfigure}
\caption{
\textsc{Tough} pair from HPatches (i\_contruction):
(a) \textsc{Left:} The two input images. \textsc{Middle:} The result in the upper row is the initial alignment. Here both images have been overlaid and blended with each other. The lower row shows the alignment computed by our estimated homography. \textsc{Right:} Zoomed-in views of a specific region of the image shows the quality of the initial and final alignments. (b) The sparse feature matches between the two images. These are computed by extracting SIFT keypoints and matching feature descriptors learned using our method. The matches are filtered using a geometric verification step that robustly fits a homography and finds inlier matches.
The outliers are not shown.
}
\end{figure*}

\begin{figure*}[t]
\begin{subfigure}[b]{1.0\textwidth}
\includegraphics[width=0.35\linewidth]{figs/results/img1_i_crownday}
\hfill
\includegraphics[width=0.35\linewidth]{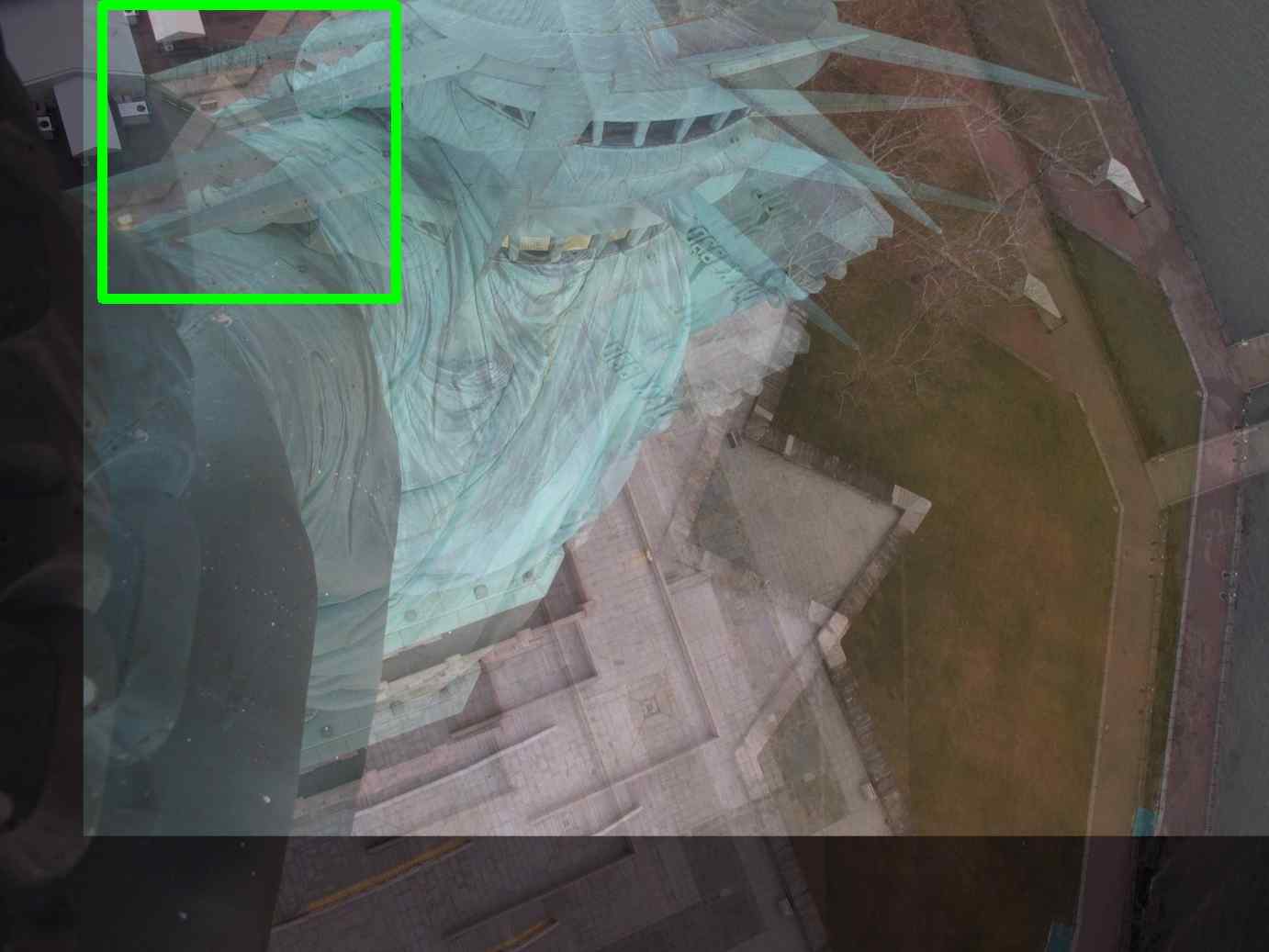}
\hfill
\includegraphics[width=0.26\linewidth]{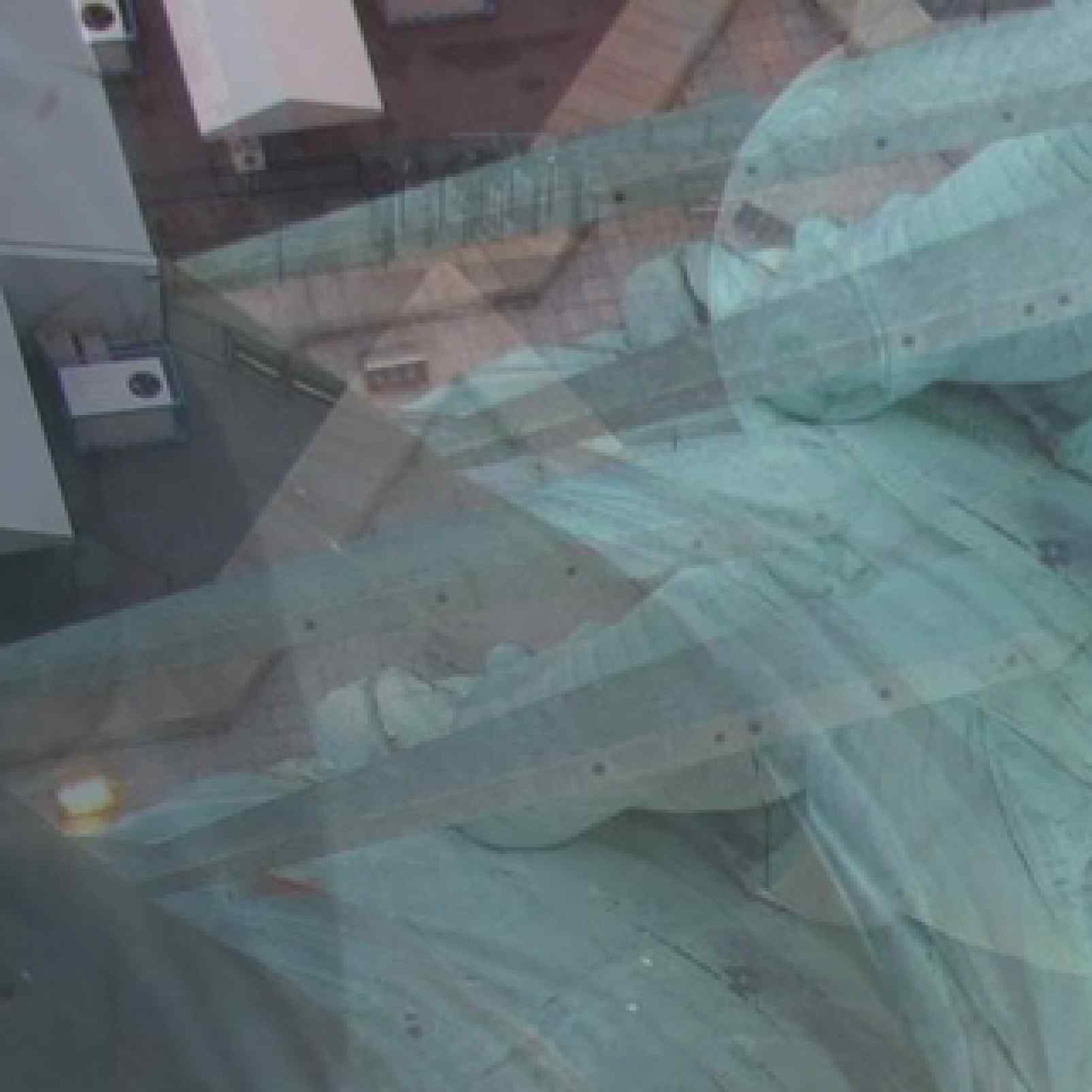}
\par\medskip
\includegraphics[width=0.35\linewidth]{figs/results/img2_i_crownday}
\hfill
\includegraphics[width=0.35\linewidth]{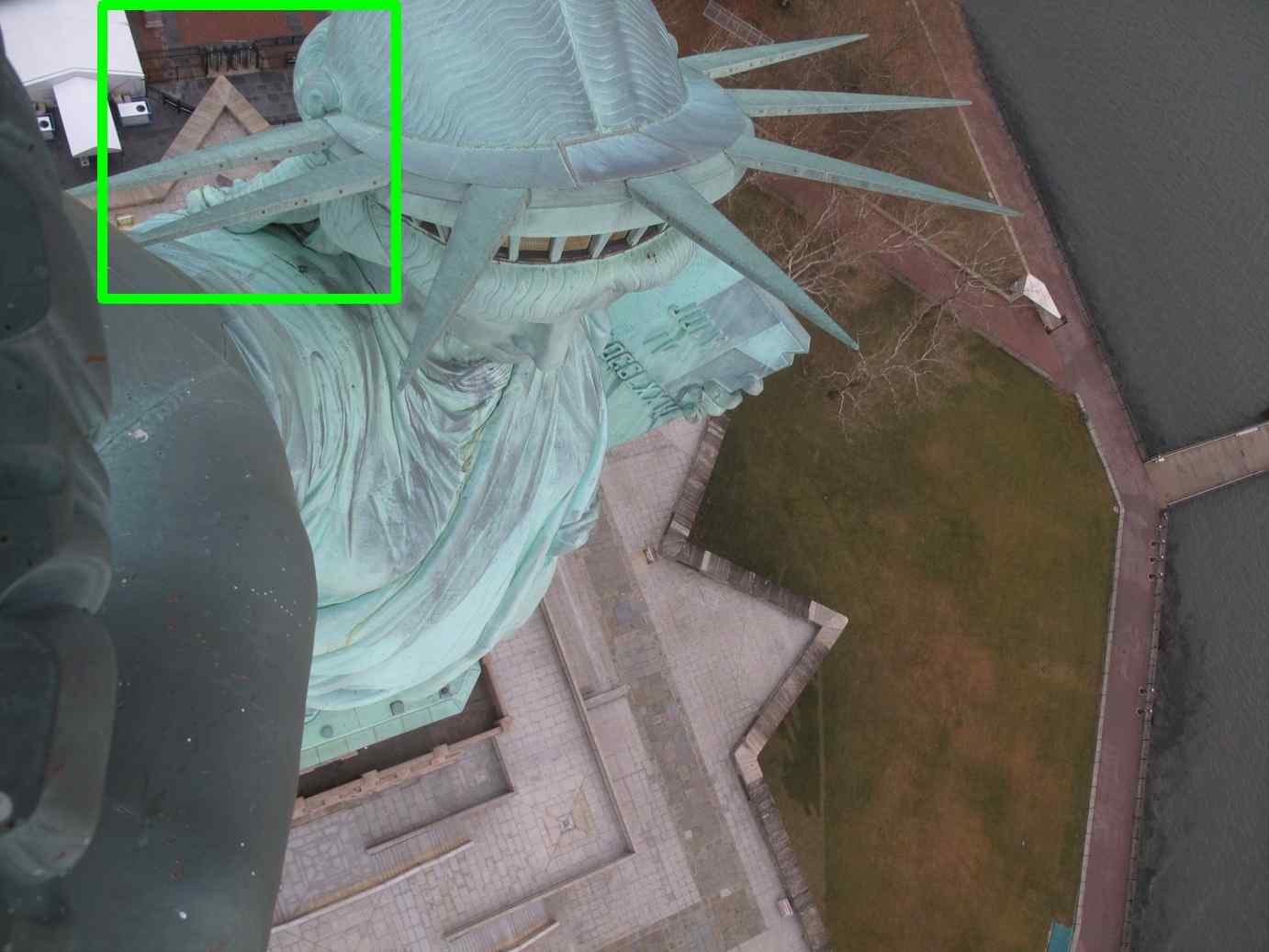}
\hfill
\includegraphics[width=0.26\linewidth]{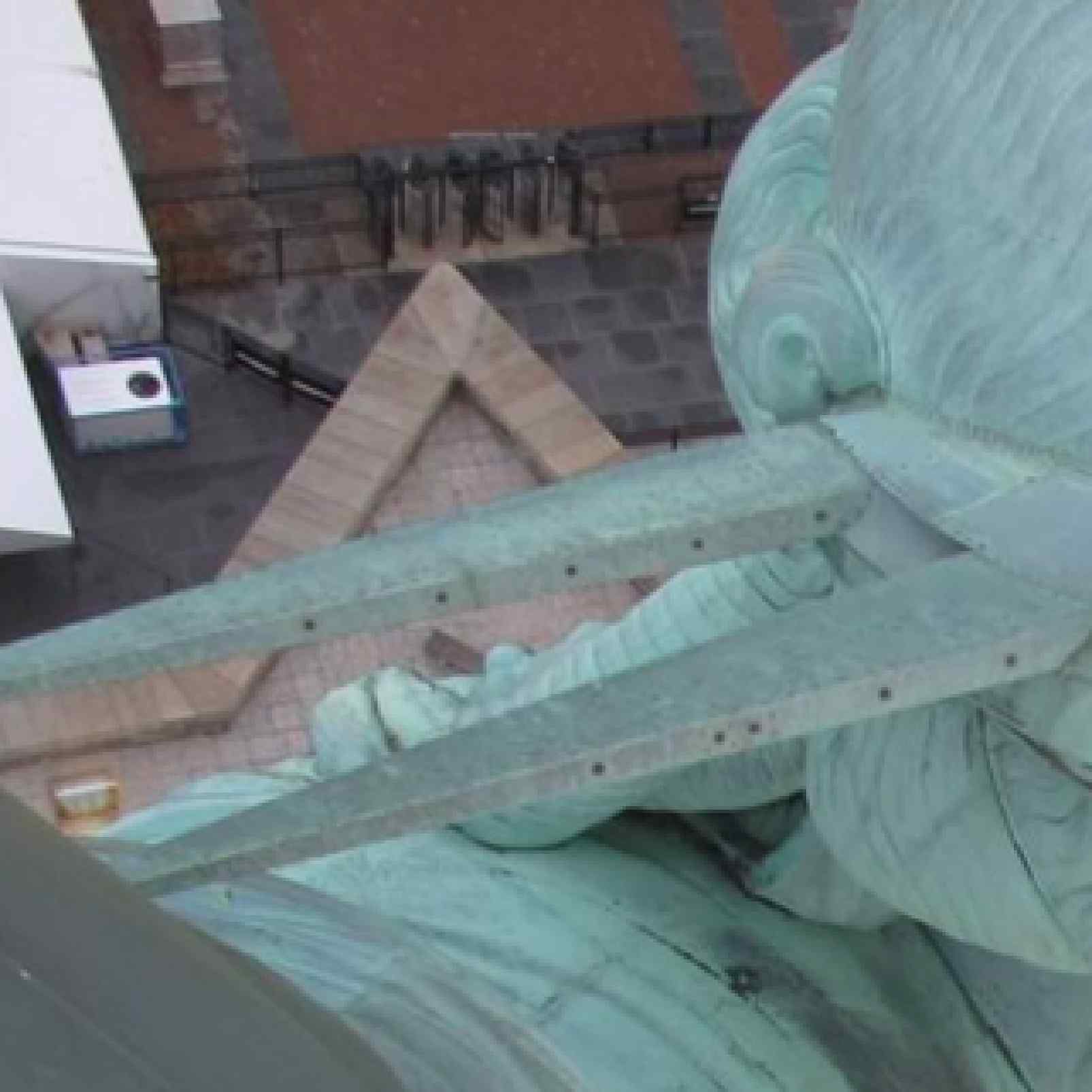}
\caption{Alignment result}
\end{subfigure}
\par\bigskip
\begin{subfigure}[b]{1.0\textwidth}
\begin{center}
\includegraphics[width=0.8\linewidth]{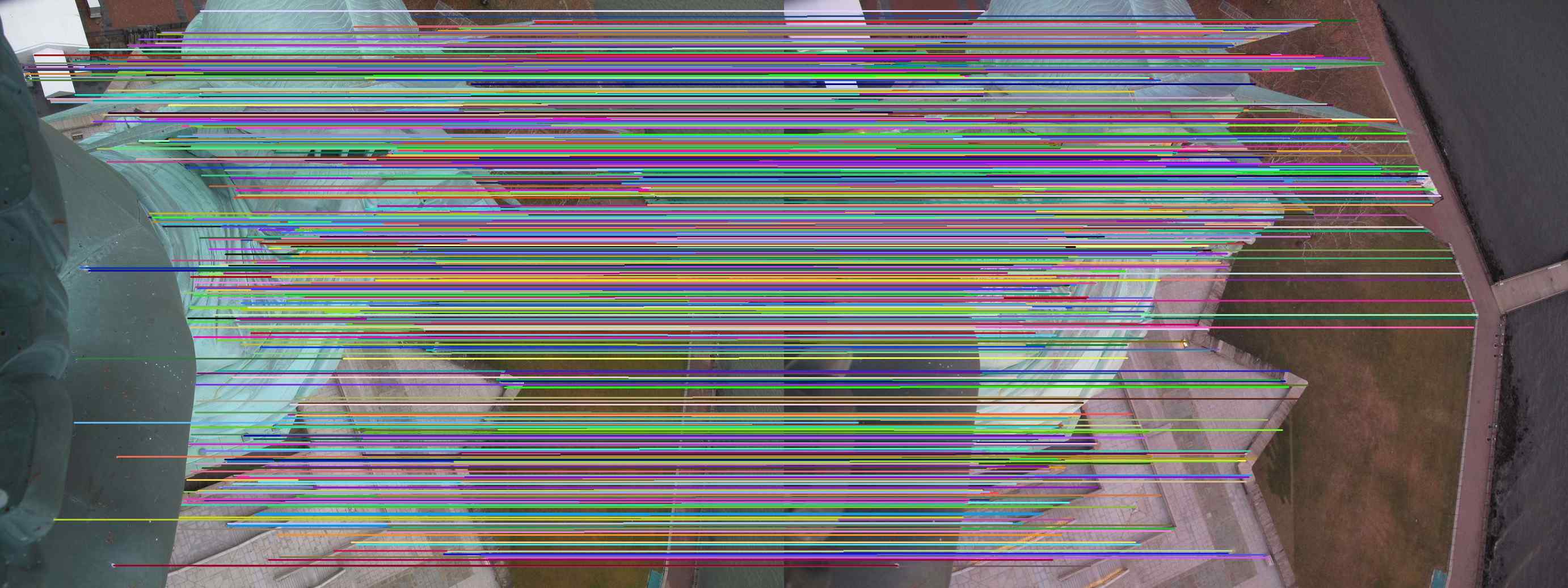}
\caption{Matching result}
\end{center}
\end{subfigure}
\caption{
\textsc{Tough} pair from HPatches (i\_crownday):
(a) \textsc{Left:} The two input images. \textsc{Middle:} The result in the upper row is the initial alignment. Here both images have been overlaid and blended with each other. The lower row shows the alignment computed by our estimated homography. \textsc{Right:} Zoomed-in views of a specific region of the image shows the quality of the initial and final alignments. (b) The sparse feature matches between the two images. These are computed by extracting SIFT keypoints and matching feature descriptors learned using our method. The matches are filtered using a geometric verification step that robustly fits a homography and finds inlier matches.
The outliers are not shown.
}
\end{figure*}

\begin{figure*}[t]
\begin{subfigure}[b]{1.0\textwidth}
\includegraphics[width=0.35\linewidth]{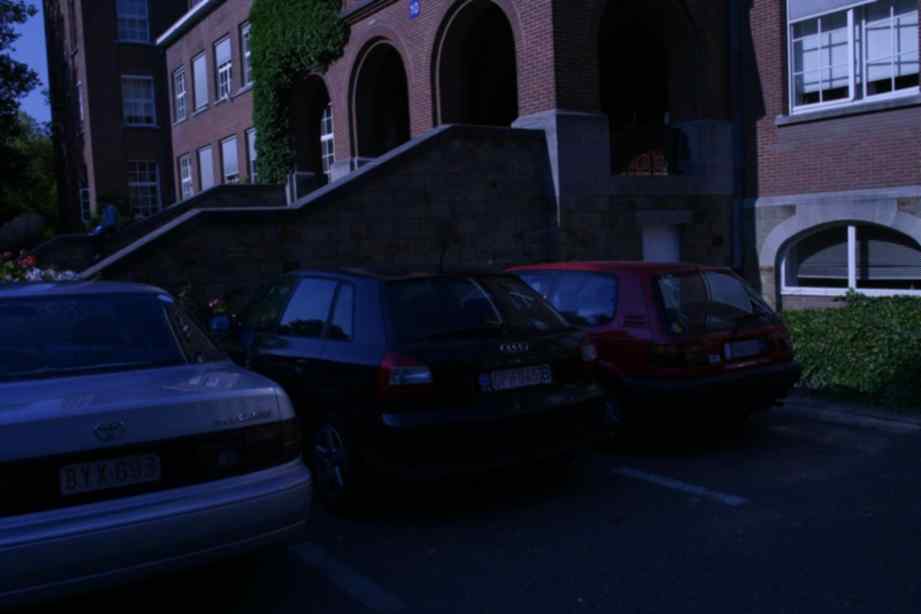}
\hfill
\includegraphics[width=0.35\linewidth]{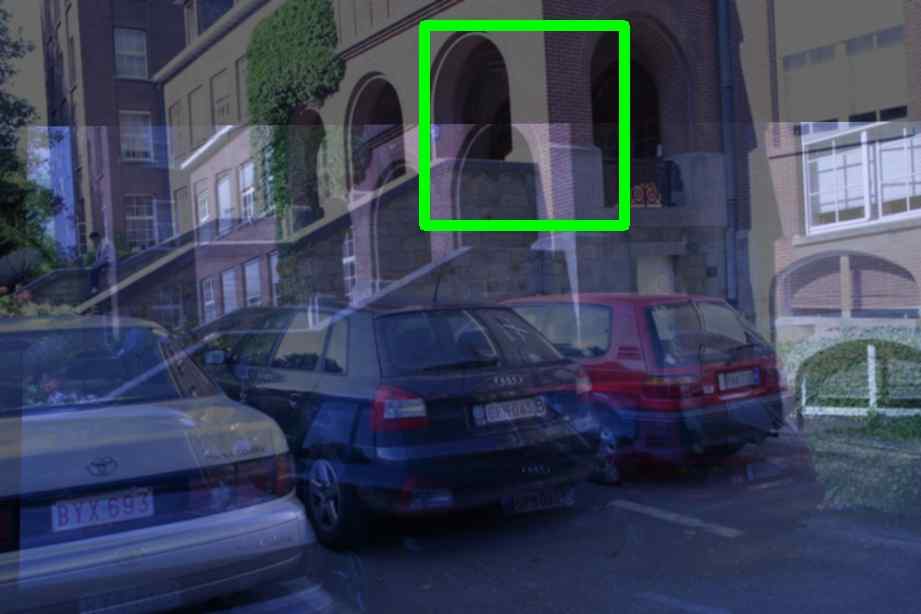}
\hfill
\includegraphics[width=0.26\linewidth]{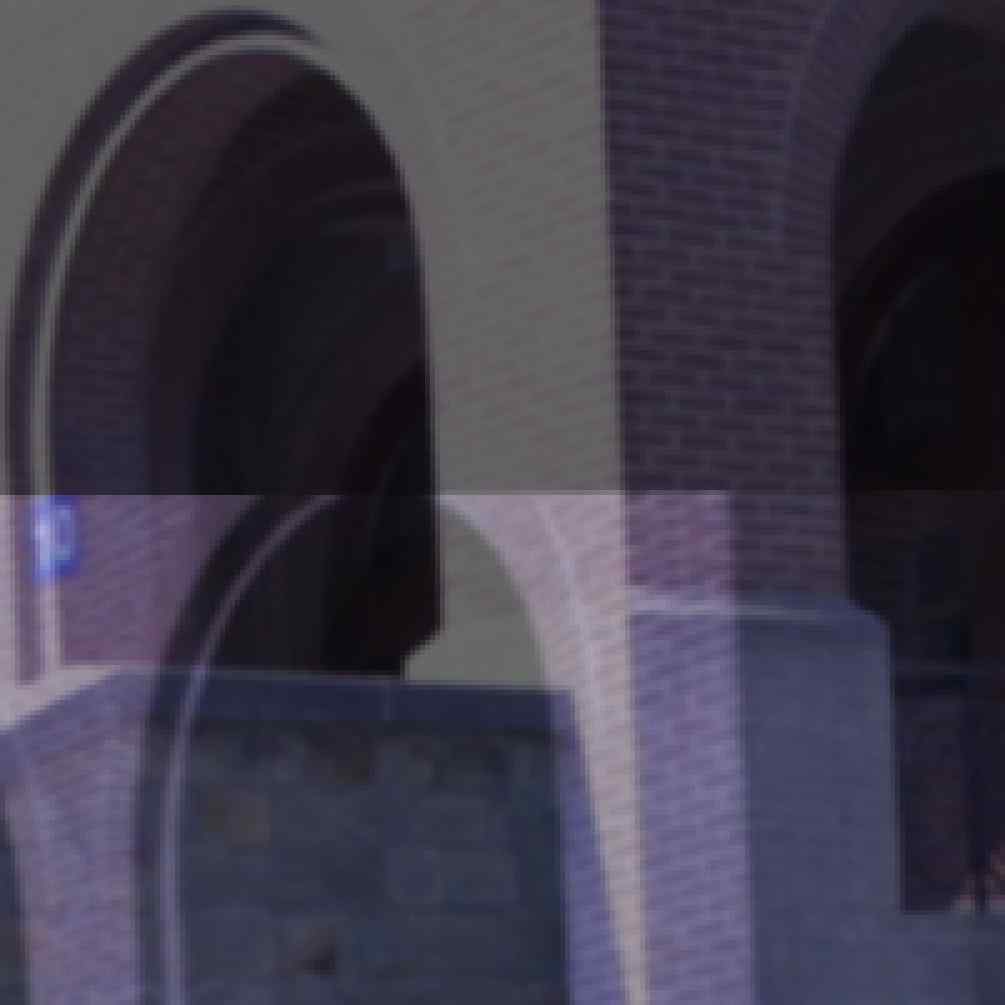}
\par\medskip
\includegraphics[width=0.35\linewidth]{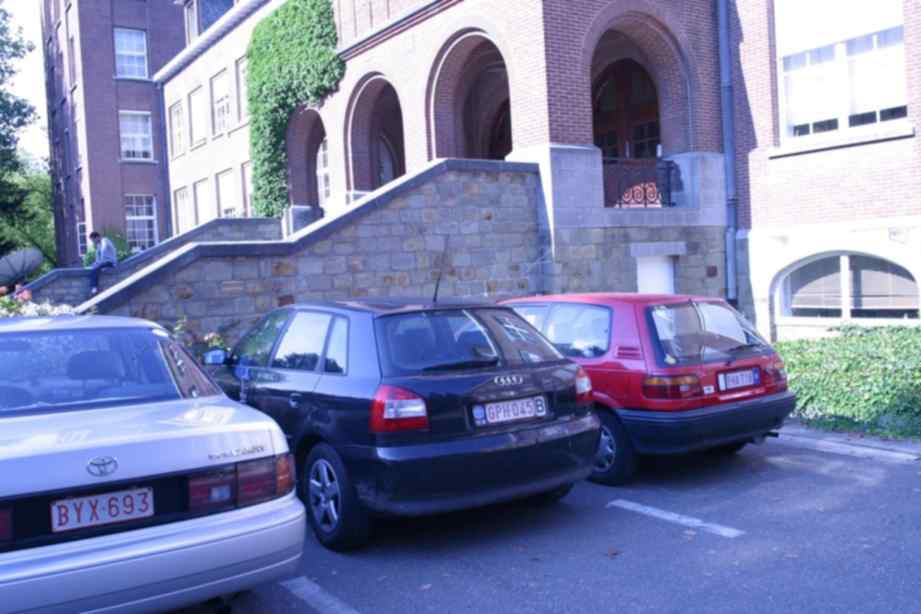}
\hfill
\includegraphics[width=0.35\linewidth]{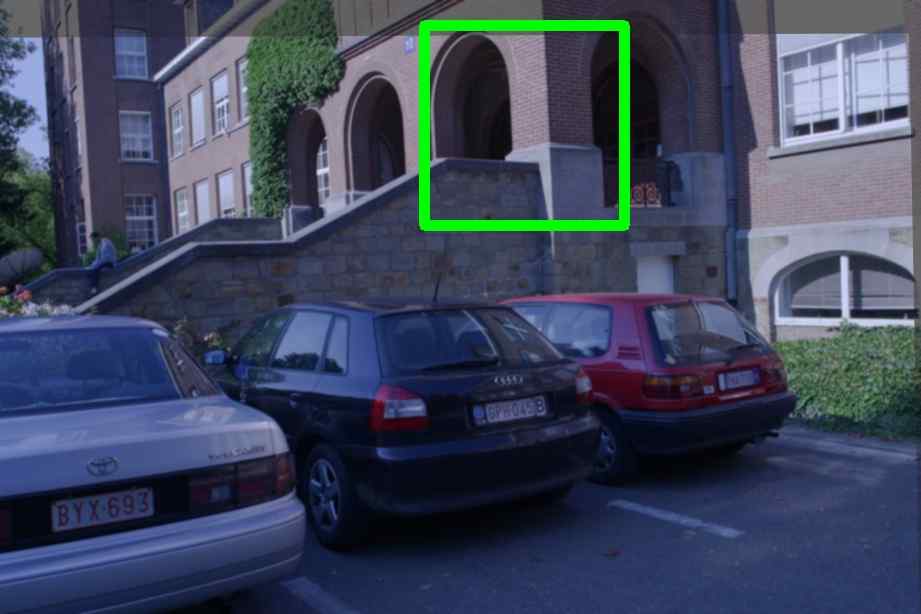}
\hfill
\includegraphics[width=0.26\linewidth]{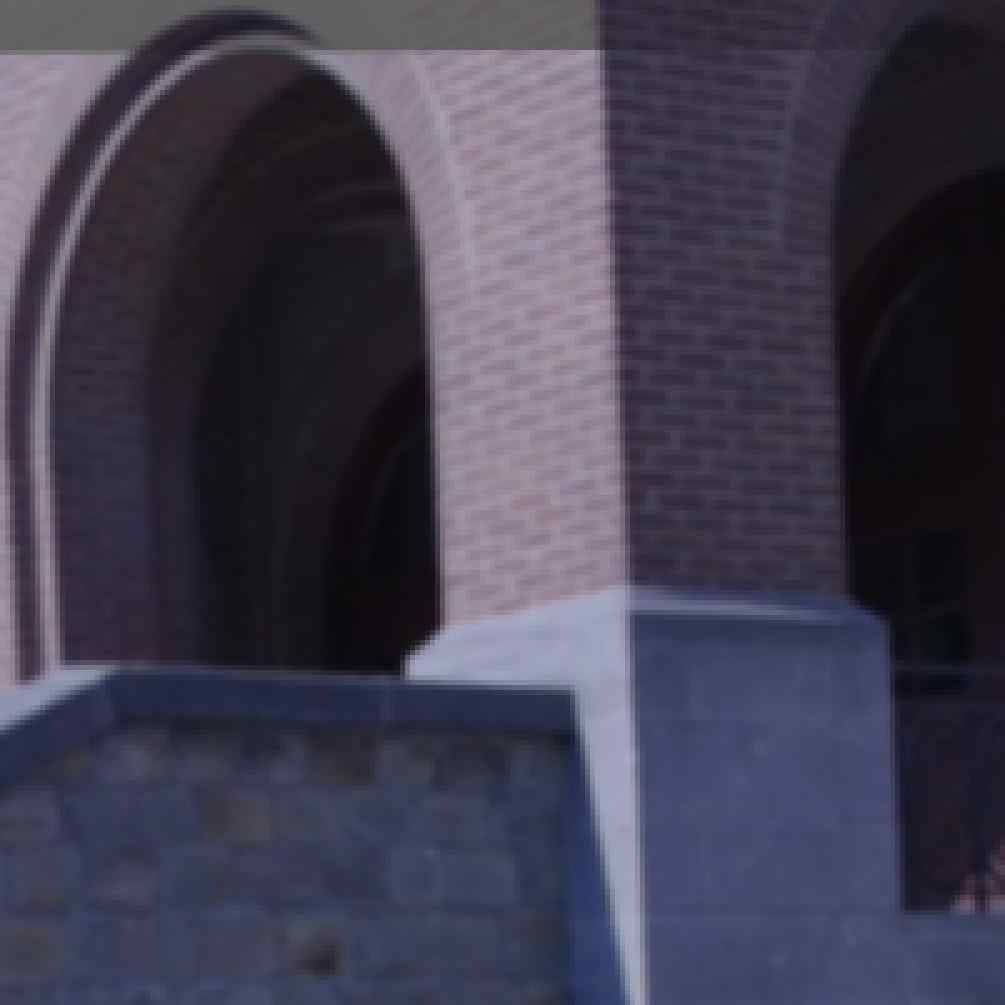}
\caption{Alignment result}
\end{subfigure}
\par\bigskip
\begin{subfigure}[b]{1.0\textwidth}
\begin{center}
\includegraphics[width=0.8\linewidth]{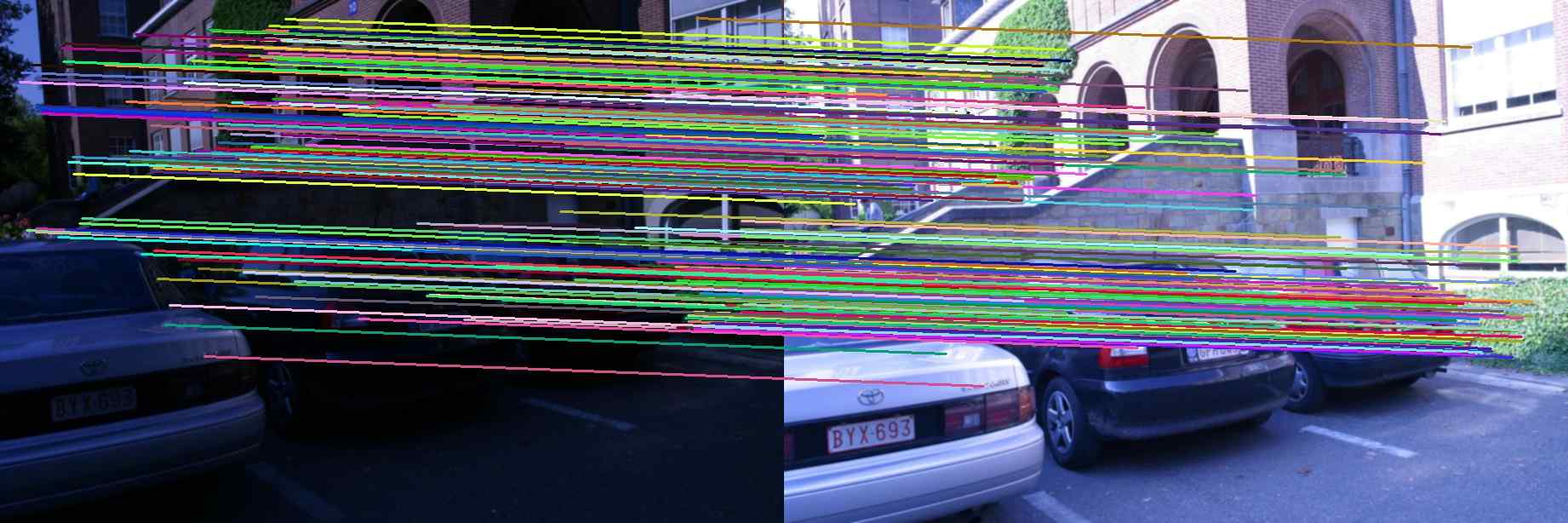}
\caption{Matching result}
\end{center}
\end{subfigure}
\caption{
\textsc{Tough} pair from HPatches (i\_leuven):
(a) \textsc{Left:} The two input images. \textsc{Middle:} The result in the upper row is the initial alignment. Here both images have been overlaid and blended with each other. The lower row shows the alignment computed by our estimated homography. \textsc{Right:} Zoomed-in views of a specific region of the image shows the quality of the initial and final alignments. (b) The sparse feature matches between the two images. These are computed by extracting SIFT keypoints and matching feature descriptors learned using our method. The matches are filtered using a geometric verification step that robustly fits a homography and finds inlier matches.
The outliers are not shown.
}
\end{figure*}

\begin{figure*}[t]
\begin{subfigure}[b]{1.0\textwidth}
\includegraphics[width=0.35\linewidth]{figs/results/img1_i_miniature}
\hfill
\includegraphics[width=0.35\linewidth]{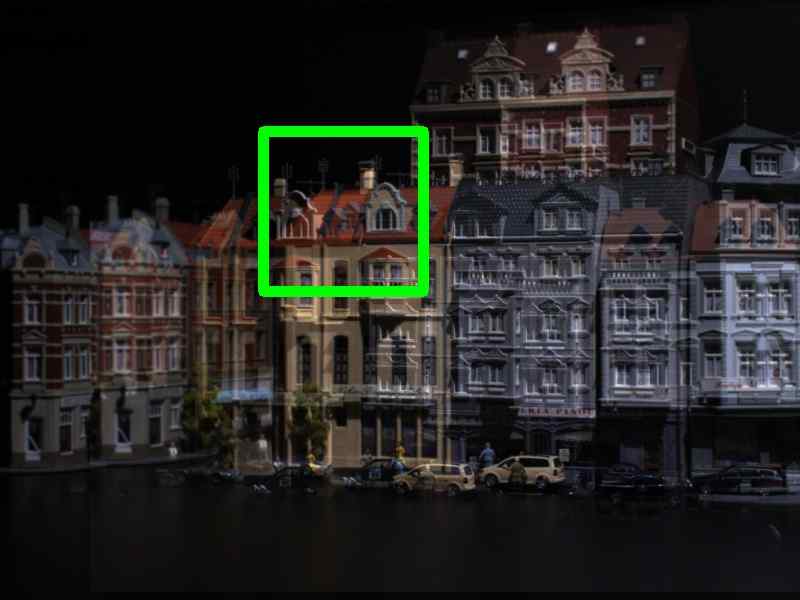}
\hfill
\includegraphics[width=0.26\linewidth]{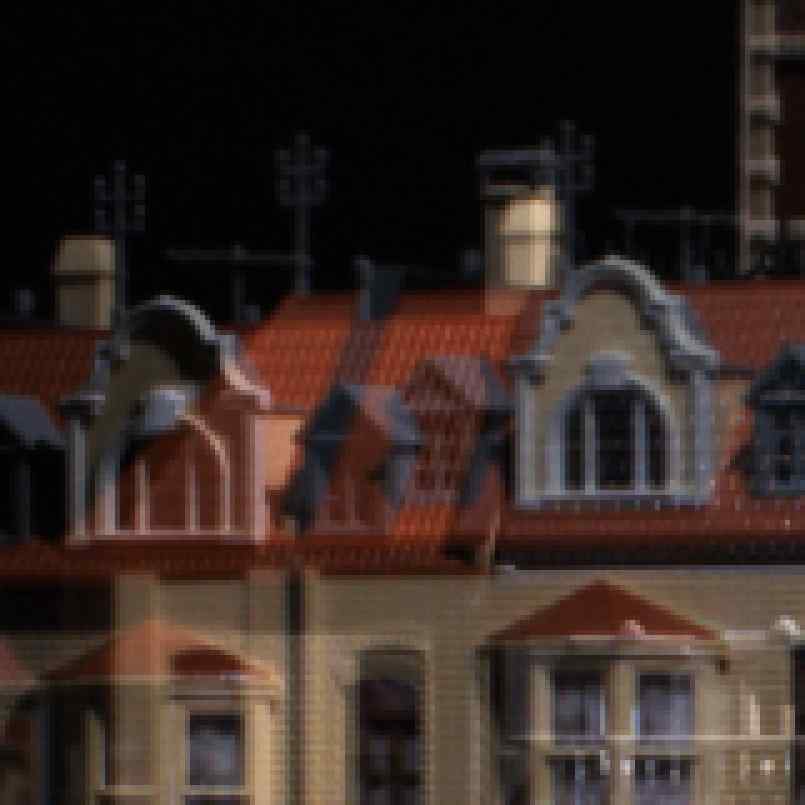}
\par\medskip
\includegraphics[width=0.35\linewidth]{figs/results/img2_i_miniature}
\hfill
\includegraphics[width=0.35\linewidth]{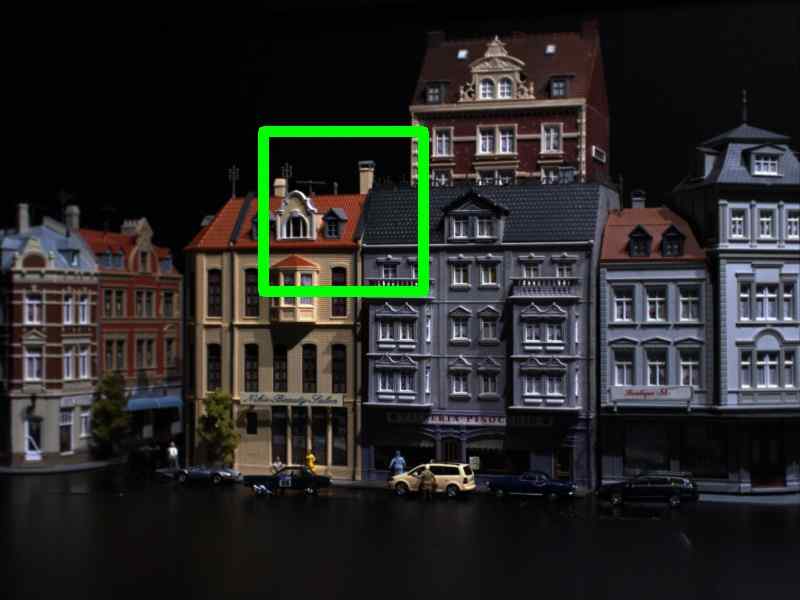}
\hfill
\includegraphics[width=0.26\linewidth]{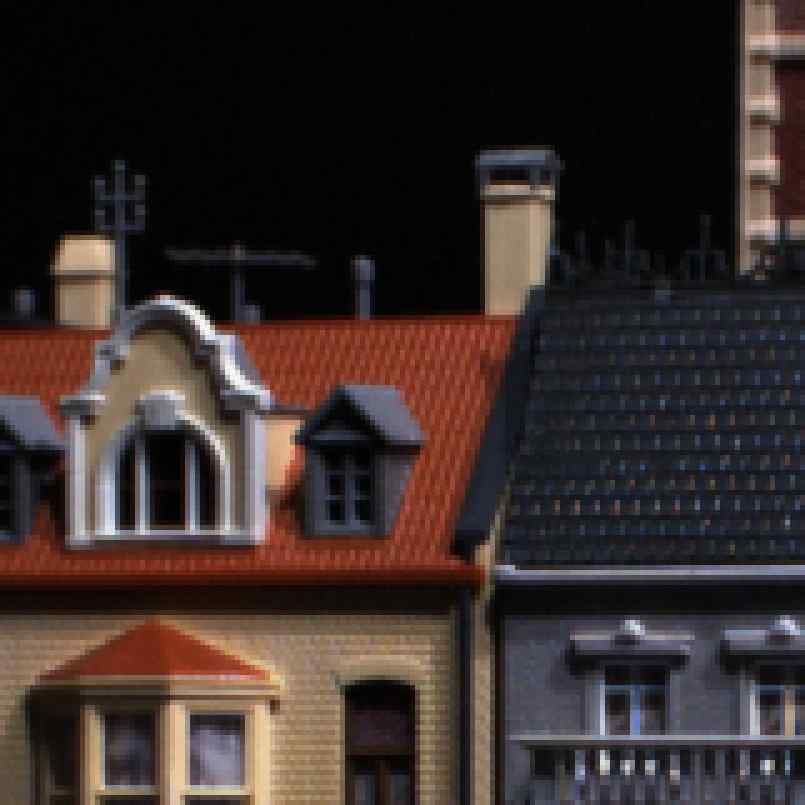}
\caption{Alignment result}
\end{subfigure}
\par\bigskip
\begin{subfigure}[b]{1.0\textwidth}
\begin{center}
\includegraphics[width=0.8\linewidth]{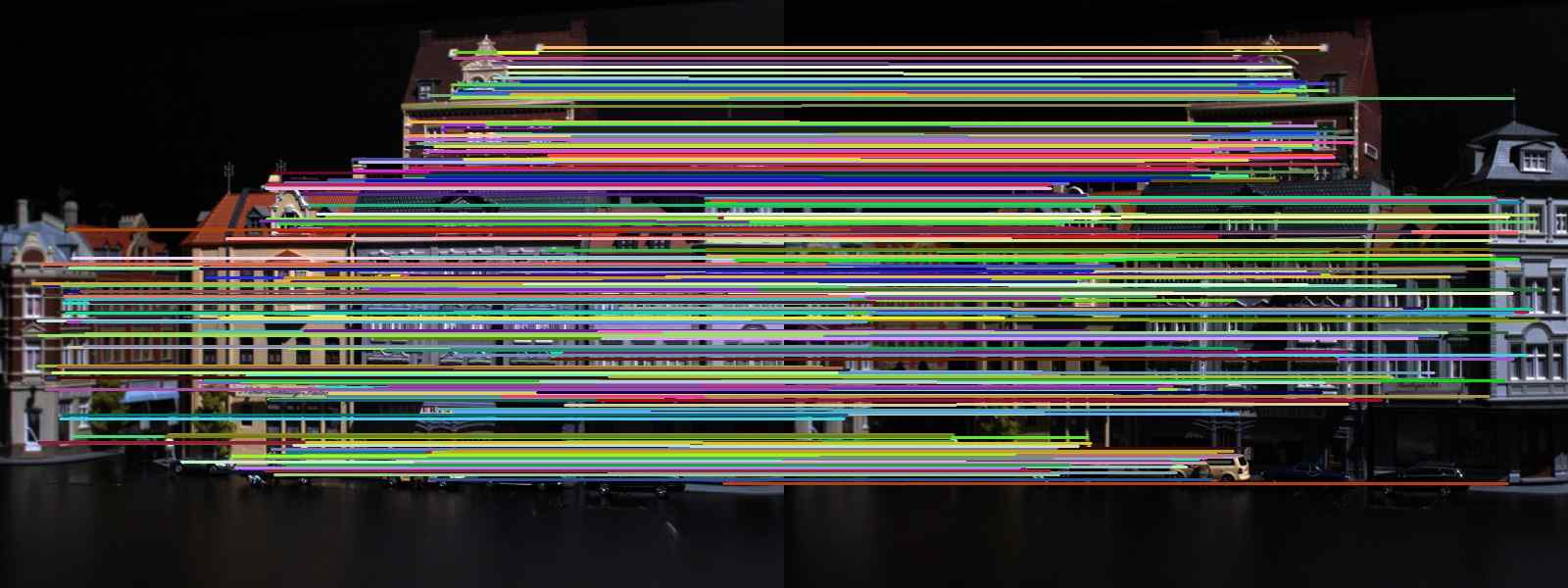}
\caption{Matching result}
\end{center}
\end{subfigure}
\caption{
\textsc{Tough} pair from HPatches (i\_miniature):
(a) \textsc{Left:} The two input images. \textsc{Middle:} The result in the upper row is the initial alignment. Here both images have been overlaid and blended with each other. The lower row shows the alignment computed by our estimated homography. \textsc{Right:} Zoomed-in views of a specific region of the image shows the quality of the initial and final alignments. (b) The sparse feature matches between the two images. These are computed by extracting SIFT keypoints and matching feature descriptors learned using our method. The matches are filtered using a geometric verification step that robustly fits a homography and finds inlier matches.
The outliers are not shown.
}
\end{figure*}

\begin{figure*}[t]
\begin{subfigure}[b]{1.0\textwidth}
\includegraphics[width=0.357\linewidth]{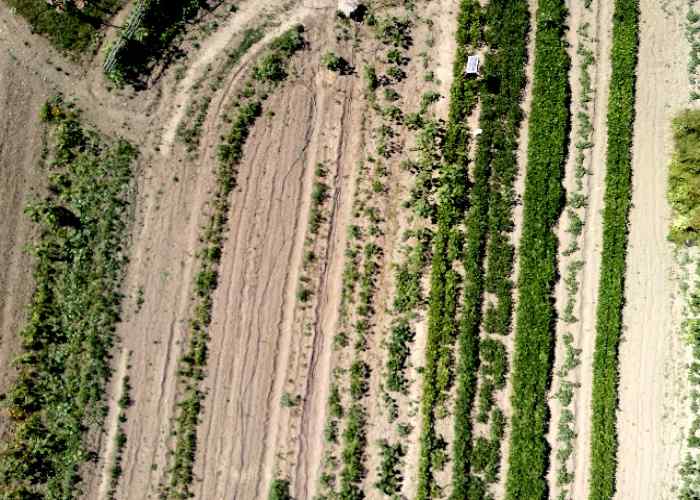}
\hfill
\includegraphics[width=0.357\linewidth]{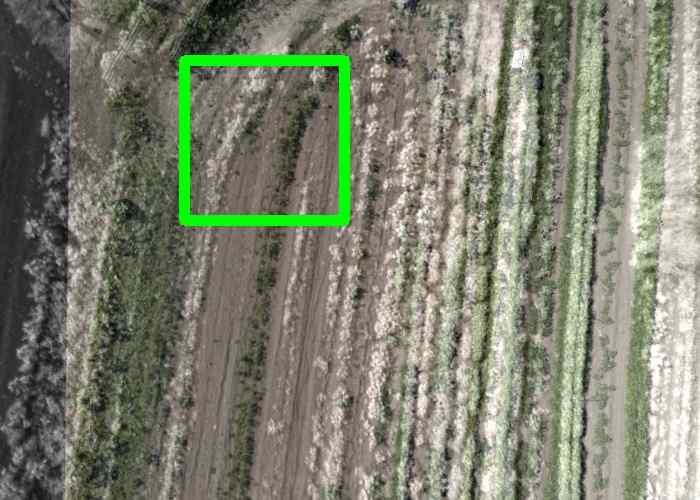}
\hfill
\includegraphics[width=0.255\linewidth]{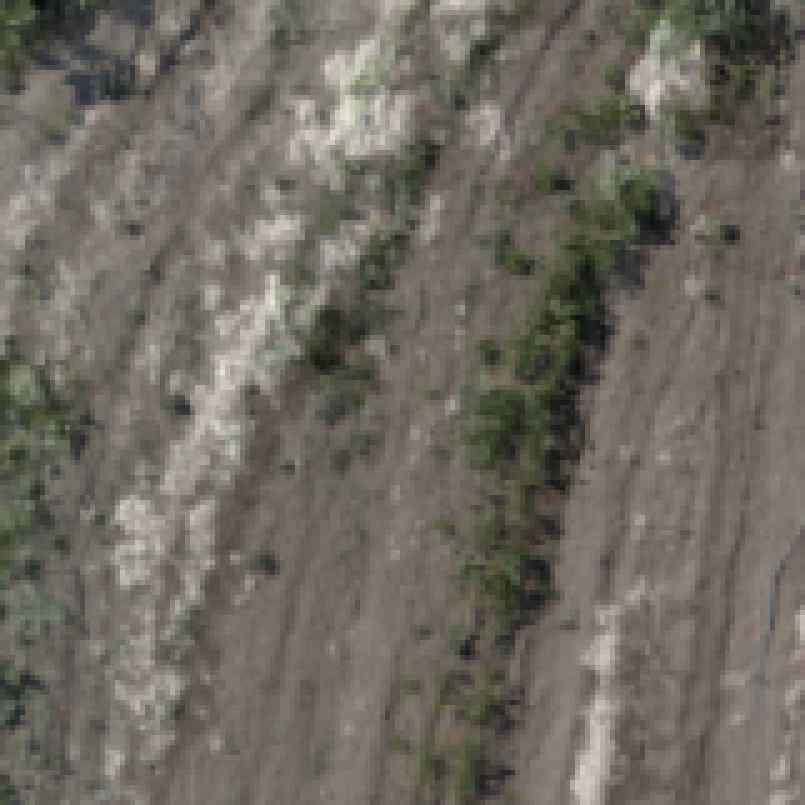}
\par\medskip
\includegraphics[width=0.357\linewidth]{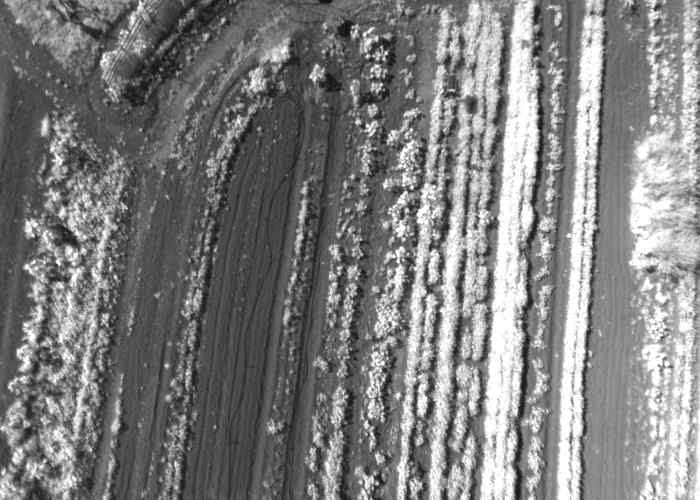}
\hfill
\includegraphics[width=0.357\linewidth]{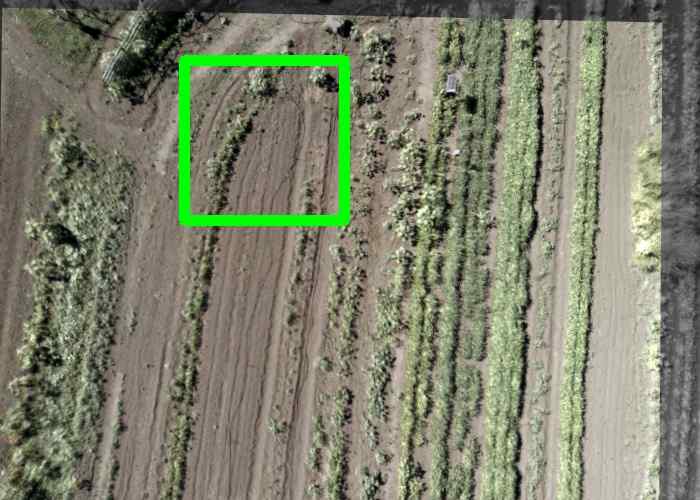}
\hfill
\includegraphics[width=0.255\linewidth]{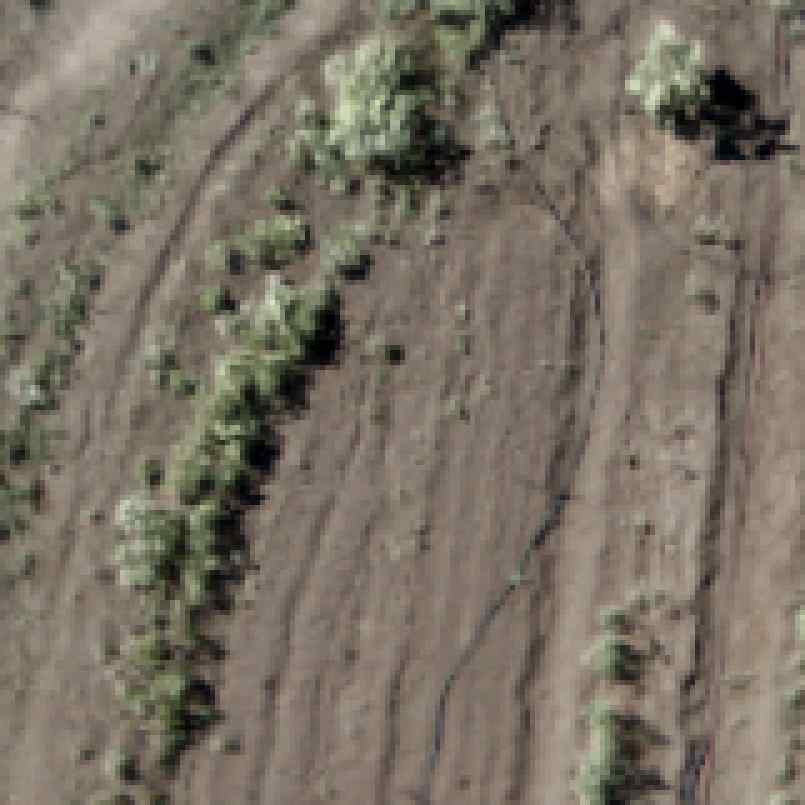}
\caption{Alignment result}
\end{subfigure}
\par\bigskip
\begin{subfigure}[b]{1.0\textwidth}
\begin{center}
\includegraphics[width=0.8\linewidth]{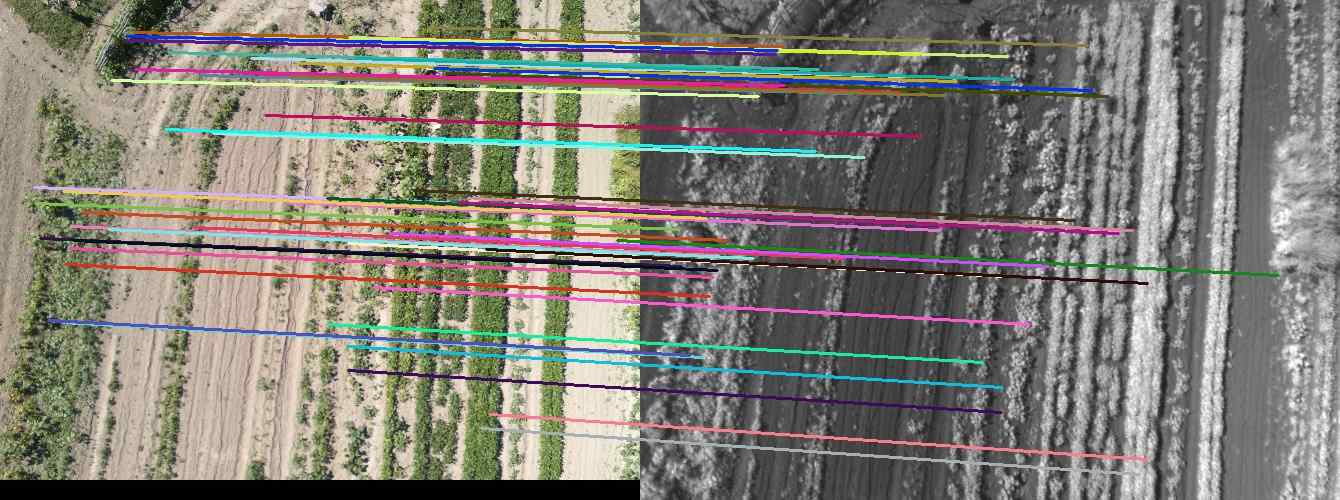}
\caption{Matching result}
\end{center}
\end{subfigure}
\caption{
RGB NIR image pair 1:
(a) \textsc{Left:} The two input images. \textsc{Middle:} The result in the upper row is the initial alignment. Here both images have been overlaid and blended with each other. The lower row shows the alignment computed by our estimated homography. \textsc{Right:} Zoomed-in views of a specific region of the image shows the quality of the initial and final alignments. (b) The sparse feature matches between the two images. These are computed by extracting SIFT keypoints and matching feature descriptors learned using our method. The matches are filtered using a geometric verification step that robustly fits a homography and finds inlier matches.
The outliers are not shown.
}
\end{figure*}

\begin{figure*}[t]
\begin{subfigure}[b]{1.0\textwidth}
\includegraphics[width=0.357\linewidth]{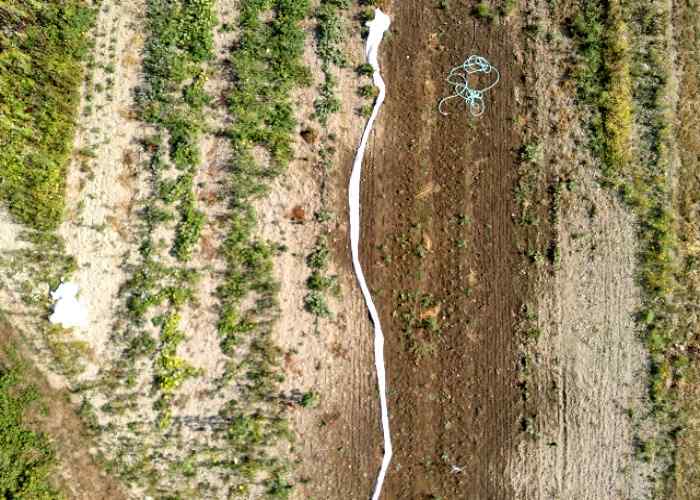}
\hfill
\includegraphics[width=0.357\linewidth]{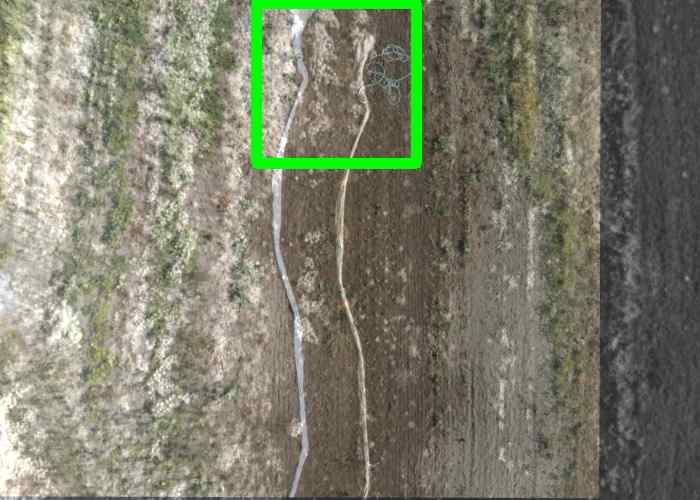}
\hfill
\includegraphics[width=0.255\linewidth]{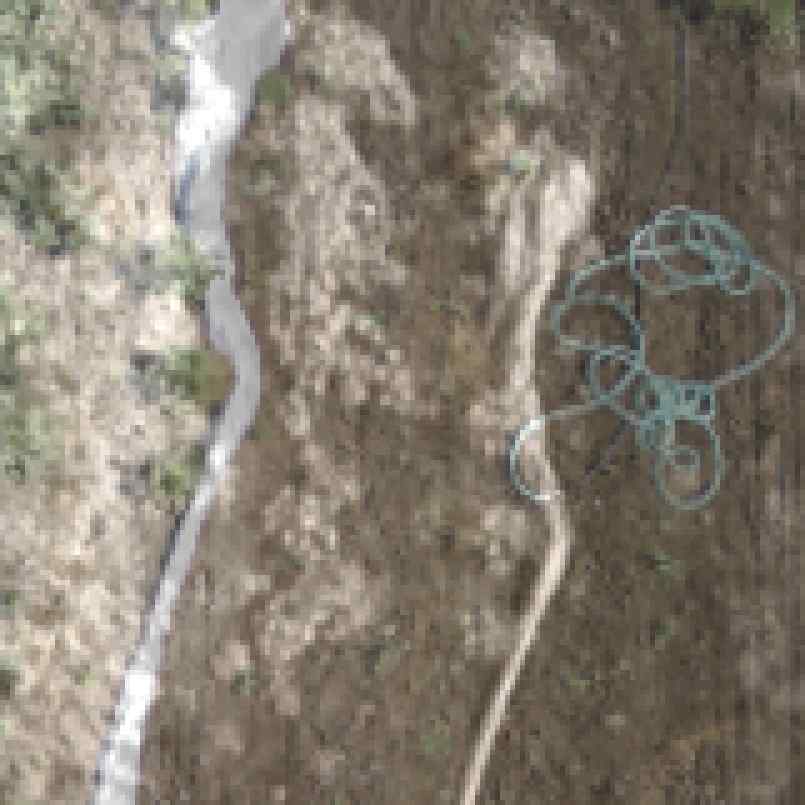}
\par\medskip
\includegraphics[width=0.357\linewidth]{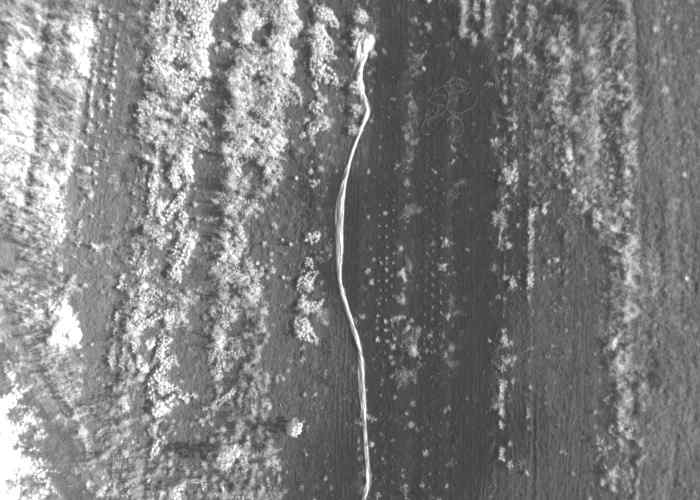}
\hfill
\includegraphics[width=0.357\linewidth]{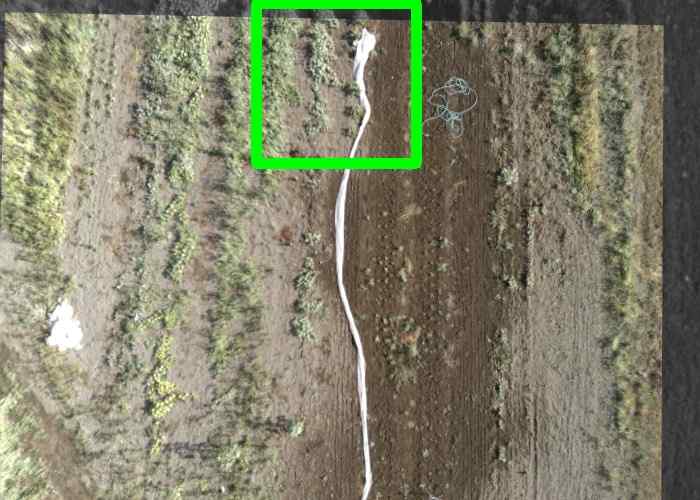}
\hfill
\includegraphics[width=0.255\linewidth]{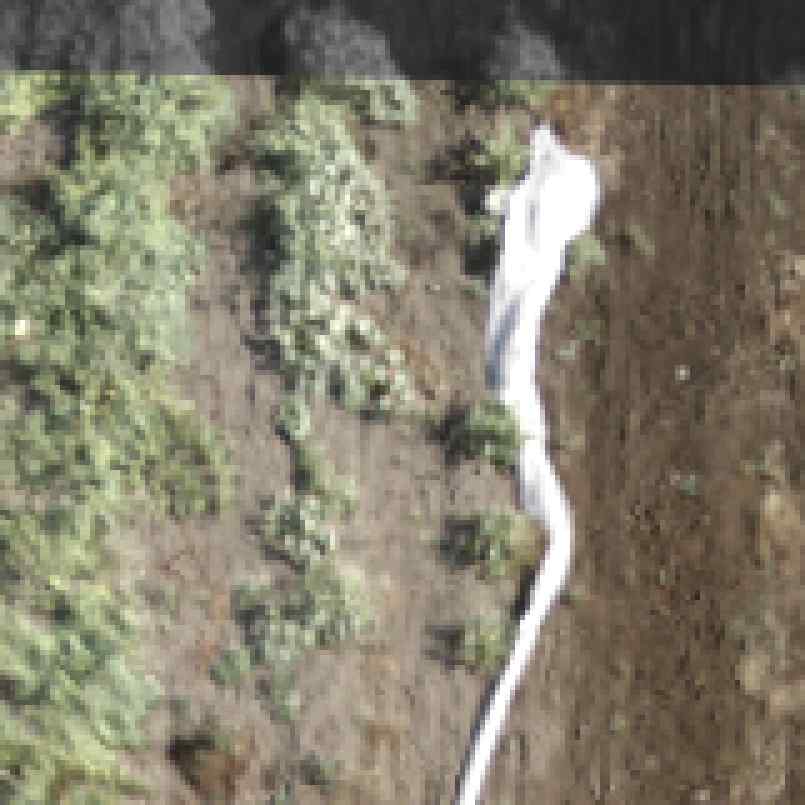}
\caption{Alignment result}
\end{subfigure}
\par\bigskip
\begin{subfigure}[b]{1.0\textwidth}
\begin{center}
\includegraphics[width=0.8\linewidth]{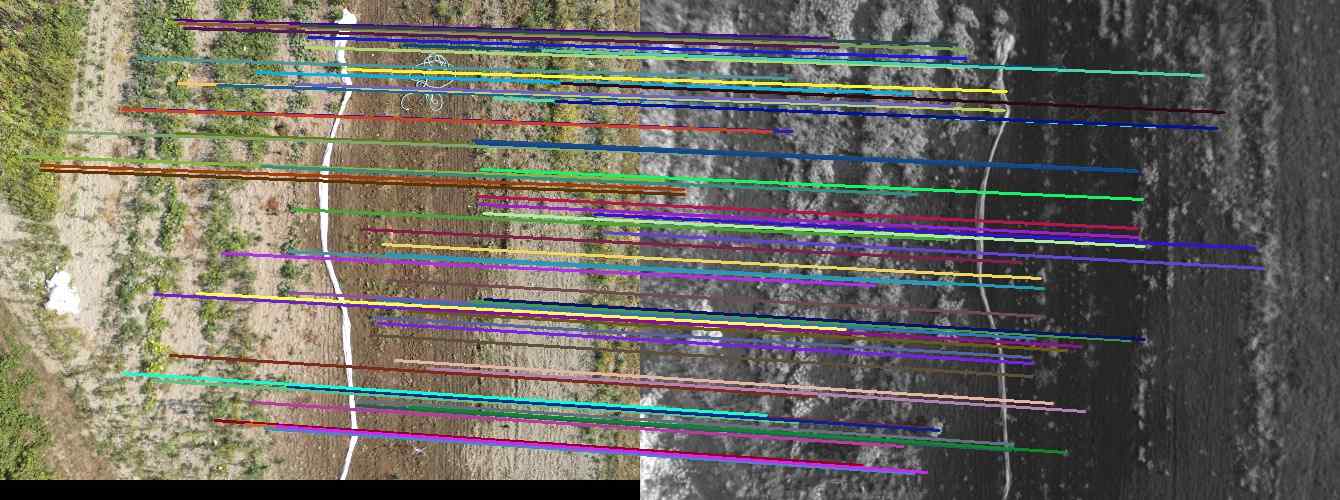}
\caption{Matching result}
\end{center}
\end{subfigure}
\caption{
RGB NIR image pair 2:
(a) \textsc{Left:} The two input images. \textsc{Middle:} The result in the upper row is the initial alignment. Here both images have been overlaid and blended with each other. The lower row shows the alignment computed by our estimated homography. \textsc{Right:} Zoomed-in views of a specific region of the image shows the quality of the initial and final alignments. (b) The sparse feature matches between the two images. These are computed by extracting SIFT keypoints and matching feature descriptors learned using our method. The matches are filtered using a geometric verification step that robustly fits a homography and finds inlier matches.
The outliers are not shown.
}
\end{figure*}

\begin{figure*}[t]
\begin{subfigure}[b]{1.0\textwidth}
\includegraphics[width=0.357\linewidth]{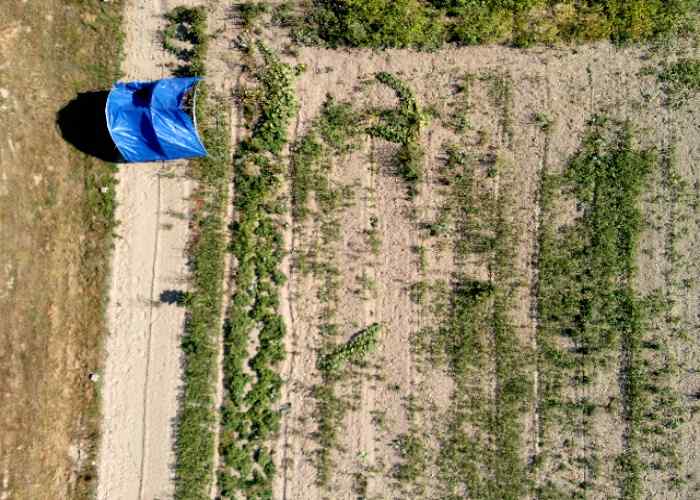}
\hfill
\includegraphics[width=0.357\linewidth]{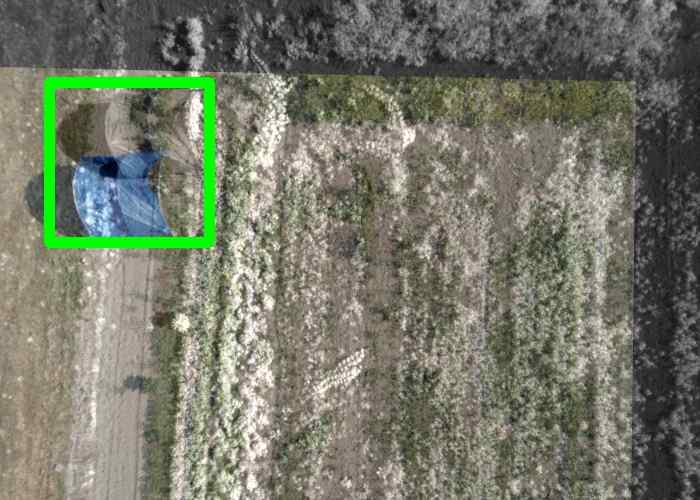}
\hfill
\includegraphics[width=0.255\linewidth]{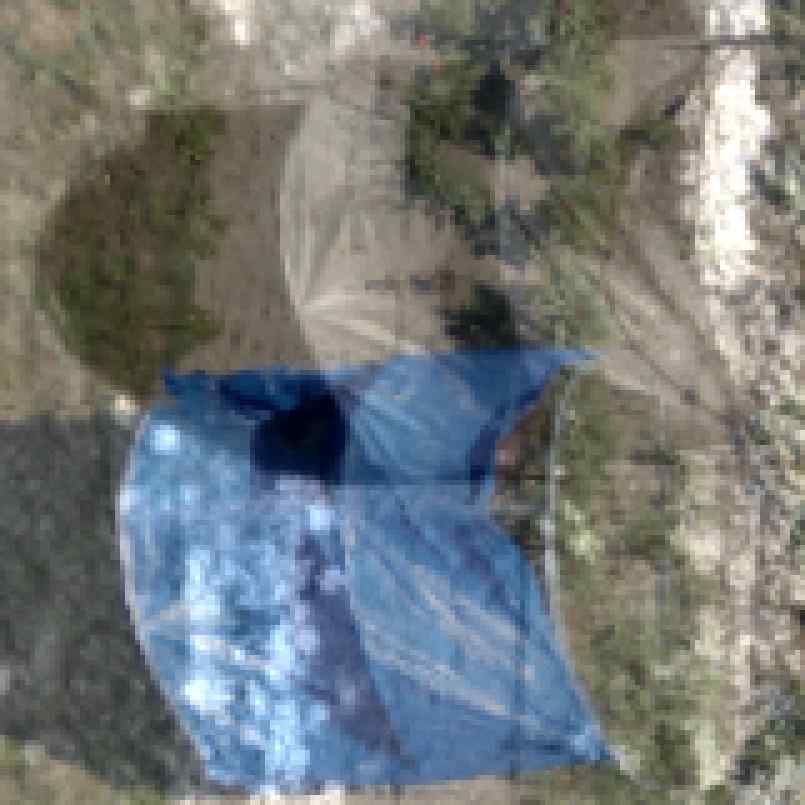}
\par\medskip
\includegraphics[width=0.357\linewidth]{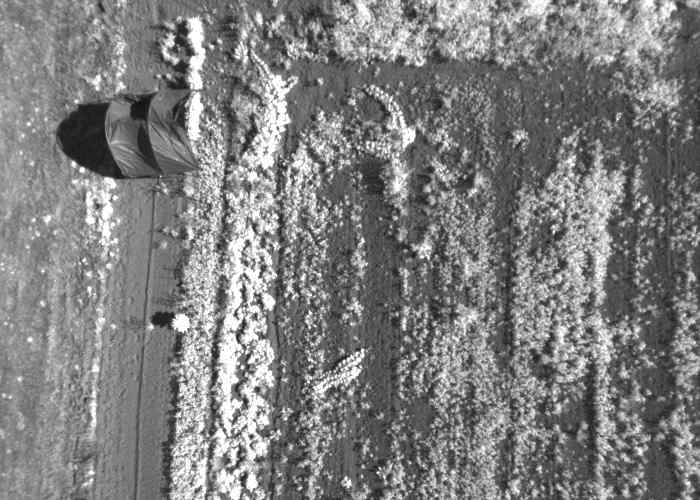}
\hfill
\includegraphics[width=0.357\linewidth]{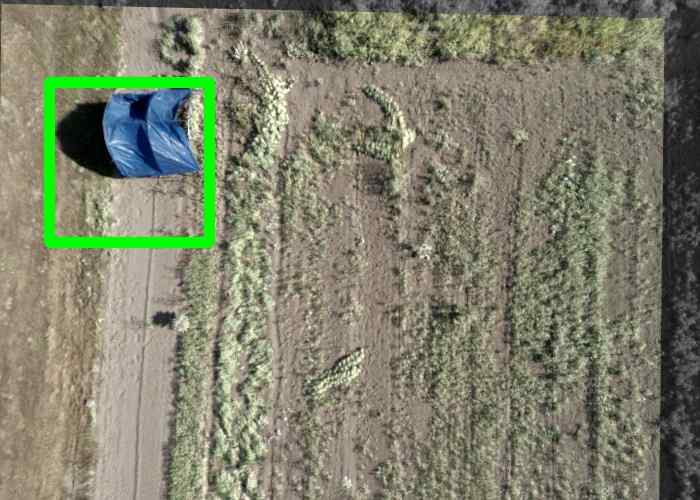}
\hfill
\includegraphics[width=0.255\linewidth]{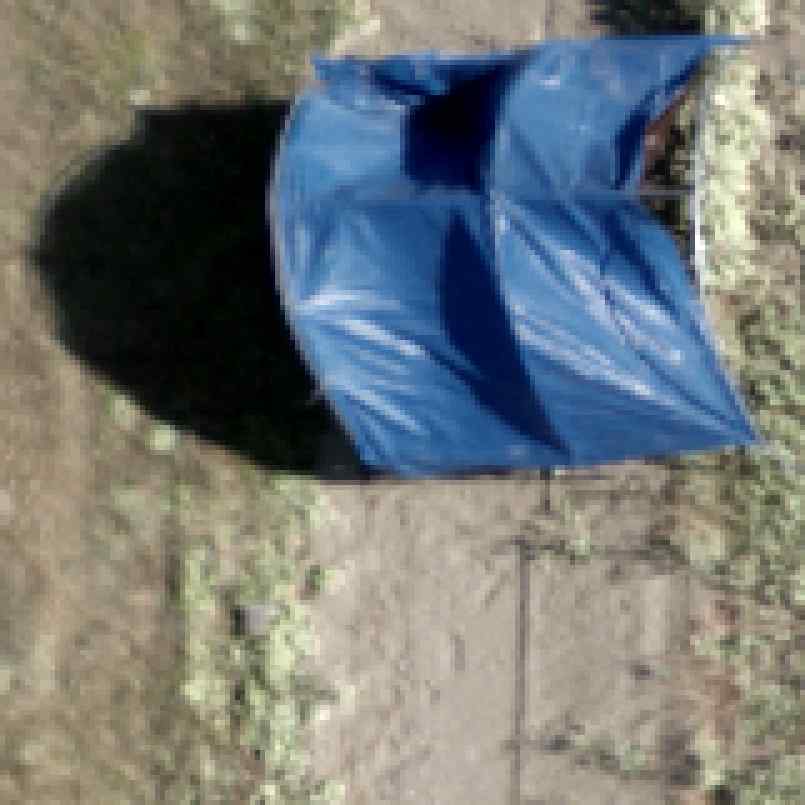}
\caption{Alignment result}
\end{subfigure}
\par\bigskip
\begin{subfigure}[b]{1.0\textwidth}
\begin{center}
\includegraphics[width=0.8\linewidth]{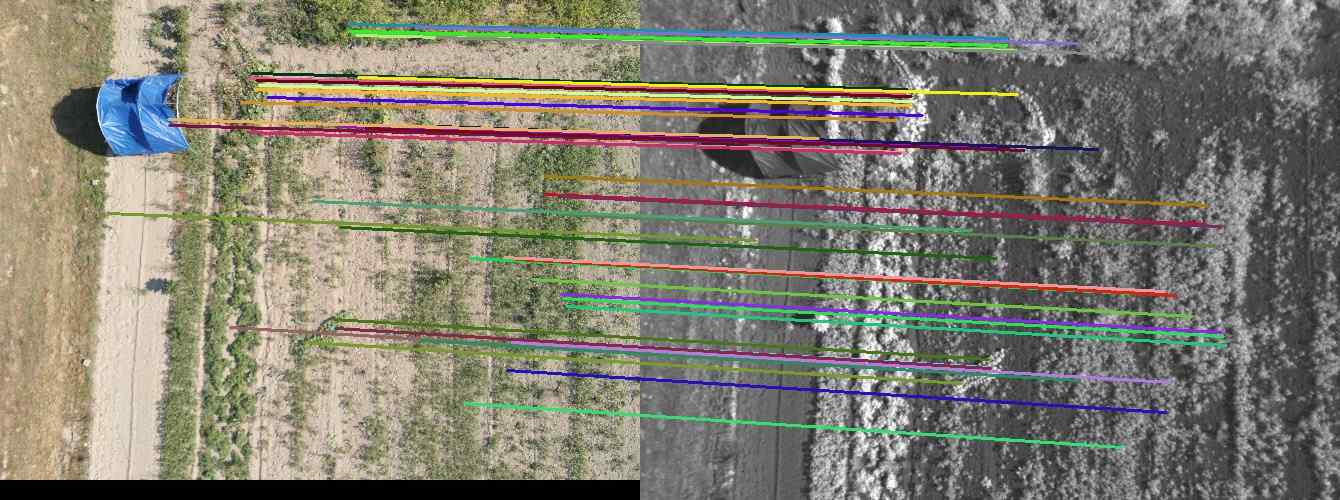}
\caption{Matching result}
\end{center}
\end{subfigure}
\caption{
RGB NIR image pair 3:
(a) \textsc{Left:} The two input images. \textsc{Middle:} The result in the upper row is the initial alignment. Here both images have been overlaid and blended with each other. The lower row shows the alignment computed by our estimated homography. \textsc{Right:} Zoomed-in views of a specific region of the image shows the quality of the initial and final alignments. (b) The sparse feature matches between the two images. These are computed by extracting SIFT keypoints and matching feature descriptors learned using our method. The matches are filtered using a geometric verification step that robustly fits a homography and finds inlier matches.
The outliers are not shown.}
\end{figure*}

\begin{figure*}[t]
\begin{subfigure}[b]{1.0\textwidth}
\includegraphics[width=0.357\linewidth]{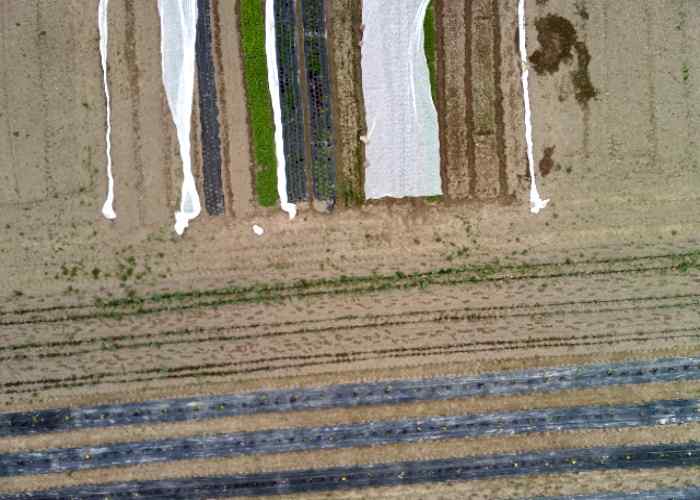}
\hfill
\includegraphics[width=0.357\linewidth]{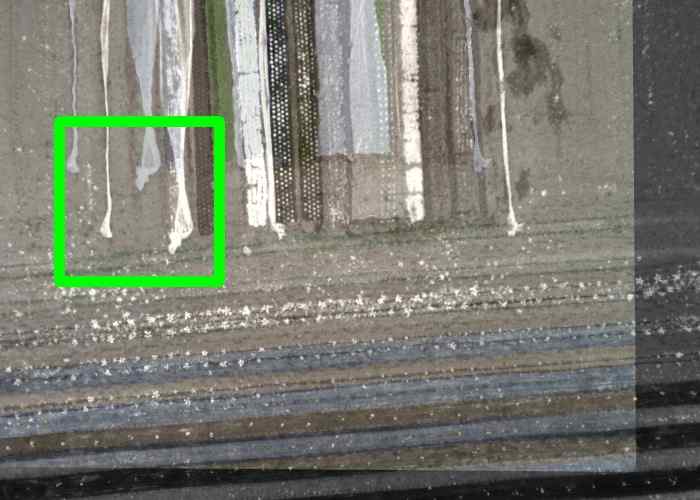}
\hfill
\includegraphics[width=0.255\linewidth]{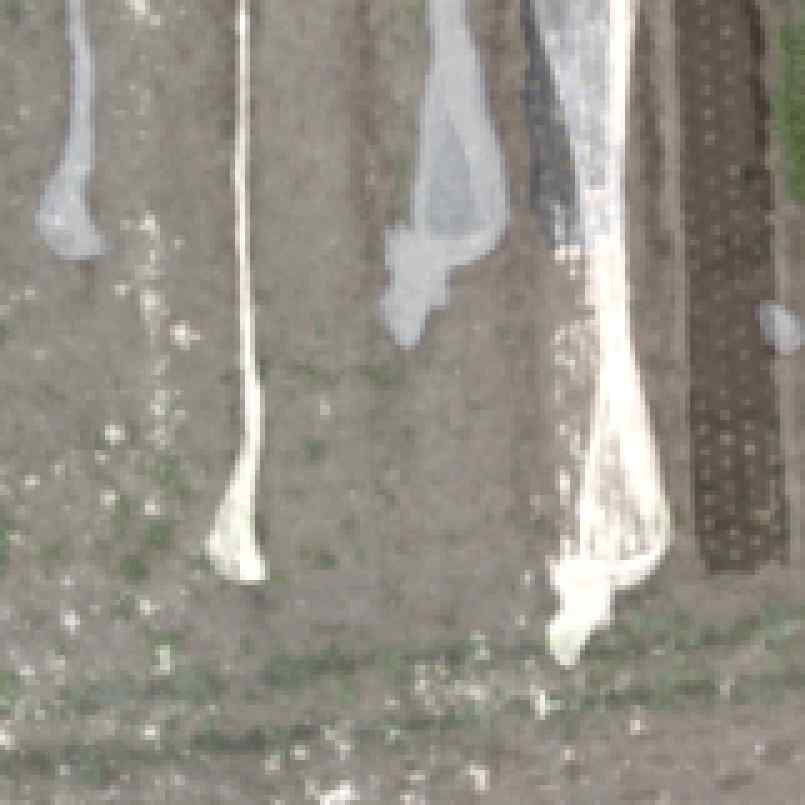}
\par\medskip
\includegraphics[width=0.357\linewidth]{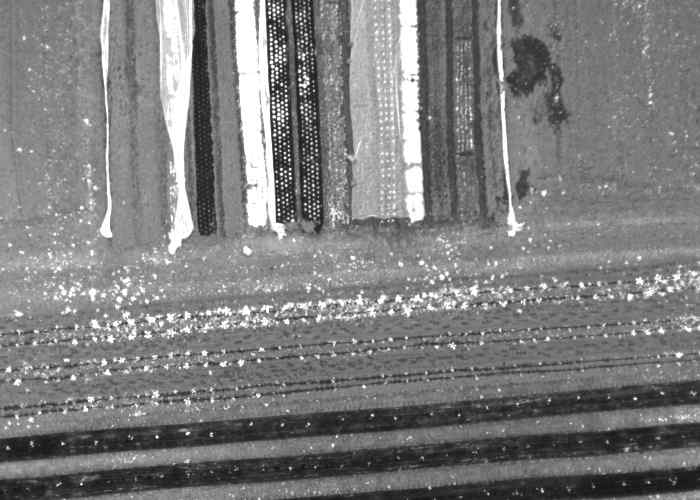}
\hfill
\includegraphics[width=0.357\linewidth]{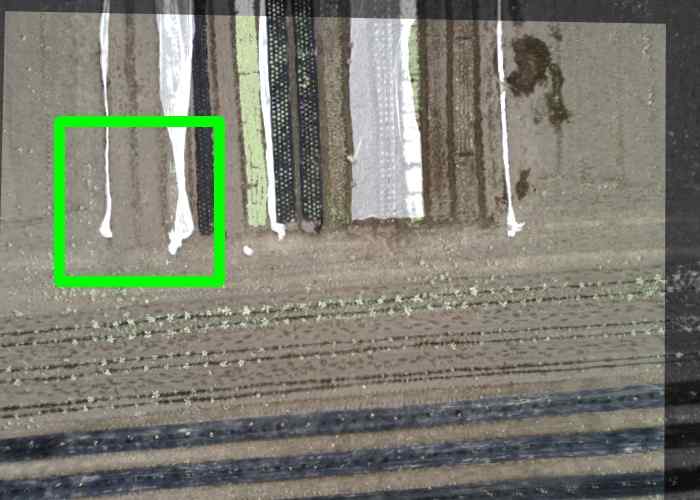}
\hfill
\includegraphics[width=0.255\linewidth]{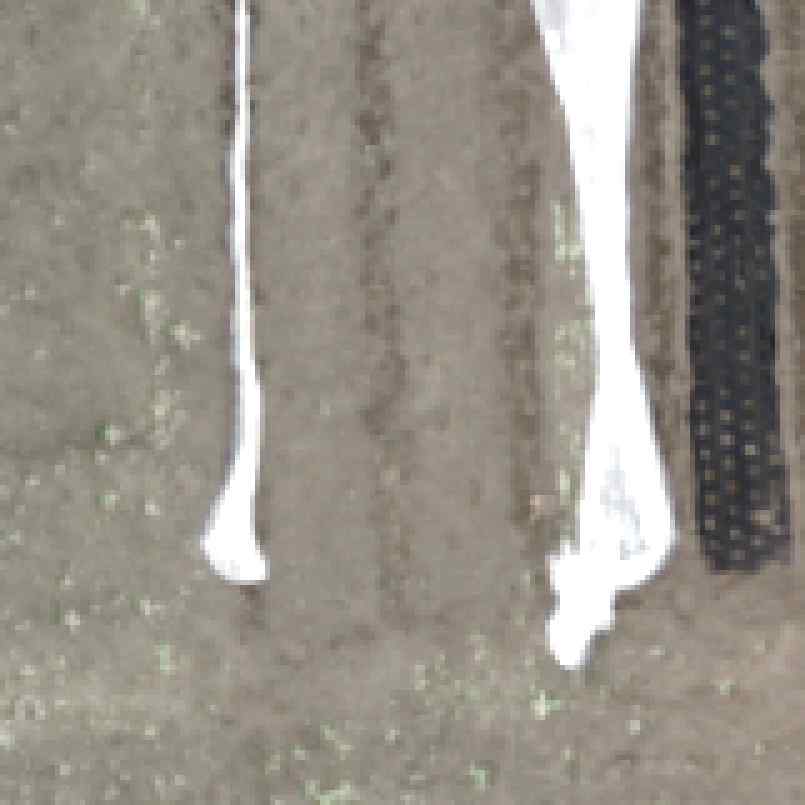}
\caption{Alignment result}
\end{subfigure}
\par\bigskip
\begin{subfigure}[b]{1.0\textwidth}
\begin{center}
\includegraphics[width=0.8\linewidth]{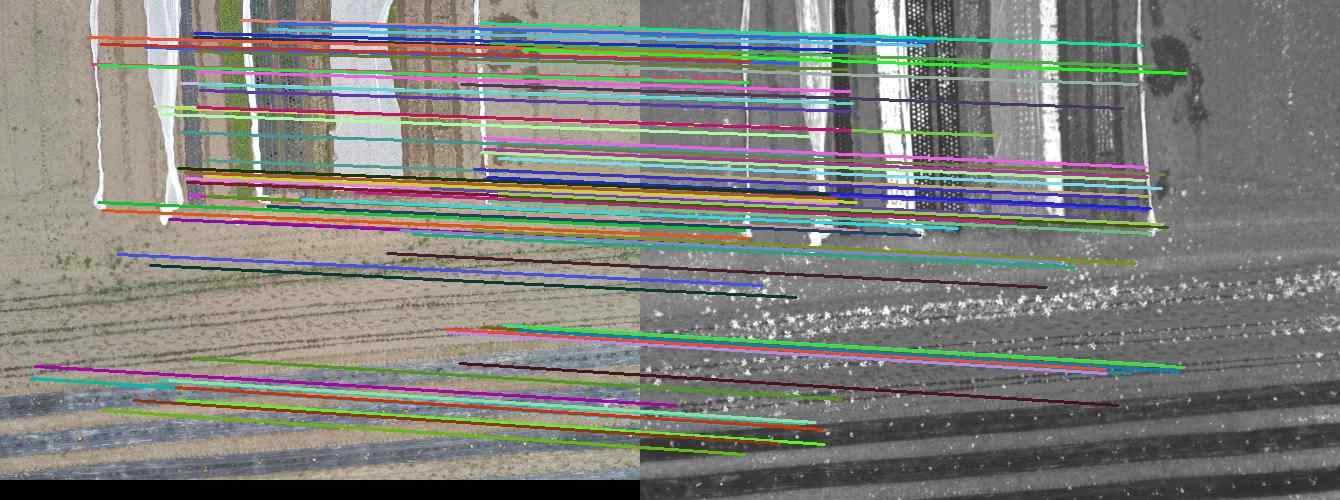}
\caption{Matching result}
\end{center}
\end{subfigure}
\caption{
RGB NIR image pair 4:
(a) \textsc{Left:} The two input images. \textsc{Middle:} The result in the upper row is the initial alignment. Here both images have been overlaid and blended with each other. The lower row shows the alignment computed by our estimated homography. \textsc{Right:} Zoomed-in views of a specific region of the image shows the quality of the initial and final alignments. (b) The sparse feature matches between the two images. These are computed by extracting SIFT keypoints and matching feature descriptors learned using our method. The matches are filtered using a geometric verification step that robustly fits a homography and finds inlier matches.
The outliers are not shown.
}
\end{figure*}

\end{document}